\documentclass{article}

\usepackage[eandd, final]{neurips_2026}
\usepackage[utf8]{inputenc}
\usepackage[T1]{fontenc}
\usepackage{hyperref}
\usepackage{url}
\usepackage{booktabs}
\usepackage{amsfonts}
\usepackage{nicefrac}
\usepackage{microtype}
\usepackage{xcolor}
\usepackage{colortbl}
\usepackage{graphicx}
\usepackage{wrapfig}
\usepackage{placeins}
\usepackage{multirow}
\usepackage[normalem]{ulem}
\usepackage{array}
\usepackage[version=4]{mhchem}

\definecolor{rmseMulti}{HTML}{4E79A7}
\definecolor{rmseTransfer}{HTML}{F28E2B}
\definecolor{rmseSingle}{HTML}{59A14F}
\colorlet{rmseMultiLight}{rmseMulti!60!white}
\colorlet{rmseTransferLight}{rmseTransfer!60!white}
\colorlet{rmseSingleLight}{rmseSingle!60!white}
\colorlet{rmseMultiDark}{rmseMulti!25!black}
\colorlet{rmseTransferDark}{rmseTransfer!25!black}
\colorlet{rmseSingleDark}{rmseSingle!25!black}
\newcommand{\rmseMM}[2]{\textcolor{rmseMultiDark!#1!rmseMultiLight}{#2}}
\newcommand{\rmseTR}[2]{\textcolor{rmseTransferDark!#1!rmseTransferLight}{#2}}
\newcommand{\rmseSG}[2]{\textcolor{rmseSingleDark!#1!rmseSingleLight}{#2}}
\colorlet{metricMultiLight}{rmseMulti!70!white}
\colorlet{metricTransferLight}{rmseTransfer!70!white}
\colorlet{metricSingleLight}{rmseSingle!70!white}
\newcommand{\metricMM}[2]{\textcolor{rmseMultiDark!#1!metricMultiLight}{#2}}
\newcommand{\metricTR}[2]{\textcolor{rmseTransferDark!#1!metricTransferLight}{#2}}
\newcommand{\metricSG}[2]{\textcolor{rmseSingleDark!#1!metricSingleLight}{#2}}

\definecolor{esOrange}{HTML}{E06C00}

\usepackage{bm,amssymb,amsmath}

\makeatletter
\def\th@plain{%
  \thm@notefont{}% same as heading font
  \itshape % body font
}
\def\th@definition{%
  \thm@notefont{}% same as heading font
  \normalfont % body font
}
\makeatother

\def\1{\bm{1}}

\DeclareMathAlphabet{\mathsfit}{\encodingdefault}{\sfdefault}{m}{sl}
\SetMathAlphabet{\mathsfit}{bold}{\encodingdefault}{\sfdefault}{bx}{n}

\title{ThousandWorlds: A benchmark for climate emulation of potentially habitable exoplanets}

\author{%
  Edward T. Stevenson \\
  \normalfont University of Cambridge \\
  \texttt{es833@cam.ac.uk} \\
  \And
  Mei Ting Mak \\
  \normalfont University of Oxford \\
  \And
  Eric Wolf \\
  \normalfont University of Colorado Boulder \\
  \And
  Denis E. Sergeev \\
  \normalfont University of Bristol \\
  \And
  Tobi Hammond \\
  \normalfont Purdue University \\
  \And
  N. J. Mayne \\
  \normalfont University of Exeter \\
  \And
  Miles Cranmer \\
  \normalfont University of Cambridge \\
}

\begin{document}

\maketitle

\begin{abstract}

The search for life beyond Earth will depend on detecting faint signatures in the atmospheres of potentially habitable exoplanets.
Interpreting those signatures requires understanding the host planet's climate: the same molecule may signal life on one planet and abiotic chemistry on another.
Global climate models (GCMs) provide this understanding, but individual runs can require up to millions of core-hours and substantial domain expert time.
Machine-learning emulators could remove this bottleneck, but progress has been limited by the absence of a curated, multi-model exoclimate dataset.
We introduce ThousandWorlds, an ML-ready benchmark for exoclimate emulation and for the broader regime of low-data, multi-simulator, parameter-to-field regression.
The dataset contains approximately 1,700 simulations from five GCMs, mapping eight planet parameters to 3D atmospheric fields including temperature, humidity, winds, clouds, and radiation.
Three nested subsets define progressively harder challenges: single-simulator regression, multi-simulator regression with complete observations, and multi-simulator regression with structured missingness.
We propose two evaluation protocols: one for ranking methods, and one that measures performance relative to the disagreement between GCMs themselves.
We evaluate ten baselines spanning simple methods, trees, deep learning, and Gaussian processes.
GP-based methods perform best, suggesting that ThousandWorlds exposes a regime where off-the-shelf deep learning does not yet succeed. \newline
Data: \url{https://doi.org/10.57967/hf/8695}. \newline Code: \url{https://github.com/astroautomata/ThousandWorlds}.
\end{abstract}

\section{Introduction}\label{sec:intro}

We may be the first generation able to answer whether life exists beyond Earth.
The most promising hosts, rocky habitable-zone planets, are common in the galaxy \citep{dressingOCCURRENCEPOTENTIALLY2015}, and JWST has just begun observing the atmospheres of the nearest candidates.
The search for life will hinge on the detection of \emph{biosignatures} -- molecular fingerprints of life imprinted as absorption lines in observed spectra.
But these signatures are ambiguous -- does an O$_2$ detection mean life, or just photodissociation of water \citep{wordsworthAbioticOxygendominated2014}?
Their interpretation requires knowledge of the planet's climate.
Temperature, circulation, clouds, and heat transport all matter, and predicting them accurately requires sophisticated 3D modelling.

This has driven an increasing use of \emph{global climate models} (GCMs)\footnote{The acronym ``GCM'' can stand for either \textbf{g}lobal \textbf{c}limate \textbf{m}odel or \textbf{g}eneral \textbf{c}irculation \textbf{m}odel; both refer to the same class of models.}, large numerical codes that simulate 3D atmospheric fluid dynamics alongside non-dynamical processes like clouds and radiation.
But a single GCM simulation typically costs $10^4\text{--}10^6$ core-hours plus substantial domain expert time to configure and monitor, limiting studies to small ensembles of hand-picked configurations.
An emulator that produces near-instant climate predictions would remove this bottleneck, opening the door to large parameter sweeps, principled uncertainty quantification, and integration with observational inference pipelines.

Despite this need, exoclimate emulation remains largely unexplored.
The main barrier has been the absence of a curated, multi-model dataset: the raw simulations exist, produced by different groups running different GCMs for different scientific questions, but they are scattered across studies in incompatible formats, on different vertical grids, with different output variables.

Such a situation is not unique to exoclimate: across the sciences, many emulation problems share the same difficult structure: a handful of input parameters, high-dimensional structured outputs, scarce simulator evaluations, and several imperfect simulators.
Existing scientific ML benchmarks cover parts of this landscape, but largely target data-rich field-to-field prediction regimes where deep learning succeeds.
The complementary regimes of parameter-to-field regression, data scarcity, and multi-simulator learning are comparatively overlooked.

We introduce ThousandWorlds, a benchmark that sits at the intersection of these three regimes and fulfils the domain need for an ML-ready exoclimate dataset.
Developed in collaboration with exoplanet climate scientists, the dataset comprises approximately 1,700 simulations from five GCMs, spanning totally ice-covered snowball worlds to steamy moist greenhouses.
Each maps eight planet parameters to 3D atmospheric variables covering temperature, humidity, winds, clouds, and radiation.
Three nested subsets stage increasingly demanding challenges: (1) single-simulator regression, (2) multi-simulator transfer with complete observations, and (3) multi-simulator transfer with structured missingness -- the full dataset.

We define two evaluation protocols.
The \emph{standard} protocol uses a larger test set for comparing methods against one another.
The \emph{shared-planets} protocol measures how the emulator's error compares to the spread between high-fidelity GCMs for identical planets.
This spread reflects epistemic uncertainty about the underlying physics, so the shared-planets protocol provides a clearer measure of the scientific utility of an emulator.

We evaluate ten baselines spanning simple methods, trees, deep learning, and Gaussian processes (GPs).
The GP-based methods prove to be the strongest baselines, suggesting that this regime poses challenges for standard deep learning.

\section{Related work}\label{sec:related}

Shared benchmarks have become central to scientific machine learning.
PDEBench \citep{takamotoPDEBENCHExtensive2024} and The Well \citep{ohanaWellLargeScale2025} provide large-scale benchmarks for learning from simulated spatiotemporal physical systems, while RealPDEBench \citep{huRealPDEBenchBenchmark2026} pairs real-world measurements with numerical simulations for sim-to-real evaluation.
CFDBench \citep{luoCFDBenchLargeScale2024} and FlowBench \citep{taliFlowBenchLarge2024} benchmark flow prediction over varied geometries.
The literature on multi-fidelity surrogates also provides a closely related methodological context, but recent surveys rely on synthetic or bespoke test cases rather than shared community datasets \citep{fernandez-godinoReviewMultifidelity,brunelSurveyMultifidelity2025}.

Earth-system ML provides the closest domain precedent to ThousandWorlds, and here benchmark datasets have already driven rapid progress.
WeatherBench/WeatherBench2 \citep{raspWeatherBenchBenchmark2020,raspWeatherBench22024} standardize data-driven medium-range weather forecasting, while ClimSim/ClimSim-Online \citep{yuClimSimLarge2023, yuClimSimOnlineLarge2024} and ClimART \citep{cachayClimARTBenchmark2021} target component emulation inside climate models, for sub-grid atmospheric physics and radiative transfer respectively.
ClimateBench \citep{watson-parrisClimateBenchV102022} maps forcing inputs to annual-mean spatial climate fields, sharing ThousandWorlds' parameter-to-field structure, but in a data-rich, single-simulator setting.
ClimateSet \citep{kaltenbornClimateSetLargeScale2023} and ClimateSuite \citep{irvinSpatiotemporalPyramid2025} extend the climate change benchmarks line to multiple simulators: ClimateSet assembles inputs and outputs from 36 CMIP6 Earth system models (ESMs) and benchmarks climate emulation from gridded forcing-emission trajectories to monthly global temperature and precipitation fields. 
ClimateSuite further scales multi-simulator data to 33{,}000 simulation-years across ten ESMs.
One can view ThousandWorlds as a benchmark within this climate modelling tradition, but with a different source of diversity: rather than varying forcings on a single planet, the planet itself varies, producing hugely diverse climate states by Earth-modelling standards.

In astronomy broadly, the CAMELS project \citep{villaescusa-navarroCAMELSProject2023} provides an adjacent precedent, assembling thousands of cosmological simulations and ML-ready multifield maps across different simulators.
Within exoplanet astronomy, the only prior work on 3D exoclimate emulation is \citet{plaschzugAcceleratingExoplanet2025}, who train a pointwise emulator for hot Jupiter climates on 60 simulations from a single GCM.
Other ML work on exoclimate has targeted individual components within GCMs (e.g., \citealp{tahseenEnhancing3D2024,malskyAcceleratingRadiative2025}), rather than whole GCMs themselves.
\citet{rothHotJupiter2024} provide the closest large-dataset precedent, with 345 hot Jupiter simulations.

ThousandWorlds is one of few benchmarks to combine parameter-to-field regression, multi-simulator transfer, and structured missingness in a single dataset, and the first large ML-ready dataset for emulating potentially habitable exoplanet climates.

\section{The ThousandWorlds dataset}\label{sec:dataset}

\subsection{Task description}\label{sec:task}

\vspace{-2pt}
\begin{figure}[t]
  \centering
  \includegraphics[width=0.94\linewidth]{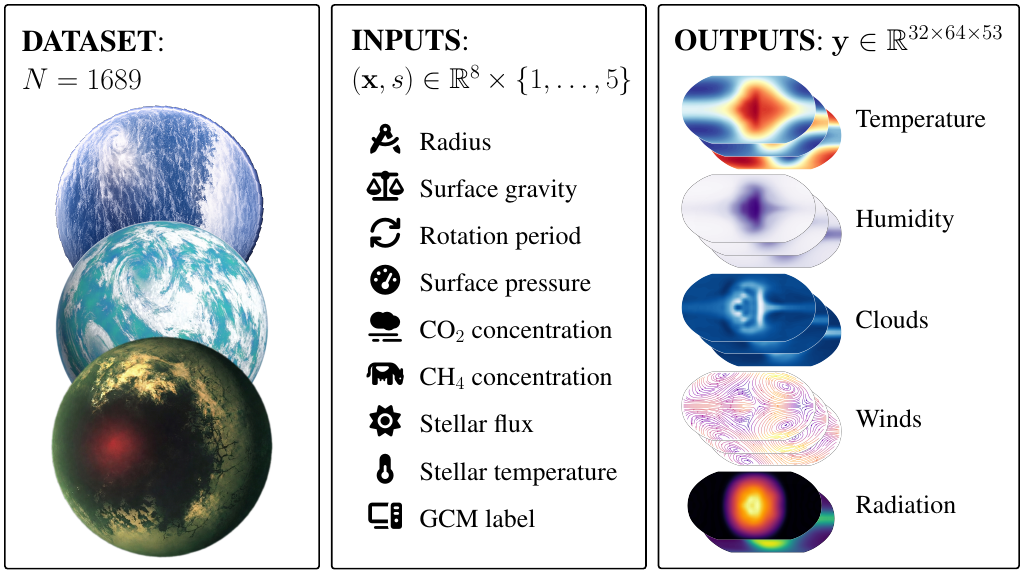}
  \vspace{-4pt}
  \caption{Dataset overview. Output fields are defined on a $32\times64$ latitude--longitude grid. There are 53 fields in total and $\sim 10^5$ output dimensions.}
  \label{fig:schema}
  \vspace{-4pt}
\end{figure}

\begin{wraptable}{r}{0.42\textwidth}
  \vspace{-20pt}
  \centering
  \small
  \setlength{\tabcolsep}{2pt}
  \renewcommand{\arraystretch}{1.05}
  \caption{Constraints on the eight continuous inputs that define the core domain.}
  \vspace{2pt}
  \label{tab:physical-domains}
  \begin{tabular}{@{}l>{\raggedleft\arraybackslash}p{2cm}@{}}
    \toprule
    Parameter (units) & Range \\
    \midrule
    Radius (Earth radii) & $\left[0.7,\,1.75\right]$ \\
    Surface gravity (m\,s$^{-2}$) & $\left[6.0,\,17.0\right]$ \\
    Rotation period (days) & $\left[1,\,200\right]$ \\
    Surface pressure (bar) & $\left[0.5,\,5.0\right]$ \\
    CO$_2$ volume fraction (\%) & $\left[0,\,100\right]$ \\
    CH$_4$ volume fraction (\%) & $\left[0,\,5\right]$ \\
    Incident stellar flux (W\,m$^{-2}$) & $\left[500,\,1500\right]$ \\
    Stellar temperature (K) & $\left[2500,\,5800\right]$ \\
    \bottomrule
  \end{tabular}
  \vspace{-20pt}
\end{wraptable}

\paragraph{Planets.}
We focus on tidally locked waterworlds (Figure~\ref{fig:schema}): ocean-covered rocky planets with one hemisphere permanently facing the host star.
Two facts make this the natural class to study.
First, most detectable habitable-zone planets orbit close to stars dimmer than the Sun, where stronger tidal forces synchronize the planet's rotation with its orbit, locking one side in permanent daylight and the other in permanent night (this is why the Moon always shows the same face to Earth).
Second, many such planets are likely to be ocean-covered, and a global ocean provides a clean idealization that avoids arbitrary choices about continent arrangement.
These planets are the most widely simulated subclass of potentially habitable exoplanets.

\paragraph{Inputs.}
Each planet is characterized by eight continuous inputs: radius, surface gravity, rotation period, surface pressure, CO$_2$ and CH$_4$ volume fractions, incident stellar flux, and stellar temperature.
A discrete GCM label $s\in\{1,\dots,5\}$ identifies the source GCM.
We restrict evaluation to planets satisfying the \emph{core domain} input constraints in Table~\ref{tab:physical-domains}.
Beyond these bounds, physical plausibility and climate regime transitions would make the scientific interpretation of emulator accuracy less clear.
However, we retain some simulations outside the core domain for training; these help anchor the response surface near the domain edges.

\paragraph{Outputs.}
GCMs are run until the atmosphere reaches a statistical steady state, plus some averaging window; the prediction target is the mean over this window, represented as 53 fields on a $32\times64$ latitude--longitude grid.
We use \emph{variable} to refer to a single physical quantity that may span multiple vertical levels, and \emph{field} to refer to a single slice at one level.
The 3D variables are temperature, (specific) humidity, east--west (E--W) wind, north--south (N--S) wind, and cloud fraction, each on 10 pressure levels.
The 2D variables are surface temperature, outgoing longwave radiation (OLR), and absorbed shortwave radiation (ASR).

\subsection{Dataset construction}\label{sec:sources}

\paragraph{GCMs.}
Our data spans five exoplanet GCMs: the UM, ExoCAM, ExoPlaSim, LFRic, and ExoCAM-pre-2022 (Table~\ref{tab:dataset-composition}).
We designate ExoCAM and the UM as \emph{target} GCMs, evaluated at test time, since they are high-fidelity models with relatively plentiful simulations in the core domain.
The remaining three serve as \emph{auxiliary} GCMs, contributing training data only.
ExoPlaSim provides the bulk of the auxiliary data but is lower fidelity than the rest.
ExoCAM's radiative transfer component was significantly updated in 2022, motivating ExoCAM-pre-2022's treatment as a separate auxiliary source. We overview the GCMs used in this work in Table~\ref{tab:gcm-overview} and give further background in Appendix~\ref{app:gcms}.

\paragraph{Data sources.}
The simulations are drawn primarily from the existing literature, where each study typically varies only a few parameters around community-favourite planets (such as TRAPPIST-1e and Proxima Centauri b).
The resulting input-space coverage is highly non-uniform.
To mitigate this, we ran 424 bespoke simulations chosen to fill gaps using a weighted coverage design.
See Appendix~\ref{app:sampling} for the sampling design and Appendix~\ref{app:gcms} for the bespoke simulation configurations.

\begin{table*}[t]
  \centering
  \small
  \setlength{\tabcolsep}{5pt}
  \renewcommand{\arraystretch}{1.2}
  \caption{Dataset composition. ThousandWorlds contains 1,689 simulations. We restrict evaluation to 346 target simulations (bold); the remaining 1,343 are eligible for training only (subject to the same-planet exclusions in Table~\ref{tab:subsets}). These comprise 48 simulations from target GCMs that are outside the core domain (Table~\ref{tab:physical-domains}) and 1,295 simulations from auxiliary GCMs.}
  \label{tab:dataset-composition}
  \begin{tabular}{@{}>{\centering\arraybackslash}m{0.7cm}|>{\raggedright\arraybackslash}p{2cm}cc|p{0.55\textwidth}@{}}
    \toprule
    \multicolumn{1}{@{}c}{} & & \multicolumn{2}{c}{{Physical domain}} & \\
    \cmidrule(lr){3-4}
    \multicolumn{1}{@{}c}{} & GCMs & \shortstack{Core} & Extended & \multicolumn{1}{c@{}}{Data sources} \\
    \midrule
    \smash{\rotatebox[origin=r]{90}{Target}} & UM & \textbf{252} & 33 & {\footnotesize \cite{mak3DSimulations2024, sergeevTRAPPIST1Habitable2022}, this work}\\
      & ExoCAM & \textbf{94} & 15 & {\footnotesize \cite{hammondClimatesThermal2025, haqq-misraSparseAtmospheric2022, sergeevTRAPPIST1Habitable2022,wolfChemistryClimate2025,woodwardExoCAMInPrep}, this work}\\
    \midrule
      & ExoCAM-pre-2022 & 109 & 50 & {\footnotesize \cite{komacekAtmosphericCirculation2019, kopparapuINNEREDGE2016Corrected, kopparapuHabitableMoist2017Corrected, wolfSimulatedPhasedependent2019, wolfAssessingHabitability2017, suissaFirstHabitablezone2020Corrected}} \\
    \smash{\rotatebox[origin=c]{90}{ \textcolor{white}{a.} Auxiliary}} & LFRic & 14 & 5 & {\footnotesize \cite{haqq-misraSparseAtmospheric2022}, this work}\\
      & ExoPlaSim & 439 & 678 & {\footnotesize \cite{macdonaldClimateTransition2025, paradiseClimateDiversity2021, paradiseFundamentalChallenges2022}, this work}\\
    \bottomrule
  \end{tabular}
\end{table*}

\subsection{Dataset challenges}\label{sec:challenges}

\paragraph{Structured missingness.}
Different simulations use different vertical grids, do not all extend to the same minimum pressure, and do not all output the same variables.
After interpolation onto the common 10-level pressure grid, this leaves a structured pattern of missingness: some simulations lack data at the uppermost or lowermost pressure levels, and some lack entire variables.
We visualize the missingness patterns in Figure~\ref{fig:missingness-patterns}.

\paragraph{Inter-GCM disagreement.}
Different GCMs use different dynamical cores, physical parameterizations, and numerical methods.
Two equal-fidelity GCMs given identical planet parameters can produce systematically different climates, reflecting the community's epistemic uncertainty about the underlying atmospheric physics.
Part of the measured inter-GCM disagreement also reflects different modelling choices made by the people running the GCMs: matching planet parameters does not require matching configuration settings, many of which are incompatible across different GCMs.
Where the appropriate choices are unclear, these differences are themselves a component of epistemic uncertainty about the planet's climate.
Achieving positive transfer from shared cross-GCM structure (ideally reflecting the shared underlying physics) while avoiding negative transfer from their systematic differences is one of the benchmark's key modelling challenges.
The shared-planets evaluation protocol (Section~\ref{sec:eval}) measures emulator error against this disagreement directly.

\paragraph{Within-GCM variability.}
Even within a single GCM, simulations from different studies use different configurations (e.g., convection schemes, cloud particle sizes, time-averaging windows, initial conditions).
These choices affect the final climate but are numerous, inconsistently documented, and far too sparsely sampled to include as inputs.
From an emulation perspective, they act as structured noise: output variability that correlates with unobserved simulation settings rather than with the eight planet parameters.

\subsection{Data processing}\label{sec:preprocessing}

\paragraph{Regridding.}
Each GCM produces output on its own native grid.
We regrid all fields onto a $32\times 64$ latitude--longitude grid with 10 pressure levels for 3D variables.
Horizontal interpolation is bilinear; vertical interpolation is in log-pressure.
The 10 pressure levels are isobars defined relative to the input surface pressure. Further details in Appendix~\ref{app:regridding}.

\paragraph{Inputs.}
Rotation period and surface pressure are log-transformed.
Gas volume fractions are transformed via $x\gets\operatorname{asinh}(x/s)$ with a fixed pivot $s$ per species (CO$_2$: $10^{-6}$; CH$_4$: $10^{-8}$), chosen so that the transformation is approximately logarithmic at climatically significant concentrations and linear near zero.

\paragraph{Outputs.}
Humidity is log-transformed.
Cloud fraction is smoothed-logit-transformed: $c\mapsto\operatorname{logit}((c+\varepsilon)/(1+2\varepsilon))$ with $\varepsilon=10^{-2}$; predictions are clamped back to $[0,1]$ after inversion.
ASR and OLR are divided by each example's incident stellar flux.

\paragraph{Equatorial symmetry.}
All planets in the dataset have symmetric forcing about the equator, so time-mean climates should be equatorially symmetric in all fields except N--S wind, which is antisymmetric.
In practice, some simulations exhibit residual asymmetry due to finite time-averaging windows and, in some cases, spontaneous symmetry breaking or amplified numerical asymmetries.
We treat these as artefacts (or at least beyond the scope of emulation) and symmetrize all fields in the released data: for symmetric fields this is equivalent to averaging the two hemispheres, and for the one antisymmetric field (N--S wind) to taking half the hemisphere difference.

\paragraph{Spectral representation.}
Spherical harmonics are a natural orthonormal basis on the sphere, analogous to sinusoids on a circle.
We release the dataset in two formats: gridded numpy arrays on the common $32\times 64$ grid, and spectral coefficients after a T21 spherical harmonic transform (SHT) of each field.
Truncation at degree 21 discards high-frequency spatial modes and yields a more compact representation.
The gridded format is the primary representation and the basis for all evaluation.
The spectral format is a convenience for methods that operate on spectral coefficients.
The benchmark package includes precomputed inverse SHT weights for mapping spectral predictions back to the grid.

\subsection{Benchmark structure}\label{sec:subsets}

We define three nested subsets that progressively expose the challenges described in Section~\ref{sec:challenges} (Table~\ref{tab:subsets}).
\textbf{Single-complete} uses UM simulations only, with 48 fields common to all GCMs -- a pure regression problem with no multi-GCM transfer and no missingness.
\textbf{Multi-complete} adds all five GCMs but retains the same 48 fields, introducing cross-GCM transfer without missingness.
\textbf{Multi-partial} uses all five GCMs and 53 fields (from a wider vertical grid), introducing structured missingness (top and bottom pressure levels and entire-variable absences); this is the full dataset.
In terms of the examples included, each subset nests within the next: Single-complete $\subset$ Multi-complete $\subset$ Multi-partial, with train--test assignments preserved across subsets.
Train--test splitting is based on the 346 target simulations.
All remaining simulations are eligible for training only.
To prevent leakage, planets simulated by multiple GCMs are never split across train and test.

\begin{table*}[htbp]
  \centering
  \small
  \setlength{\tabcolsep}{3pt}
  \renewcommand{\arraystretch}{1.1}
  \caption{Benchmark subsets: their features and split counts.
  Multi-partial is the full dataset; Multi-complete simplifies to complete observations only; Single-complete further simplifies to a single GCM (the UM).
  The shared-planets test sets are the restrictions of the standard test sets to planets simulated by both target GCMs (the UM and ExoCAM).
  To prevent leakage, we exclude simulations from auxiliary GCMs that correspond to identical planets (the same eight physical inputs, up to numerical rounding) present in the test set.}
  \label{tab:subsets}
  \begin{tabular}{@{}>{\raggedright\arraybackslash}p{2.6cm}>{\raggedright\arraybackslash}p{1.65cm}>{\centering\arraybackslash}p{1cm}>{\centering\arraybackslash}p{1.65cm}>{\raggedleft\arraybackslash}p{1.5cm}>{\raggedleft\arraybackslash}p{2.15cm}>{\raggedleft\arraybackslash}p{2.15cm}@{}}
    \toprule
    Subset & GCMs & Fields & Missingness & Train size & Standard test & Shared-planets test \\
    \midrule
    Multi-partial    & all 5 GCMs & 53 & yes & 1527 & 128 & 64 \\
    Multi-complete   & all 5 GCMs & 48 & no  & 1443 & 121 & 62 \\
    Single-complete  & UM only    & 48 & no  & 203   & 74  & --- \\
    \bottomrule
  \end{tabular}
\end{table*}

\subsection{Metrics and evaluation protocols}\label{sec:eval}

\paragraph{Metrics.}
Our primary benchmark metric is the area-weighted RMSE, $\mathrm{RMSE}(\hat{\mathbf{y}},\mathbf{y}) = \|\hat{\mathbf{y}}-\mathbf{y}\|_G$, where $\|\mathbf{e}\|_G = \sqrt{\mathbf{e}^\top\mathbf{G}\mathbf{e}}$ and $\mathbf{G}$ contains latitude weights.
For probabilistic methods, we also use the energy score, a proper scoring rule which can be seen as a multivariate generalization of the continuous ranked probability score (Appendix~\ref{app:probabilistic-results}).
We explore additional metrics (anomaly correlation coefficient, spread--skill ratio, and RAMSE, a variant of RMSE that does not reward smoothing) in Appendix~\ref{app:additional-results}.

\paragraph{Evaluation protocols.}
We evaluate under two protocols.
The \emph{standard} protocol reports metrics as usual on the full test sets.
The \emph{shared-planets} protocol restricts to planets simulated by both target GCMs.
For each simulation in these same-planet pairs, the source GCM is treated as the ground truth and the other GCM as a competing predictor.
Each metric is then reported as the ratio of the emulator's score to the other GCM's score.
For example, the relative RMSE $\mathrm{RMSE}_\text{rel}=\mathrm{RMSE}(\hat{\mathbf{y}},\mathbf{y}^s)/\mathrm{RMSE}(\mathbf{y}^{s'},\mathbf{y}^s),$
where $\mathbf{y}^s$ is the output of target GCM $s$ and $\mathbf{y}^{s'}$ is the output of the other target GCM on the same planet.
Values below 1 mean the emulator's error is smaller than the disagreement between GCMs for the same planet; values above 1 mean it is larger. 
The shared-planets protocol applies only to the Multi- subsets, since it requires planets simulated by both target GCMs.
Metric aggregation follows Appendix~\ref{app:metric-aggregation}.

The standard protocol has larger test sets and is more reliable for ranking methods; the shared-planets protocol puts errors on a scientifically meaningful scale.

\subsection{Baselines}\label{sec:baselines}
We evaluate ten baselines.
Appendix~\ref{app:baselines} gives full descriptions and hyperparameter tuning details.
\textbf{Train-mean} predicts the training-set mean for every test example.
\textbf{kNN} averages the $k$ nearest training examples in input space.
\textbf{PCA-GBT} regresses PCA scores of the spectral coefficients (Section~\ref{sec:preprocessing}) with gradient-boosted trees.
\textbf{Coord-MLP} follows \citet{plaschzugAcceleratingExoplanet2025}: a pointwise MLP maps planet parameters, GCM label, variable identity, and spatial coordinates to a single field value.
\textbf{Coord-DeepONet} uses a DeepONet architecture \citep{luLearningNonlinear2021}: a branch network encodes the inputs (planet parameters and GCM label), while a trunk network encodes the output query coordinates (variable, pressure level, latitude, longitude), and their inner product produces a single output value.
\textbf{PCA-MLP} regresses the PCA scores of the spectral coefficients with an MLP (our PCA-MLP baseline is also essentially equivalent to what gets called a ``POD-DeepONet'' in the operator-learning literature).
\textbf{ConvDec} is a convolutional decoder similar to the DCGAN generator of \citet{radfordUnsupervisedRepresentation2016}, here trained as a deterministic regressor.
It maps inputs to a coarse feature grid, then progressively upsamples and refines it with shared local filters and circular padding in longitude.
\textbf{SFNO} adapts the Spherical Fourier Neural Operator \citep{bonevSphericalFourier2023}, which learns global convolutions as spectral filters in the spherical harmonic domain. SFNO is designed for field-to-field mappings; to adapt it for parameter-to-field regression, we broadcast the inputs as spatially constant channels and add a learned position embedding, which supplies the spatially varying component.
\textbf{PPCA-ICM} compresses outputs with probabilistic PCA and regresses the scores with a multi-task GP over GCM labels (ICM; \citealp{alvarezKernelsVectorValued2012}).
\textbf{GPLFR} (Gaussian process latent factor regression; \citealp{stevensonGaussianProcessLatent2026}) replaces PPCA-ICM's compress-then-predict pipeline with end-to-end optimization.
Both GP methods produce probabilistic predictions.

We also evaluate two physically motivated extensions of GPLFR but find their performance effects to be small (Appendix~\ref{app:gplfr-extensions}).

\section{Experiments}\label{sec:experiments}

We evaluate ten baselines (Section~\ref{sec:baselines}) on the three dataset subsets (Section~\ref{sec:subsets}) under both evaluation protocols (Section~\ref{sec:eval}).
Section~\ref{sec:baseline-comparison} compares baselines using standard RMSE; Section~\ref{sec:scientific-utility} uses the shared-planets protocol to assess scientific utility.
Probabilistic evaluation and additional metrics are reported in Appendix~\ref{app:additional-results}.
Example predictions and climate diagnostics are in Appendix~\ref{app:example-predictions}.

\subsection{Baseline comparison}\label{sec:baseline-comparison}

\begin{table}[t]
  % \vspace{-2pt}
  \centering
  \small
  \setlength{\tabcolsep}{1.5pt}
  \renewcommand{\arraystretch}{1.0}
  \caption{RMSE by subset, variable, and method. Darker colours indicate lower (better) scores, with bold the lowest. The geometric mean is reported relative to Train-mean's geometric mean. Scores are the mean of five training seeds (one deterministic run for Train-mean and kNN); Table~\ref{tab:rmse-partial-splits-std} includes the standard deviations. E--W: east--west; N--S: north--south; ASR: absorbed shortwave radiation; OLR: outgoing longwave radiation. Humidity denotes specific humidity.}
  \label{tab:rmse-partial-splits}
  \resizebox{\linewidth}{!}{%
  \begin{tabular}{@{}>{\centering\arraybackslash}m{0.65cm}|>{\raggedright\arraybackslash}p{2.38cm} >{\raggedleft\arraybackslash}p{1.0cm} >{\raggedleft\arraybackslash}p{1.05cm} *{8}{>{\raggedleft\arraybackslash}p{1.32cm}}@{}}
    \toprule
    \shortstack{Sub-\\set} & Variable & Train-mean & kNN & PCA-GBT & Coord-MLP & Coord-DeepONet & PCA-MLP & ConvDec & SFNO & PPCA-ICM & GPLFR \\
    \midrule
     & Surface temp. (K)                      & \rmseMM{10}{24.52} & \rmseMM{19}{18.79} & \rmseMM{54}{12.33} & \rmseMM{28}{17.65} & \rmseMM{46}{12.45} & \rmseMM{63}{12.24} & \rmseMM{37}{12.75} & \rmseMM{72}{12.24} & \rmseMM{81}{10.06} & \rmseMM{90}{\textbf{9.89}} \\
     & Temperature (K)                        & \rmseMM{10}{20.34} & \rmseMM{19}{14.66} & \rmseMM{54}{10.23} & \rmseMM{28}{11.94} & \rmseMM{46}{10.43} & \rmseMM{72}{10.08} & \rmseMM{63}{10.10} & \rmseMM{37}{10.52} & \rmseMM{81}{8.79} & \rmseMM{90}{\textbf{8.26}} \\
     & Humidity (dex)                         & \rmseMM{10}{1.119} & \rmseMM{19}{0.784} & \rmseMM{37}{0.536} & \rmseMM{28}{0.627} & \rmseMM{54}{0.527} & \rmseMM{72}{0.5251} & \rmseMM{63}{0.5253} & \rmseMM{46}{0.531} & \rmseMM{81}{0.446} & \rmseMM{90}{\textbf{0.432}} \\
    \smash{\rotatebox[origin=c]{90}{Multi-partial \textcolor{white}{aa}}}
     & Cloud fraction (1)                     & \rmseMM{10}{0.1277} & \rmseMM{28}{0.0998} & \rmseMM{54}{0.0879} & \rmseMM{19}{0.1073} & \rmseMM{46}{0.0897} & \rmseMM{37}{0.0898} & \rmseMM{81}{0.0816} & \rmseMM{63}{0.0845} & \rmseMM{72}{0.0836} & \rmseMM{90}{\textbf{0.0778}} \\
     & E--W wind (m\,s$^{-1}$)                & \rmseMM{10}{16.73} & \rmseMM{37}{10.92} & \rmseMM{54}{10.64} & \rmseMM{19}{12.12} & \rmseMM{46}{10.88} & \rmseMM{28}{11.19} & \rmseMM{72}{9.97} & \rmseMM{63}{10.46} & \rmseMM{81}{9.63} & \rmseMM{90}{\textbf{9.50}} \\
     & N--S wind (m\,s$^{-1}$)                & \rmseMM{10}{6.57} & \rmseMM{46}{4.72} & \rmseMM{37}{4.75} & \rmseMM{19}{5.13} & \rmseMM{28}{4.91} & \rmseMM{54}{4.67} & \rmseMM{90}{\textbf{4.08}} & \rmseMM{72}{4.18} & \rmseMM{63}{4.34} & \rmseMM{81}{4.12} \\
     & ASR (W\,m$^{-2}$)                      & \rmseMM{19}{59.0} & \rmseMM{46}{33.2} & \rmseMM{63}{27.1} & \rmseMM{10}{77.1} & \rmseMM{54}{29.8} & \rmseMM{72}{26.2} & \rmseMM{28}{37.7} & \rmseMM{37}{33.5} & \rmseMM{81}{26.0} & \rmseMM{90}{\textbf{22.8}} \\
     & OLR (W\,m$^{-2}$)                      & \rmseMM{10}{41.0} & \rmseMM{28}{24.8} & \rmseMM{37}{19.5} & \rmseMM{19}{29.0} & \rmseMM{54}{18.9} & \rmseMM{63}{18.7} & \rmseMM{81}{18.0} & \rmseMM{46}{19.0} & \rmseMM{72}{18.2} & \rmseMM{90}{\textbf{16.1}} \\
    \arrayrulecolor{gray!50}\cmidrule(l){2-12}\arrayrulecolor{black}
     & Geometric mean                         & \rmseMM{10}{1.00} & \rmseMM{28}{0.685} & \rmseMM{54}{0.550} & \rmseMM{19}{0.754} & \rmseMM{37}{0.560} & \rmseMM{72}{0.546} & \rmseMM{63}{0.547} & \rmseMM{46}{0.551} & \rmseMM{81}{0.492} & \rmseMM{90}{\textbf{0.462}} \\
    \midrule
     & Surface temp. (K)                      & \rmseTR{10}{24.00} & \rmseTR{19}{19.58} & \rmseTR{46}{12.59} & \rmseTR{28}{16.66} & \rmseTR{72}{12.07} & \rmseTR{54}{12.36} & \rmseTR{37}{12.75} & \rmseTR{63}{12.29} & \rmseTR{81}{11.20} & \rmseTR{90}{\textbf{9.79}} \\
     & Temperature (K)                        & \rmseTR{10}{19.32} & \rmseTR{19}{15.22} & \rmseTR{54}{9.90} & \rmseTR{28}{11.44} & \rmseTR{63}{9.68} & \rmseTR{81}{9.47} & \rmseTR{46}{9.97} & \rmseTR{37}{10.40} & \rmseTR{72}{9.59} & \rmseTR{90}{\textbf{8.14}} \\
     & Humidity (dex)                         & \rmseTR{10}{1.087} & \rmseTR{19}{0.765} & \rmseTR{54}{0.506} & \rmseTR{28}{0.62} & \rmseTR{72}{0.49} & \rmseTR{63}{0.506} & \rmseTR{46}{0.512} & \rmseTR{37}{0.513} & \rmseTR{81}{0.459} & \rmseTR{90}{\textbf{0.404}} \\
    \smash{\rotatebox[origin=c]{90}{Multi-complete \textcolor{white}{aa}}}
     & Cloud fraction (1)                     & \rmseTR{10}{0.1356} & \rmseTR{28}{0.1074} & \rmseTR{54}{0.0913} & \rmseTR{19}{0.112} & \rmseTR{37}{0.0968} & \rmseTR{46}{0.0936} & \rmseTR{81}{0.0865} & \rmseTR{63}{0.089} & \rmseTR{72}{0.0878} & \rmseTR{90}{\textbf{0.0816}} \\
     & E--W wind (m\,s$^{-1}$)                & \rmseTR{10}{15.25} & \rmseTR{19}{10.16} & \rmseTR{63}{8.92} & \rmseTR{28}{10.14} & \rmseTR{37}{9.63} & \rmseTR{46}{9.57} & \rmseTR{72}{8.64} & \rmseTR{54}{9.13} & \rmseTR{81}{8.63} & \rmseTR{90}{\textbf{8.24}} \\
     & N--S wind (m\,s$^{-1}$)                & \rmseTR{10}{6.15} & \rmseTR{19}{4.90} & \rmseTR{54}{4.43} & \rmseTR{28}{4.89} & \rmseTR{37}{4.87} & \rmseTR{46}{4.66} & \rmseTR{90}{\textbf{4.00}} & \rmseTR{72}{4.12} & \rmseTR{63}{4.22} & \rmseTR{81}{4.07} \\
     & ASR (W\,m$^{-2}$)                      & \rmseTR{19}{59.7} & \rmseTR{46}{31.5} & \rmseTR{63}{26.5} & \rmseTR{10}{68.6} & \rmseTR{54}{29.6} & \rmseTR{81}{25.5} & \rmseTR{28}{36.5} & \rmseTR{37}{31.7} & \rmseTR{72}{26.1} & \rmseTR{90}{\textbf{22.1}} \\
     & OLR (W\,m$^{-2}$)                      & \rmseTR{10}{41.2} & \rmseTR{28}{23.8} & \rmseTR{37}{19.0} & \rmseTR{19}{26.2} & \rmseTR{63}{18.2} & \rmseTR{46}{18.5} & \rmseTR{81}{17.5} & \rmseTR{54}{18.3} & \rmseTR{72}{18.2} & \rmseTR{90}{\textbf{15.5}} \\
    \arrayrulecolor{gray!50}\cmidrule(l){2-12}\arrayrulecolor{black}
     & Geometric mean                         & \rmseTR{10}{1.00} & \rmseTR{28}{0.700} & \rmseTR{72}{0.540} & \rmseTR{19}{0.723} & \rmseTR{37}{0.554} & \rmseTR{63}{0.542} & \rmseTR{54}{0.546} & \rmseTR{46}{0.548} & \rmseTR{81}{0.513} & \rmseTR{90}{\textbf{0.458}} \\
    \midrule
     & Surface temp. (K)                      & \rmseSG{10}{20.85} & \rmseSG{28}{14.04} & \rmseSG{46}{11.63} & \rmseSG{19}{14.76} & \rmseSG{37}{11.94} & \rmseSG{54}{11.40} & \rmseSG{63}{10.72} & \rmseSG{72}{10.04} & \rmseSG{81}{9.65} & \rmseSG{90}{\textbf{9.38}} \\
     & Temperature (K)                        & \rmseSG{10}{18.71} & \rmseSG{19}{11.71} & \rmseSG{28}{9.88} & \rmseSG{37}{9.85} & \rmseSG{46}{9.75} & \rmseSG{54}{9.72} & \rmseSG{63}{8.64} & \rmseSG{72}{8.55} & \rmseSG{81}{8.33} & \rmseSG{90}{\textbf{8.08}} \\
     & Humidity (dex)                         & \rmseSG{10}{1.067} & \rmseSG{19}{0.583} & \rmseSG{37}{0.555} & \rmseSG{28}{0.582} & \rmseSG{46}{0.549} & \rmseSG{54}{0.523} & \rmseSG{63}{0.477} & \rmseSG{72}{0.472} & \rmseSG{81}{0.455} & \rmseSG{90}{\textbf{0.447}} \\
    \smash{\rotatebox[origin=c]{90}{Single-complete \textcolor{white}{aa}}}
     & Cloud fraction (1)                     & \rmseSG{10}{0.1287} & \rmseSG{28}{0.0902} & \rmseSG{54}{0.0866} & \rmseSG{19}{0.1021} & \rmseSG{46}{0.0896} & \rmseSG{37}{0.0900} & \rmseSG{90}{\textbf{0.0761}} & \rmseSG{72}{0.0809} & \rmseSG{63}{0.0816} & \rmseSG{81}{0.0767} \\
     & E--W wind (m\,s$^{-1}$)                & \rmseSG{10}{12.75} & \rmseSG{54}{8.53} & \rmseSG{46}{8.54} & \rmseSG{37}{9.01} & \rmseSG{19}{9.55} & \rmseSG{28}{9.39} & \rmseSG{90}{\textbf{7.34}} & \rmseSG{63}{8.16} & \rmseSG{72}{8.14} & \rmseSG{81}{7.49} \\
     & N--S wind (m\,s$^{-1}$)                & \rmseSG{10}{5.16} & \rmseSG{63}{3.82} & \rmseSG{46}{3.95} & \rmseSG{37}{4.20} & \rmseSG{19}{4.23} & \rmseSG{28}{4.22} & \rmseSG{90}{\textbf{3.36}} & \rmseSG{72}{3.59} & \rmseSG{54}{3.95} & \rmseSG{81}{3.55} \\
     & ASR (W\,m$^{-2}$)                      & \rmseSG{19}{45.6} & \rmseSG{54}{28.6} & \rmseSG{63}{28.0} & \rmseSG{10}{67.0} & \rmseSG{46}{30.5} & \rmseSG{72}{26.0} & \rmseSG{28}{36.3} & \rmseSG{37}{32.1} & \rmseSG{90}{\textbf{24.4}} & \rmseSG{81}{25.1} \\
     & OLR (W\,m$^{-2}$)                      & \rmseSG{10}{33.7} & \rmseSG{28}{20.4} & \rmseSG{37}{20.0} & \rmseSG{19}{26.3} & \rmseSG{54}{19.0} & \rmseSG{46}{19.0} & \rmseSG{63}{17.7} & \rmseSG{72}{17.1} & \rmseSG{90}{\textbf{16.2}} & \rmseSG{81}{16.3} \\
    \arrayrulecolor{gray!50}\cmidrule(l){2-12}\arrayrulecolor{black}
     & Geometric mean                         & \rmseSG{10}{1.00} & \rmseSG{28}{0.646} & \rmseSG{46}{0.610} & \rmseSG{19}{0.756} & \rmseSG{37}{0.630} & \rmseSG{54}{0.609} & \rmseSG{63}{0.561} & \rmseSG{72}{0.560} & \rmseSG{81}{0.538} & \rmseSG{90}{\textbf{0.518}} \\
    \bottomrule
  \end{tabular}
  }
\end{table}

\paragraph{Results overview.}
Table~\ref{tab:rmse-partial-splits} reports per-variable RMSE for the ten baselines across the three subsets.
The two GP methods are the strongest baselines: GPLFR achieves the lowest RMSE on seven of the eight variables on the Multi- subsets, and three of eight on Single-complete, and PPCA-ICM is the second-best method on every subset by geometric-mean RMSE.
Among the deep learning baselines, PCA-MLP, ConvDec, and SFNO are closely matched: PCA-MLP is narrowly the strongest on the Multi- subsets (ConvDec within 1\%) and SFNO is the strongest on Single-complete.
Coord-DeepONet sits a step behind these three. Coord-MLP is further behind, being the weakest deep learning baseline and overall weaker than kNN.
The trees baseline PCA-GBT is competitive with the best deep learning methods across the subsets.

Comparing to the simple baselines (Train-mean, kNN), the learned methods improve most clearly on temperature and humidity, and least on cloud fraction, winds, and ASR.
We briefly discuss the physical characteristics of these variables and how they may explain these results in Appendix~\ref{app:knn-discussion}.

Replacing PCA-MLP's MLP with ridge regression increases seed-0 RMSE by 25\% on average, affirming the value of nonlinear regression (Appendix~\ref{app:nonlinearity-ablation}).
Appendix~\ref{app:rmse-bootstrap-intervals} reports paired bootstrap intervals for the Multi-partial RMSE results to quantify uncertainty from the test-set draw, together with a paired test of the method ordering.
These intervals are notably wider than the training-seed standard deviations in Appendix~\ref{app:tables-with-seed-variability}, so test-set composition is the dominant source of uncertainty in the RMSE numbers.

\paragraph{Learning compression and regression separately or jointly.}
Two pairs of baselines can be viewed as differing primarily in whether they learn their output basis (compression) and input-to-latent map (regression) separately or jointly: PCA-MLP versus Coord-DeepONet, and PPCA-ICM versus GPLFR (see \citealp{luComprehensiveFair2022} and \citealp{stevensonGaussianProcessLatent2026} for these perspectives).
The \emph{separate} (two-stage) approach is simpler and more stable but can waste capacity on high-variance output structure that is irrelevant or hard to predict.
The \emph{joint} approach can better allocate capacity to predictable structure but pays for this flexibility with weaker identifiability, which can manifest as harder optimization or lower data efficiency.
Our results appear consistent with this interpretation: both joint methods improve relative to their two-stage counterpart when moving from Single-complete to the larger Multi- subsets.
However, which approach wins overall differs between the two families: in the deep learning pair, the two-stage method (PCA-MLP) leads on all three subsets; in the GP pair, the joint method (GPLFR) does.
This is plausibly due to the stronger inductive biases of the GP methods allowing the joint learning advantages to win out even at small dataset sizes.

% \vspace{-2pt}
\paragraph{Dataset composition ablations.}
The effect of auxiliary-GCM data is method-dependent: it slightly benefits PCA-GBT, leaves GPLFR essentially unchanged, slightly harms PCA-MLP and PPCA-ICM and clearly harms the remaining four deep learning methods (Coord-MLP, Coord-DeepONet, ConvDec, SFNO). 
Better ways to leverage multi-simulator data are therefore a promising direction for improving on these baselines.
Data from outside the core domain is beneficial for all learned methods except PCA-GBT, for which it is essentially neutral.
See Appendix~\ref{app:dataset-ablations} for further discussion.

\subsection{Scientific utility of emulators}\label{sec:scientific-utility}

\begin{table}[htbp]
  \centering
  \small
  \setlength{\tabcolsep}{1.5pt}
  \renewcommand{\arraystretch}{1.0}
  \caption{Relative RMSE by variable and method under the shared-planets protocol on the Multi-partial subset, with the Multi-complete geometric mean in the final row. Darker colours indicate lower (better) scores and bold the lowest in each row. Scores are the mean of five training seeds; Table~\ref{tab:relative-rmse-std} gives per-variable results for both subsets, with the standard deviations.}
  \label{tab:relative-rmse}
  \resizebox{\linewidth}{!}{%
  \begin{tabular}{@{}>{\raggedright\arraybackslash}m{2.5cm} >{\raggedleft\arraybackslash}m{1.0cm} >{\raggedleft\arraybackslash}m{1.05cm} *{8}{>{\raggedleft\arraybackslash}m{1.32cm}}@{}}
    \toprule
    Variable & Train-mean & kNN & PCA-GBT & Coord-MLP & Coord-DeepONet & PCA-MLP & ConvDec & SFNO & PPCA-ICM & GPLFR \\
    \midrule
    Surface temp.                    & \rmseMM{10}{1.680} & \rmseMM{28}{1.283} & \rmseMM{46}{1.039} & \rmseMM{19}{1.380} & \rmseMM{63}{1.005} & \rmseMM{37}{1.057} & \rmseMM{54}{1.007} & \rmseMM{72}{0.975} & \rmseMM{81}{0.927} & \rmseMM{90}{\textbf{0.739}} \\
    Temperature                        & \rmseMM{10}{1.655} & \rmseMM{19}{1.209} & \rmseMM{46}{0.925} & \rmseMM{28}{0.998} & \rmseMM{37}{0.949} & \rmseMM{72}{0.902} & \rmseMM{63}{0.903} & \rmseMM{54}{0.921} & \rmseMM{81}{0.829} & \rmseMM{90}{\textbf{0.664}} \\
    Humidity                         & \rmseMM{10}{1.515} & \rmseMM{19}{1.247} & \rmseMM{37}{0.896} & \rmseMM{28}{1.008} & \rmseMM{46}{0.879} & \rmseMM{72}{0.818} & \rmseMM{54}{0.851} & \rmseMM{63}{0.824} & \rmseMM{81}{0.787} & \rmseMM{90}{\textbf{0.670}} \\
    Cloud fraction                     & \rmseMM{10}{0.867} & \rmseMM{19}{0.752} & \rmseMM{46}{0.644} & \rmseMM{28}{0.733} & \rmseMM{37}{0.653} & \rmseMM{54}{0.639} & \rmseMM{81}{0.595} & \rmseMM{72}{0.606} & \rmseMM{63}{0.624} & \rmseMM{90}{\textbf{0.562}} \\
    E--W wind                & \rmseMM{10}{1.265} & \rmseMM{63}{0.832} & \rmseMM{46}{0.897} & \rmseMM{19}{0.937} & \rmseMM{28}{0.918} & \rmseMM{37}{0.909} & \rmseMM{90}{\textbf{0.757}} & \rmseMM{54}{0.833} & \rmseMM{72}{0.785} & \rmseMM{81}{0.765} \\
    N--S wind                & \rmseMM{10}{1.205} & \rmseMM{54}{0.828} & \rmseMM{37}{0.885} & \rmseMM{28}{0.898} & \rmseMM{19}{0.919} & \rmseMM{46}{0.837} & \rmseMM{90}{\textbf{0.714}} & \rmseMM{81}{0.746} & \rmseMM{63}{0.802} & \rmseMM{72}{0.749} \\
    ASR                      & \rmseMM{19}{1.342} & \rmseMM{46}{0.884} & \rmseMM{81}{0.699} & \rmseMM{10}{2.007} & \rmseMM{54}{0.794} & \rmseMM{72}{0.699} & \rmseMM{28}{1.044} & \rmseMM{37}{0.920} & \rmseMM{63}{0.711} & \rmseMM{90}{\textbf{0.594}} \\
    OLR                      & \rmseMM{10}{1.571} & \rmseMM{28}{1.023} & \rmseMM{54}{0.775} & \rmseMM{19}{1.218} & \rmseMM{46}{0.776} & \rmseMM{63}{0.767} & \rmseMM{81}{0.741} & \rmseMM{72}{0.765} & \rmseMM{37}{0.776} & \rmseMM{90}{\textbf{0.640}} \\
    \arrayrulecolor{gray!50}\cmidrule{1-11}\arrayrulecolor{black}
    Geometric mean                       & \rmseMM{10}{1.360} & \rmseMM{28}{0.988} & \rmseMM{46}{0.836} & \rmseMM{19}{1.097} & \rmseMM{37}{0.854} & \rmseMM{54}{0.819} & \rmseMM{72}{0.814} & \rmseMM{63}{0.816} & \rmseMM{81}{0.776} & \rmseMM{90}{\textbf{0.669}} \\
    \midrule
    Multi-complete\newline geometric mean & \rmseTR{10}{1.365} & \rmseTR{28}{1.040} & \rmseTR{81}{0.809} & \rmseTR{19}{1.061} & \rmseTR{37}{0.862} & \rmseTR{72}{0.816} & \rmseTR{46}{0.827} & \rmseTR{54}{0.823} & \rmseTR{63}{0.822} & \rmseTR{90}{\textbf{0.688}} \\
    \bottomrule
  \end{tabular}
  }
\end{table}

Table~\ref{tab:relative-rmse} reports relative RMSE under the shared-planets protocol, which normalizes emulator error by inter-GCM disagreement for the same planet.
Most learned baselines achieve relative RMSE below 1 on most variables, with GPLFR, PPCA-ICM, ConvDec, SFNO, PCA-MLP, Coord-DeepONet, and PCA-GBT achieving it on nearly every variable.
So these emulators already reach accuracies within the spread of frontier GCMs on these planets, while kNN sits right around the GCM-spread threshold.

In Section~\ref{sec:baseline-comparison} we saw that, for RMSE, the variables that complex methods struggled most to improve on over kNN were cloud fraction, winds, and ASR.
Here we see that kNN already sits below the GCM disagreement on each of these.
This leaves learned methods less headroom on the GCM-relative scale, for some combination of two reasons: kNN captures most of the predictable variation, or the inter-GCM disagreement itself is large (see Appendix~\ref{app:knn-discussion}).
Cloud fraction is the clearest case of the latter: even Train-mean achieves relative RMSE below 1.
This indicates that, for these planets, cloud fraction is dominated by inter-GCM variability rather than inter-planet variability.\footnote{This is consistent with the long-standing difficulty of modelling clouds. For Earth (where we have better ground truth access), clouds are usually cited as the largest source of uncertainty for climate change predictions \citep{sherwoodAssessmentEarths2020}.}

Beyond RMSE, both probabilistic methods show a notable drop in relative energy score compared to relative RMSE, supporting the value of their predictive uncertainties (the denominator for both scores is the same, since the energy score of the other GCM's point prediction reduces to its RMSE). Full results are in Appendix~\ref{app:relative-energy-score-results}.

For tasks where predictive accuracy matters more than physical consistency or interpretability -- such as identifying promising observing targets or surveying habitability trends across parameter space -- emulators operating within the GCM spread are likely already scientifically useful.
That said, significant headroom remains: for example, our best method's surface temperature RMSE (9.9\,K) is large enough to affect habitability assessments, and many individual predictions do fall outside the GCM spread (Figure~\ref{fig:per-example-relative-rmse}).

\section{Conclusion}\label{sec:conclusions}

We have introduced ThousandWorlds, a benchmark for emulating exoplanet climates comprising approximately 1,700 simulations from five GCMs.
We have defined three dataset subsets of increasing difficulty and realism, and two evaluation protocols: one for comparing methods and one for measuring scientific utility. For limitations, see Appendix~\ref{app:limitations}.

On method comparison, our baseline results show that Gaussian-process-based methods outperform standard deep learning, making ThousandWorlds a useful challenge problem for novel deep learning surrogate methods in the low-data, parameter-to-field, and multi-simulator regime.
GPLFR is the current best, with PPCA-ICM second on every subset; PCA-GBT and the deep learning methods PCA-MLP, ConvDec, and SFNO form a closely spaced group below them and are the natural targets for new methods.

On scientific utility, the best baselines emulate individual GCM outputs more closely than GCMs agree with each other -- a threshold of practical relevance for exoplanet astronomers -- while leaving substantial headroom in absolute error terms.

We hope ThousandWorlds serves as both a useful benchmark problem for an underserved regime of scientific ML problems, and an invitation to the ML community to contribute their ideas and techniques to accelerating a field that aims to answer one of humanity's oldest questions.

\section*{Data and code availability}
The dataset is available in full, in ready-to-use gridded and spectral formats: \url{https://doi.org/10.57967/hf/8695} (we use v2.0.0, the current version). All baselines, evaluation code, and result tables are available in the benchmark code: \url{https://github.com/astroautomata/ThousandWorlds}.
The dataset is released under the Creative Commons Attribution 4.0 (CC BY 4.0) license and hosted on Hugging Face; the benchmark code is released under the MIT license on GitHub.
The Appendix documents the steps needed to reproduce the results independently.

\acksection

We are grateful to Hamza Ali Shahjahan for contributing the PCA-GBT baseline implementation.
Edward Stevenson is supported by the Science and Technology Facilities Council (STFC) Centre for Doctoral Training in Data Intensive Science at the University of Cambridge.
Miles Cranmer is grateful for support from the Isaac Newton Trust and the AI2050 program at Schmidt Sciences.
Mei Ting Mak acknowledges support from the Croucher Postdoctoral Fellowship, funded by the Croucher Foundation.
The GCM results are produced using Met Office Software and the Monsoon3 system, a collaborative facility supplied under the Joint Weather and Climate Research Programme, a strategic partnership between the Met Office and the Natural Environment Research Council in the UK.
Eric Wolf acknowledges funding from the Consortium on Habitability and Atmospheres of M-dwarf Planets team and the Virtual Planetary Laboratory, supported by NASA grant numbers 80NSSC21K0905, 80NSSC23K1399, 80NSSC23K1398 and 80NSSC18K0829 respectively.
Tobi Hammond was supported by a NASA FINESST Award (80NSSC25K0320).
Nathan Mayne acknowledges support from a UK Research and Innovation (UKRI) Future Leaders Fellowship MR/T040866/1, and partly from the Leverhulme Trust through a research project grant RPG-2020-82 alongside a Science and Technology Facilities Council (STFC) Consolidated Grant ST/R000395/1.
This work used the Dawn AI service, part of the UK AI Research Resource (AIRR), operated by the University of Cambridge Research Computing Service (\url{www.hpc.cam.ac.uk/d-w-n}) and supported by UK Research and Innovation, with Intel and Dell Technologies as technology partners.

\bibliography{My_Library,manual}
\bibliographystyle{plainnat}
\clearpage

\appendix

\section{Benchmark details}\label{app:benchmark-details}

\subsection{Sampling design}\label{app:sampling}

The bespoke simulations in Table~\ref{tab:dataset-composition} were selected using a weighted coverage design constructed by greedy sequential selection from a Sobol candidate pool.
The design balances two competing goals: filling gaps in the eight-dimensional input space left by the heterogeneous literature data, and concentrating samples near the manifold of physically plausible planets.
Pursuing coverage alone pushes towards combinations such as low radius with high surface gravity, producing unrealistically dense planets that often cause the GCMs to crash.
The weighting function described below mitigates this by downweighting implausible regions.

We standardize the input space to $[0,1]^8$ by recasting surface gravity as density $\rho$; taking logarithms of radius, rotation period, surface pressure, CO$_2$ volume fraction, and CH$_4$ volume fraction (using $\log(x+10^{-16})$ for the gas fractions); and rescaling the transformed coordinates to the unit cube.
We then draw a scrambled Sobol candidate set $U$ and an independent scrambled Sobol probe set $U'$, both in $[0,1]^8$.
Given an existing design $E \subset [0,1]^8$ (the literature simulations, already standardized), we evaluate the weighted coverage objective
\begin{equation*}
  C_p(E \cup \{\mathbf{x}\}) = \int_{[0,1]^8} w(\mathbf{u})\, r(\mathbf{u}; E \cup \{\mathbf{x}\})^p \, d\mathbf{u},
\end{equation*}
where $r(\mathbf{u}; E) = \min_{\mathbf{y} \in E} \|\mathbf{u} - \mathbf{y}\|_2$ is the distance from $\mathbf{u}$ to its nearest design point and $w(\mathbf{u})$ is the weighting function defined below.
The integral is approximated by quadrature over the probe set $U'$.
We use exponent $p=2$, which biases selection towards shrinking large gaps without the brittleness of the minimax criterion ($p=\infty$).
For a later batch of simulations we used $p=4$ to more aggressively target remaining holes once a stable backbone of $p=2$ samples was in place.

New points are selected greedily: at step $t$, we choose
\begin{equation*}
  \mathbf{x}_t = \mathop{\arg\min}_{\mathbf{x} \in U} C_p(E \cup \{\mathbf{x}_1, \ldots, \mathbf{x}_{t-1}, \mathbf{x}\}).
\end{equation*}

The weighting function biases selection towards physically likely densities $\rho$ given radius $R$, and physically likely rotation periods $P_\mathrm{rot}$ given stellar temperature $T_*$ and incident flux $F$, leaving the remaining parameters unweighted.
We set $w(\mathbf{x}) \propto w(\rho \mid R)w(P_\mathrm{rot} \mid T_*, F)$, where: 
\begin{itemize}
  \item $w(\rho \mid R) = \mathrm{Lognormal}(\rho;\mu_\rho, 0.15^2)$, with the mean set by the empirical radius--density relation for rocky planets from \citet{mullerMassradiusRelation2024}: $\mu_\rho / \mathrm{g\,cm}^{-3} = 5.11 \times (R/R_\oplus)^{0.73}$.
  \item $w(P_\mathrm{rot} \mid T_*, F) = \mathrm{Lognormal}(P_\mathrm{rot}; \mu_{P_\mathrm{rot}}, 0.3^2)$.
  The mean $\mu_{P_\mathrm{rot}}$ is computed from Kepler's third law and the Stefan--Boltzmann law using a three-piece function that chains empirical stellar relations to obtain stellar mass $M_*$ from $T_*$: for $T_* \in [2500, 3300]\,\mathrm{K}$, a radius--temperature relation from \citet{cassisiEffectiveTemperature2019}, Stefan--Boltzmann to obtain luminosity $L_*$, and the mass--luminosity relation of \citet{duricAdvancedAstrophysics2004}; for $T_* \in [3300, 4800]\,\mathrm{K}$, the mass--temperature and luminosity--temperature relations of \citet{mannSPECTROTHERMOMETRYDWARFS2013}; and for $T_* \in [4800, 5800]\,\mathrm{K}$, the mass--temperature and mass--luminosity relations of \citet{moyaEmpiricalRelations2018}.  
\end{itemize}

For ExoPlaSim, we further restrict the sampling domain to CO$_2$ partial pressures $P_0 x_{\mathrm{CO}_2} \leq 0.1\,\mathrm{bar}$ and $x_{\mathrm{CH}_4}=0$, since ExoPlaSim is known to be unreliable outside this region \citep{paradiseExoPlaSimExtending2022}.
A small number of candidate points were discarded because the resulting simulations entered runaway greenhouse or CO$_2$ atmospheric collapse.

\subsection{GCM simulations}\label{app:gcms}

Below we describe the five GCMs used in ThousandWorlds (summarized in Table~\ref{tab:gcm-overview}) alongside the configurations used for our bespoke simulations. 
For simulations drawn from the literature, configurations are documented in the source publications listed in Table~\ref{tab:dataset-composition}, and the mapping from individual simulations to source publications is provided in the dataset repository. 

\begin{table}[b]
  \centering
  \small
  \setlength{\tabcolsep}{3pt}
  \renewcommand{\arraystretch}{1.15}
  \caption{Overview of the five GCMs in ThousandWorlds.}
  \label{tab:gcm-overview}
  \begin{tabular}{@{}>{\raggedright\arraybackslash}p{2.15cm} >{\raggedright\arraybackslash}p{2.95cm} >{\raggedright\arraybackslash}p{4.95cm} >{\raggedright\arraybackslash}p{2.65cm}@{}}
    \toprule
    GCM & Lineage & Key notes & Key reference(s) \\
    \midrule
    ExoCAM & NCAR CESM/CAM adapted for exoplanets & Finite volume dynamical core; CAM4 cloud and convection physics; updated radiative transfer & \citep{wolfExoCAM3D2022} \\
    ExoCAM-pre-2022 & Earlier versions of ExoCAM & Evolving radiative transfer schemes; known CO$_2$-atmosphere bias & \citep{yang2016differences, kopparapuHabitableMoist2017Corrected, wolfSimulatedPhasedependent2019} \\
    The UM & The UK Met Office's Unified Model & Lat-lon grid; LLCS cloud scheme & \citep{boutleExploringClimate2017} \\
    LFRic & The UM's successor & Cubed-sphere grid & \citep{sergeevSimulationsIdealised2023} \\
    ExoPlaSim & PlaSim adapted for exoplanets & Simplified parameterizations; spectral dynamical core prone to Gibbs ringing & \citep{paradiseExoPlaSimExtending2022} \\
    \bottomrule
  \end{tabular}
\end{table}

\paragraph{ExoCAM.}
ExoCAM is a planetary climate model based on National Center for Atmospheric Research (NCAR) Community Earth System Model (CESM) version 1.2.1, which expands the native capabilities to allow simulation of a broad parameter space of geophysical properties and atmospheric compositions.  ExoCAM operates as a patch for CESM.  The user must first download the core CESM code, which is freely available provided by NCAR\footnote{https://www.cesm.ucar.edu/models/cesm1.2/tags/}, before modifying CESM with ExoCAM specific configuration scripts, source code, and namelists, also freely available on GitHub\footnote{https://github.com/storyofthewolf/ExoCAM}.  ExoCAM is accompanied by a radiative transfer code ExoRT\footnote{https://github.com/storyofthewolf/ExoRT}, designed for flexibility across a broad range of atmospheric compositions which is linked at build time.  \citet{wolfExoCAM3D2022} describes ExoCAM, ExoRT, and their relationship to CESM1.2.1.

ExoCAM provides basic controls for setting the geophysical and atmospheric properties of the modelled planet, along with several options for controlling the modes of operation. ExoCAM permits flexible setting of surface type (land vs. ocean), planet radius, surface gravity, rotation rate and period, incident stellar flux, stellar spectrum, orbital eccentricity, obliquity, and partial pressure of atmospheric gases including \ce{N2},  \ce{CO2}, \ce{CH4}, \ce{H2O}, \ce{O3}, \ce{O2}, \ce{H2}, \ce{CO}, \ce{NH3}, along with cloud and aerosol species.  ExoCAM is multi-configurable, allowing the user to configure with a variety of horizontal and vertical grid resolutions, along with different choices for dynamical core, convection, cloud, and aerosol physics.  The simulations featured in this dataset uniformly use the finite volume (FV) dynamical core  \citep{lin&rood:1996} along with CAM4 cloud and convection physics \citep{hack:1994, zhang&mcfarlane:1995, rasch&kristjansson:1998}.  Likewise the simulations featured here use a horizontal resolution of $4^\circ\times 5^\circ$, but vertical levels used are mixed between  40 or 51 layers, representing 3 to 5 orders of magnitude extents in pressure space.

Post-2022 ExoCAM studies featured in the main training set are distinguished from early studies by a significant update to the radiative transfer component, ExoRT.  Pre-2022 studies included in the auxiliary training set used an evolving set of radiative transfer configurations (e.g. \citet{yang2016differences, kopparapuHabitableMoist2017Corrected, wolfSimulatedPhasedependent2019}).
By 2022, development had settled down to more-or-less its current form as described in \citet{wolfExoCAM3D2022}.

\paragraph{UM.}
The Unified Model (UM) developed by the UK Met Office consists of the dynamical core, Even Newer Dynamics for General atmospheric modelling of the environment (ENDGame), which uses a semi-implicit semi-Lagrangian scheme to solve the non-hydrostatic, full deep-atmosphere equations of motion with varying gravity within the atmosphere \citep[see][for discussion]{Wood_etal2014,Mayne_etal2014a,Mayne_etal2014b}. The UM has been used to perform 3D climate simulation across a wide range of planets, ranging across the modern Earth \citep{Walters_etal2019,Andrews_etal2020}, rocky planets \citep{boutleExploringClimate2017,sergeevTRAPPIST1Habitable2022,mak3DSimulations2024} and gas giant exoplanets \citep{christieCAMEMBERTMiniNeptunes2022,Zamyatina_etal2023,Zamyatina_etal2024,Mak_etal2025}. The UM also consists of a 2-stream radiative transfer scheme, the ``Suite of Community RAdiative Transfer codes based on \citet{Edwards_and_Slingo_1996} (Socrates), and uses the correlated-\textit{k} method to solve for gaseous absorption from \ce{H2O}, \ce{CO2}, \ce{O3}, \ce{N2O}, \ce{CH4}, \ce{O2}, \ce{SO2}, \ce{OCS} from HITRAN \citep{Rothman_etal2013} and collision-induced absorption from \ce{N2}-\ce{N2}, \ce{N2}-\ce{CH4}, \ce{CO2}-\ce{CO2} from HITRAN \citep{Karman_etal2019}, and \ce{CH4}-\ce{CO2} \citep{Turbet_etal2020_CIA}. Socrates is also used to construct the configuration file for the climate simulations which contains the optical properties of the gases in the shortwave (stellar) and longwave (planetary) part of the spectrum. The stellar spectrum is generated with the PHOENIX stellar model \citep{Husser_etal2013} and the star is assumed to have $\log(g)=6.0$ and [Fe/H]$=0.0$ for simplicity. The shortwave range spans 0.2--10\,$\mu$m and is binned into 6 bands, whereas the longwave range extends from 3.34--10$^4$\,$\mu m$ and is binned into 9 bands.

The climate simulations are performed in a horizontal grid spacing of 2.5$^\circ$ in longitude and 2$^\circ$ in latitude, and a quadratically stretched vertical grid spacing of 38 layers to allow for higher resolution near the planetary surface. The vertical grid of the UM is altitude-based and the model domain height is fixed at 39.25\,km across all simulations to maintain model stability. Lambert-Lewis (LLCS) simple moist adjustment scheme \citep{LLCS_cloud_scheme} is used in the cloud treatment within the UM. All simulations are run for at least 20\,Earth years to reach an equilibrium state. This is diagnosed by requiring that fluctuations in surface temperature and the top-of-atmosphere flux remain below 1\% over the last 10\,Earth years of the simulation time. The results presented in this work are temporally averaged over this final 10-year period.

\paragraph{LFRic-Atmosphere.}
LFRic-Atmosphere is the next-generation 3D GCM of the Met Office, designed for exascale computing \citep{Adams19_lfric, Bull24_performance}.
This model combines a new dynamical core GungHo and a suite of well-tested physical parameterizations inherited from its forerunner, the Unified Model (see previous section).
Like the UM, GungHo solves the fully compressible non-hydrostatic Euler equations on a sphere but it uses a quasi-uniform cubed-sphere finite-element discretisation \citep{Melvin19_mixed, Melvin24_mixed}.
For transport, the latest GungHo version uses a flux-form semi-Lagrangian scheme that ensures local conservation of mass and entropy, while maintaining other key properties of numerical schemes such as preservation of a constant, monotonicity and positivity \citep{Bendall25_swift}.
LFRic-Atmosphere is capable of reproducing a variety of atmospheric flows in different model configurations and domain geometries \citep{Kent23_mixed, Brown24_physics, Johnson25_regional}, including idealised Earth-like exoplanet setups \citep{sergeevSimulationsIdealised2023, Sergeev24_impact}.

The LFRic-Atmosphere simulation used for these runs is derived from \citet{sergeevSimulationsIdealised2023} and based on the \ce{N2}-dominated aquaplanet case of the TRAPPIST-1 Habitable Atmosphere Intercomparison \citep[THAI,][]{fauchezTRAPPIST1Habitable2020, sergeevTRAPPIST1Habitable2022}.
The three parameters that we vary in this study are (i) planet's radius (0.5--1.7$R_\text{T1e}$), (ii) stellar flux (0.6--1.7$S_\text{T1e}$) and (iii) surface gravity (0.9--1.15$g_\text{T1e}$).
We use the C24 cubed-sphere mesh resolution (i.e., $24\times24\times6$ atmospheric columns).
In the vertical, the model uses the same vertical spacing as that in the UM: 38 quadratically stretched layers to allow for higher resolution near the surface.
All simulations were run until a steady state.

\paragraph{ExoPlaSim.}
ExoPlaSim is an intermediate-complexity GCM that is a modification of the Planet Simulator (PlaSim) for planets beyond Earth.
It has been used to study exoplanet habitability (e.g., \citealp{paradiseExoPlaSimExtending2022, chenSporadicSpinorbit2023, cohenHazeOptical2024, macdonaldClimateTransition2025}) and for rocky planet parameter sweeps generally \citep{paradiseLargeRepository2020, macdonaldClimateUncertainties2022}.
ExoPlaSim has also been compared with higher complexity models like ExoCAM and found to replicate global climate patterns to first order \citep{paradiseExoPlaSimExtending2022}.
However, ExoPlaSim lacks some of the advanced physics that higher complexity models have.
For instance, it is prone to Gibbs oscillations when simulating tidally locked planets (a known problem in climate simulation more generally; \citealp{landerBelievableScales1997}).
For the simulations generated for this work, we use a T42 grid resolution and an exponential filter (the most consistent filter at low resolutions; \citealp{paradiseExoPlaSimExtending2022}) to mitigate Gibbs oscillations.

\subsection{Regridding details}\label{app:regridding}

The 10 pressure levels are defined as relative isobars
$\sigma_k=(P_k-P_\text{top})/(f_\text{bottom}P_0-P_\text{top})$,
where $P_0$ is the input surface pressure, $P_\text{top}=10\,\text{mbar}$,
and $f_\text{bottom}=0.925$ lifts the lowest level above near-surface
pressure fluctuations. The $\sigma_k$ are spaced according to a
fourth-order polynomial that increases resolution near the top and
bottom of the atmosphere. The $32\times 64$ horizontal grid is a T21
Gaussian grid, supporting an exact spherical harmonic transform up to
total wavenumber $\ell_\text{max}=21$. Horizontal interpolation is
bilinear; vertical interpolation is in log-pressure. Any field with
partially missing values across the spatial grid (which can arise from
pressure fluctuations in GCMs using height-based coordinates) is
treated as fully unobserved.

\subsection{Missingness patterns}\label{app:missingness-patterns}

\begin{figure}[t]
  \centering
  \includegraphics[width=\linewidth]{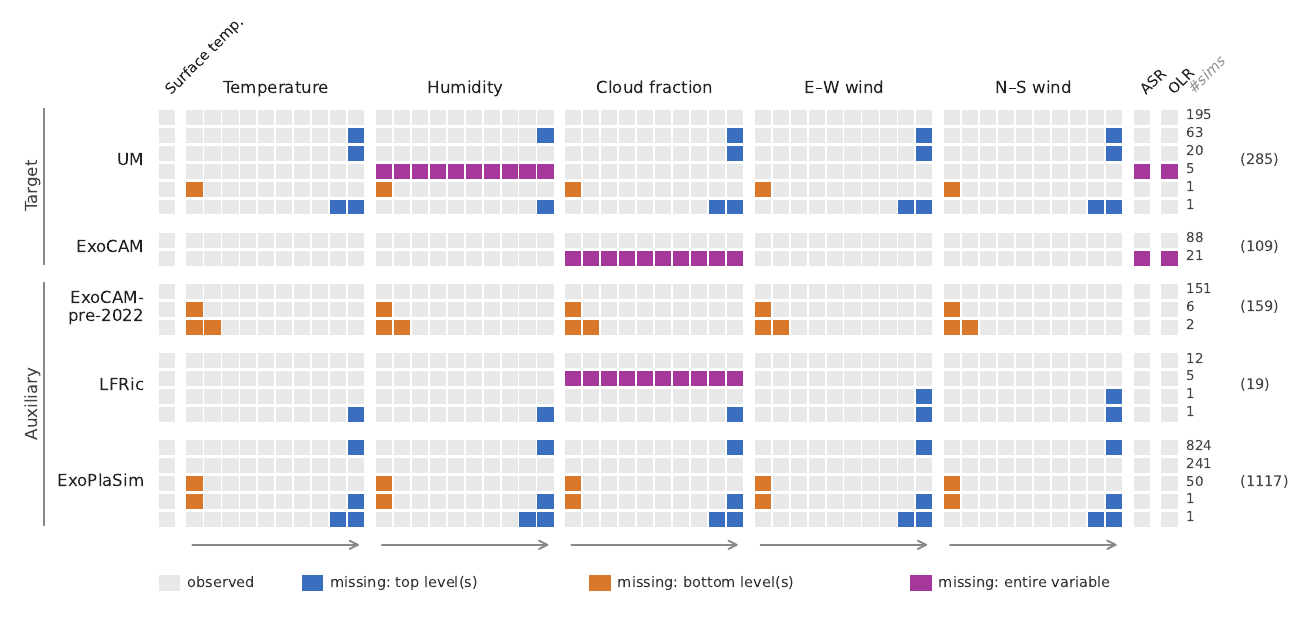}
  \caption{Observation patterns in the full dataset (Multi-partial). Each row is a distinct observation pattern occurring within one GCM, and the right-hand column counts the simulations sharing it. Columns are the 53 fields: three single-level variables, then five 3D variables on the common 10-level pressure grid ordered surface to top. Unobserved fields come in three kinds (Section~\ref{sec:challenges}): uppermost pressure levels (blue), lowermost pressure levels (orange), and entire variables (purple). Missingness is blockwise: a simulation either provides a field on the full spatial grid or not at all, and patterns are shared across simulator and study blocks rather than varying per example. The 48-field complete-observation subsets (Section~\ref{sec:subsets}) drop the top level of each 3D variable and exclude the simulations still missing any remaining field.}
  \label{fig:missingness-patterns}
\end{figure}

Figure~\ref{fig:missingness-patterns} shows every distinct observation pattern in the dataset.
The dominant pattern is a missing top level: the native vertical grids of most ExoPlaSim simulations do not extend to 10\,mbar, leaving the top level of each 3D variable unobserved in 826 of the 1,117 ExoPlaSim simulations.
Entire-variable absences are generally studywise; for example, the 21 ExoCAM simulations from \citet{hammondClimatesThermal2025} do not include cloud fraction, ASR, or OLR.
Overall, 5.8\% of simulation--field pairs are unobserved.

\subsection{Metric aggregation}\label{app:metric-aggregation}

For evaluation under the standard protocol, we average each field's score over test simulations (excluding those missing that field), then uniformly over pressure levels for 3D variables.
Under the shared-planets protocol, both GCMs take turns as target and predictor for each planet (using only fields available from both).
For a 3D variable, we first average the emulator errors over the two target GCMs and the shared pressure levels, and average the GCM disagreement over those same levels; we then form their ratio for that planet and average the ratios uniformly across planets.

\subsection{Test-set composition}\label{app:test-composition}

Of the 128 Multi-partial test simulations, 86 are bespoke simulations from the sampling design of Appendix~\ref{app:sampling} and 42 come from literature studies. 
The literature-derived test simulations are easier for every baseline than the bespoke ones. This is likely because literature studies tend to cluster around community-favourite planets and vary only a subset of the eight inputs, so most of them have a close neighbour in training, whereas the bespoke simulations are chosen to fill gaps in input space (subject to physical plausibility), so their neighbours in training are further away.
When the two parts are scored separately, GPLFR's geometric-mean RMSE relative to Train-mean is 0.32 on the literature-derived part and 0.53 on the bespoke part.
GPLFR and PPCA-ICM remain first and second on the bespoke part, and kNN falls to ninth of the ten baselines, ahead of only Train-mean.
Otherwise, only the order of the closely matched middle methods changes.

\section{Baseline details}\label{app:baselines}

\subsection{Shared settings}\label{app:baseline-shared-settings}

\paragraph{Inputs and outputs.}
All baselines start from the transformed inputs and outputs described in Section~\ref{sec:preprocessing}.
Inputs are then z-scored using training-set statistics.
Output processing depends on the method family.
Grid-space methods (Coord-MLP, Coord-DeepONet, ConvDec, SFNO) operate on the transformed gridded fields, z-scored per field using training-set statistics.
Spectral-space methods (PCA-MLP, PCA-GBT, PPCA-ICM, GPLFR) first expand each horizontal field in the T21 spherical harmonic basis, then centre and scale per field, i.e., for field $k$, we compute the training-set mean $\bar{\mathbf{a}}^{(k)}$ and the root-mean anomaly energy
\begin{equation*}
  \sigma^{(k)} = \sqrt{\frac{1}{N} \sum_{i=1}^{N} \|\mathbf{a}_i^{(k)} - \bar{\mathbf{a}}^{(k)}\|_2^2}
\end{equation*}
and set $\mathbf{y}_i^{(k)} = (\mathbf{a}_i^{(k)} - \bar{\mathbf{a}}^{(k)}) / \sigma^{(k)}$.
Dividing by $\sigma^{(k)}$ equalizes the total anomaly variance across fields.
Predictions are denormalized by inverting these steps before evaluation.
The simple baselines (training mean, kNN) operate directly on the transformed gridded fields without further standardization.

\paragraph{Equatorial symmetry.}
The released targets are symmetrized as described in Section~\ref{sec:preprocessing}.
All predictions are symmetrized in the same way before evaluation.
For grid-space methods, symmetric fields are averaged across hemispheres, except for N--S wind which is replaced by half the hemisphere difference.
For spectral-space methods, this can be done more neatly by zeroing the complementary spherical harmonic coefficients (retaining $\ell+m$ even for symmetric fields, $\ell+m$ odd for the antisymmetric N--S wind).

\paragraph{Output-space linear trend function.}
The spectral-space methods subtract an affine function of the planet parameters with per-GCM intercepts from each spectral coefficient before training, fit by ridge regression (penalty $10^{-3}$) on the training set and added back at prediction time.

\paragraph{Missing data handling.}
Due to differences in GCM output grids, in the full dataset (Multi-partial), some examples have unobserved output fields (Section~\ref{sec:challenges}).
All methods handle this by restricting to observed fields only.
The training mean and kNN average fields only over examples where that field is observed.
Coord-MLP, Coord-DeepONet, ConvDec, and SFNO exclude missing fields from the training loss.
PCA-MLP, PCA-GBT, and PPCA-ICM implement PCA as probabilistic PCA (PPCA), omitting missing fields from the likelihood.
PCA-MLP and PCA-GBT retain only the point-estimate scores, while PPCA-ICM combines PPCA noise with GP uncertainty for its ensemble predictions.
GPLFR handles missingness directly in its collapsed decoder likelihood by restricting each output dimension's contribution to only examples where it is observed.

\subsection{Models}\label{app:baseline-models}

\paragraph{Training mean.}
The prediction for every test input is the per-field area-weighted mean over the training set.

\paragraph{kNN.}
For each test input, we find the $k$ nearest training examples in standardized continuous input space using Euclidean distance.
GCM identity is encoded as a scaled one-hot vector appended to the input, so that different GCMs contribute an additional distance penalty $\lambda\sqrt{2}$, encouraging same-GCM neighbours while still permitting cross-GCM matches.
The prediction is the uniform average of the neighbours' spatial output fields.
We tuned $k$ and $\lambda$ (Table~\ref{tab:hyperparameter-tuning}).

\paragraph{PCA-GBT.}
Uses the same compress-then-predict pipeline as PCA-MLP -- a shared linear trend removed before PPCA compression of the normalized spectral coefficients -- but replaces the MLP with histogram-based gradient-boosted regression trees, fitting one ensemble per latent score from the eight standardized planet parameters and a one-hot GCM label.
Predictions are decoded through the PCA loadings and means without noise sampling.
We tuned the number of PCs, boosting learning rate and maximum leaves per tree (Table~\ref{tab:hyperparameter-tuning}); the remaining settings are fixed off-the-shelf (600 boosting iterations with early stopping, a minimum of 10 samples per leaf, and L2 regularization of 1.0).

\paragraph{Coord-MLP.}
This baseline adapts the pointwise emulator strategy of \citet{plaschzugAcceleratingExoplanet2025}, who trained a network to predict 3D temperature and wind fields for hot Jupiters from a 2D parameter space of 60 GCM simulations.
The model takes as input the concatenation of the eight standardized planet parameters, a one-hot GCM label, a one-hot variable indicator, and the three spatial coordinates (pressure level, latitude, longitude), and predicts a scalar field value.
We tuned the network width, depth, and learning rate (Table~\ref{tab:hyperparameter-tuning}).
The remaining model settings were fixed to Plaschzug et al.-style choices: RMSProp optimizer, tanh activations, batch size 128, zero weight decay, trained with unweighted pointwise MSE.
The main adaptation relative to Plaschzug et al.\ is the input specification: their model conditions on a 2D parameter space (stellar effective temperature and global-mean temperature) with four output variables, whereas ours conditions on eight planet parameters with GCM and variable identity tokens, predicting across all output fields.
We also condition on a discrete pressure-level coordinate rather than continuous pressure, since ThousandWorlds outputs are defined on fixed relative isobars (Appendix~\ref{app:regridding}).

\paragraph{Coord-DeepONet.}
Uses a DeepONet architecture \citep{luLearningNonlinear2021}, which models nonlinear operators via an inner product between a branch network and a trunk network over a learned basis of rank $R$.
The branch takes the eight standardized planet parameters concatenated with a one-hot GCM label.
The trunk takes a one-hot variable indicator, normalized vertical level, latitude coordinate $\sin(\lambda)$, and periodic longitude features $(\sin\phi, \cos\phi)$.
The predicted value at a query point is
\begin{equation*}
  \hat{y}(\mathbf{x}, s, f, \ell, \lambda, \phi) =
  b_0(f, \ell, \sin\lambda, \sin\phi, \cos\phi) +
  \langle \mathbf{b}(\mathbf{x}, s), \mathbf{t}(f, \ell, \sin\lambda, \sin\phi, \cos\phi) \rangle / \sqrt{R},
\end{equation*}
where $\mathbf{x}$ is the planet parameter vector, $s$ the GCM label, $f$ the variable, $\ell$ the vertical level, and $\mathbf{b}$, $\mathbf{t}$, and $b_0$ are the branch, trunk, and bias networks respectively.
All three are MLPs with SiLU activations.

Training uses scalar coordinate minibatches.
Minibatches are sampled by first choosing a variable uniformly, then a field within that variable (a single selection for single-level variables, a pressure level for 3D variables), then a planet observed for that field, then latitude proportional to area weights and longitude uniformly.
For simplicity, branch, trunk, and bias MLPs share a single depth parameter.
We use AdamW with weight decay $10^{-4}$ and batch size $2^{15}$.
We tuned rank, branch and trunk widths, shared depth, and learning rate (Table~\ref{tab:hyperparameter-tuning}).

\paragraph{PCA-MLP.}
Fits PCA on the normalized spectral coefficients to extract latent scores.
(When fields are missing, the PCA fit omits them from the likelihood, making it technically PPCA, but only point estimates are retained.)
A two-layer MLP (SiLU activations) then maps the inputs (eight standardized planet parameters and a one-hot GCM label) to latent scores.
We minimize MSE on the PCA scores using AdamW.
Predictions are decoded through the PCA loadings and means without noise sampling.
We tuned depth, number of PCs, hidden width (shared across layers), learning rate, and weight decay (Table~\ref{tab:hyperparameter-tuning}).

\paragraph{ConvDec.}
A deterministic, parameter-conditioned convolutional decoder.
A linear layer maps the input vector (eight standardized planet parameters and a one-hot GCM label) to a $4\times 8$ feature grid with $C$ channels, establishing the global spatial layout.
Three decoder stages then each apply $\times 2$ nearest-neighbour upsampling followed by $3\times 3$ convolutions (SiLU activations), reaching the full $32\times 64$ resolution, with the channel count halved between successive stages; a final $1\times 1$ convolution produces one channel per field (53 for Multi-partial, 48 for the complete subsets).
This coarse-to-fine design -- project a vector to a small feature grid, then progressively upsample it with shared convolutional filters -- follows the DCGAN generator of \citet{radfordUnsupervisedRepresentation2016} (we borrow the architecture only, training it by supervised regression rather than adversarially).
We separate resizing from convolution, with nearest-neighbour resizing, to avoid the uneven-overlap artifacts associated with transposed convolutions \citep{odenaDeconvolutionCheckerboard2016}.
Convolutions use circular padding in longitude and zero padding in latitude.
Training minimizes latitude-weighted MSE over observed fields using AdamW with batch size 32.
We tuned the seed channel count $C$, the number of convolutions per stage, the learning rate, and the weight decay (Table~\ref{tab:hyperparameter-tuning}).

\paragraph{SFNO.}
Uses the Spherical Fourier Neural Operator of \citet{bonevSphericalFourier2023} via the \texttt{torch-harmonics} reference implementation\footnote{\url{https://github.com/NVIDIA/torch-harmonics}, version 0.8.0.} with its linear, zonal-kernel spectral filters: each SFNO block applies a spherical harmonic transform, a learned complex channel-mixing matrix for each degree $\ell$ shared across orders $m$, the inverse transform, and a pointwise GELU MLP with a residual connection.
Pointwise linear projections map the input channels to the embedding dimension and back to one output channel per field.
We use the native $32\times 64$ Legendre--Gauss grid without internal down-scaling and retain degrees $0\leq\ell\leq 21$.
We disable spatial normalization layers and the input-residual path, and add a learned position embedding to the encoded features; these are explicit configuration choices rather than reference defaults.

To adapt this field-to-field architecture to parameter-to-field regression, we broadcast the inputs (eight standardized planet parameters and a one-hot GCM label) as spatially constant channels, similar in spirit to the FNO input-space extensions of \citet{luComprehensiveFair2022}.
A constant field excites only the $\ell=0$ mode, and per-degree filters and pointwise operations preserve spatial constancy, so the learned position embedding is the sole source of spatial symmetry breaking (spatial normalization would remove precisely the $\ell=0$ conditioning component, hence its exclusion).
This anchoring to fixed spatial coordinates is appropriate because the grids are substellar-fixed -- the same property exploited by Coord-MLP and Coord-DeepONet -- though it means the full predictor, unlike the spectral filters themselves, is not rotation-equivariant.
Training follows ConvDec (latitude-weighted MSE over observed fields, AdamW, batch size 32).
We tuned the embedding dimension, number of blocks, learning rate, and weight decay (Table~\ref{tab:hyperparameter-tuning}).

\paragraph{PPCA-ICM.}
PPCA-ICM is a two-stage compress-then-predict pipeline on normalized spectral coefficients.

\emph{Stage 1: PPCA compression.}
PPCA compresses each example's spectral coefficients to a latent score vector, fit by EM.
Missing fields are handled by omitting their likelihood terms.

\emph{Stage 2: ICM-GP regression.}
The PPCA scores are regressed against inputs $(\mathbf{x}, s)$ using GPs with an ICM kernel: $k((\mathbf{x}, s), (\mathbf{x}', s')) = k_x(\mathbf{x}, \mathbf{x}') B^\text{in}_{ss'}$, where $k_x$ is an ARD kernel on the continuous planet parameters and $\mathbf{B}^\text{in} \in \mathbb{R}^{S \times S}$ is a coregionalization matrix across GCMs; Table~\ref{tab:hyperparameter-tuning} gives the selected kernel family for each subset.
All scores share common lengthscales, amplitudes, and $\mathbf{B}^\text{in}$; a per-component regime (separate lengthscales and amplitudes per score, with shared $\mathbf{B}^\text{in}$) performed worse.
Kernel hyperparameters are fit by maximizing the GP marginal likelihood using Adam.

At prediction, ensemble members are generated by sampling scores from the GP predictive distribution, decoding through the PPCA loadings and mean, and adding PPCA noise.
Deterministic predictions use the GP predictive mean decoded through the PPCA loadings.
We tuned the number of PCs, learning rate, and kernel family for each subset (Table~\ref{tab:hyperparameter-tuning}).

\paragraph{GPLFR.}
GPLFR is a GP-based model designed for high-dimensional structured outputs \citep{stevensonGaussianProcessLatent2026}.
It models each output as a linear decoding of a low-dimensional latent state drawn from a GP prior over the inputs.
The latents and GP kernel hyperparameters are jointly optimized under a tempered MAP objective.
In our configuration, GPLFR is essentially a more flexible, end-to-end version of PPCA-ICM.
It likewise uses an ARD kernel with ICM coregionalization over GCMs.
Optimization uses Adam with separate learning rates for the latent variables and global parameters.
Ensemble predictions are generated as in PPCA-ICM: sampling latents, decoding, and sampling observation noise.
We tuned latent dimensionality, regularization parameters (inverse-temperature and latent noise), learning rates and kernel family (Table~\ref{tab:hyperparameter-tuning}).

\begin{table}[htbp]
  \centering
  \scriptsize
  \setlength{\tabcolsep}{1pt}
  \renewcommand{\arraystretch}{1.0}
  \caption{Hyperparameter search grids and selected values. Final steps were the median best validation steps across CV folds.}
  \label{tab:hyperparameter-tuning}
  \begin{tabular}{@{}>{\raggedright\arraybackslash}p{2.7cm} >{\raggedright\arraybackslash}p{1.4cm} >{\raggedright\arraybackslash}p{1.4cm} >{\raggedright\arraybackslash}p{1.45cm} >{\raggedright\arraybackslash}p{1.55cm} >{\raggedright\arraybackslash}p{1.4cm} >{\raggedright\arraybackslash}p{3.6cm}@{}}
    \toprule
    & \multicolumn{5}{c}{Subset(s)} & \\
    \cmidrule(lr){2-6}
    Hyperparameter & Multi-partial & Multi-complete & Single-complete & Target-GCMs-only ablation & Target-only ablation & Candidates \\
    \midrule
    \multicolumn{7}{@{}l}{\textbf{kNN}} \\
    $k$ & 3 & 3 & 2 & 1 & 2 & [1, 2, 3, 5, 10] \\
    GCM penalty & 1 & 1 & 0 & 1 & 10 & [0, 0.3, 1, 3, 10] \\
    \midrule
    \multicolumn{7}{@{}l}{\textbf{PCA-GBT}} \\
    Number of PCs & 150 & 150 & 50 & 50 & 50 & [50, 150] \\
    Learning rate & 0.03 & 0.05 & 0.1 & 0.1 & 0.1 & [0.03, 0.05, 0.1] \\
    Max leaf nodes & 63 & 31 & 15 & 15 & 15 & [15, 31, 63] \\
    \midrule
    \multicolumn{7}{@{}l}{\textbf{Coord-MLP}} \\
    Width & 1024 & 1024 & 512 & 1024 & 1024 & [128, 256, 512, 1024] \\
    Depth & 4 & 4 & 6 & 4 & 4 & [2, 4, 6] \\
    Learning rate & $3\cdot 10^{-4}$ & $3\cdot 10^{-4}$ & $3\cdot 10^{-4}$ & $3\cdot 10^{-4}$ & $3\cdot 10^{-4}$ & [$3\cdot 10^{-5}$, $10^{-4}$, $3\cdot 10^{-4}$, 0.001, 0.003] \\
    Final step & 7800 & 7500 & 7900 & 7600 & 7500 & --- \\
    \midrule
    \multicolumn{7}{@{}l}{\textbf{Coord-DeepONet}} \\
    Rank & 128 & 128 & 32 & 64 & 32 & [32, 64, 128, 256] \\
    Branch width & 256 & 256 & 256 & 512 & 512 & [64, 128, 256, 512] \\
    Trunk width & 512 & 512 & 256 & 512 & 512 & [64, 128, 256, 512] \\
    Depth & 3 & 3 & 3 & 3 & 3 & [2, 3] \\
    Learning rate & 0.003 & 0.003 & 0.003 & 0.003 & 0.003 & [$3\cdot 10^{-5}$, $10^{-4}$, $3\cdot 10^{-4}$, 0.001, 0.003] \\
    Final step & 1500 & 1000 & 2000 & 4250 & 1500 & --- \\
    \midrule
    \multicolumn{7}{@{}l}{\textbf{PCA-MLP}} \\
    Number of PCs & 50 & 50 & 20 & 50 & 50 & [20, 50, 100, 150] \\
    Depth & 2 & 2 & 2 & 2 & 2 & [1, 2] \\
    Hidden widths & 1024 & 1024 & 512 & 1024 & 1024 & [32, 64, 128, 256, 512, 1024] \\
    Learning rate & $3\cdot 10^{-4}$ & $3\cdot 10^{-4}$ & 0.001 & $10^{-4}$ & $10^{-4}$ & [$10^{-4}$, $3\cdot 10^{-4}$, 0.001, 0.003, 0.01] \\
    Weight decay & $10^{-4}$ & $10^{-4}$ & $10^{-4}$ & $3\cdot 10^{-5}$ & $3\cdot 10^{-5}$ & [$3\cdot 10^{-5}$, $10^{-4}$, $3\cdot 10^{-4}$, 0.001, 0.003, 0.01, 0.03] \\
    Final step & 600 & 500 & 300 & 900 & 700 & --- \\
    \midrule
    \multicolumn{7}{@{}l}{\textbf{PCA-Ridge}} \\
    Number of PCs & 150 & 150 & 20 & --- & --- & [20, 50, 100, 150] \\
    Ridge penalty & $10^{-4}$ & $10^{-6}$ & $10^{-6}$ & --- & --- & [$10^{-6}$, $10^{-4}$, 0.001, 0.01, 0.1, 1] \\
    \midrule
    \multicolumn{7}{@{}l}{\textbf{ConvDec}} \\
    Seed channels & 2048 & 2048 & 2048 & 2048 & 2048 & [64, 128, 256, 512, 1024, 2048] \\
    Convs per stage & 1 & 1 & 2 & 2 & 2 & [1, 2, 3] \\
    Learning rate & $10^{-4}$ & $10^{-4}$ & $10^{-4}$ & $10^{-4}$ & $10^{-4}$ & [$10^{-4}$, $3\cdot 10^{-4}$, 0.001, 0.003, 0.01] \\
    Weight decay & 0.3 & 0.3 & 0.3 & 0.3 & 0.1 & [$3\cdot 10^{-5}$, $10^{-4}$, $3\cdot 10^{-4}$, 0.001, 0.003, 0.01, 0.03, 0.1, 0.3] \\
    Final step & 2700 & 2900 & 2100 & 2500 & 2500 & --- \\
    \midrule
    \multicolumn{7}{@{}l}{\textbf{SFNO}} \\
    Embedding dim. & 768 & 768 & 768 & 768 & 512 & [64, 128, 256, 384, 512, 768] \\
    Number of blocks & 2 & 2 & 2 & 4 & 4 & [1, 2, 4, 6, 8] \\
    Learning rate & $3\cdot 10^{-4}$ & $3\cdot 10^{-4}$ & 0.001 & 0.001 & 0.001 & [$10^{-4}$, $3\cdot 10^{-4}$, 0.001, 0.003, 0.01] \\
    Weight decay & 0.3 & 0.3 & 0.3 & 0.3 & 0.3 & [$3\cdot 10^{-5}$, $10^{-4}$, $3\cdot 10^{-4}$, 0.001, 0.003, 0.01, 0.03, 0.1, 0.3] \\
    Final step & 2800 & 3000 & 2400 & 2800 & 2800 & --- \\
    \midrule
    \multicolumn{7}{@{}l}{\textbf{PPCA-ICM}} \\
    Kernel family & Mat\'ern-3/2 & Mat\'ern-3/2 & Mat\'ern-3/2 & Mat\'ern-3/2 & Mat\'ern-3/2 & [Mat\'ern-3/2, Mat\'ern-5/2, RBF] \\
    Number of PCs & 150 & 150 & 150 & 150 & 150 & [20, 50, 100, 150] \\
    Learning rate & 0.003 & 0.003 & 0.001 & 0.03 & 0.03 & [$3\cdot 10^{-4}$, 0.001, 0.003, 0.01, 0.03] \\
    Final step & 620 & 600 & 960 & 60 & 40 & --- \\
    \midrule
    \multicolumn{7}{@{}l}{\textbf{GPLFR}} \\
    Kernel family & Mat\'ern-5/2 & Mat\'ern-5/2 & Mat\'ern-3/2 & Mat\'ern-3/2 & Mat\'ern-3/2 & [Mat\'ern-3/2, Mat\'ern-5/2, RBF] \\
    Latent dimensionality & 150 & 150 & 150 & 150 & 150 & [50, 100, 150] \\
    Inverse-temperature & 0.1 & 0.1 & 0.03 & 0.03 & 0.03 & [0.01, 0.03, 0.1] \\
    Latent noise & 0.10 & 0.03 & 0.03 & 0.03 & 0.03 & [0.01, 0.03, 0.1] \\
    Latent learning rate & 0.1 & 0.1 & 0.1 & 0.1 & 0.1 & [0.03, 0.1, 0.3] \\
    Global learning rate & 0.3 & 0.3 & 0.3 & 0.1 & 0.1 & [0.03, 0.1, 0.3] \\
    Final step & 115 & 120 & 30 & 60 & 60 & --- \\
    \bottomrule
  \end{tabular}
\end{table}

\subsection{Hyperparameter tuning}\label{app:hyperparameter-tuning}

We select hyperparameters via 5-fold cross-validation (CV).
Validation sets contain only target simulations, grouped by planet, excluding any planet that an auxiliary GCM has also simulated.
Remaining training simulations appear in every fold.
For each candidate and fold, we fit all data-dependent preprocessing on its training set, and evaluate RMSE (in normalized spectral-coefficient or grid space depending on the method) on its validation set.
We select the setting with the best mean performance across folds, then refit on the full training set, using early stopping at the median best validation step across folds.
Tuned hyperparameters and search grids are shown in Table~\ref{tab:hyperparameter-tuning}.

For the learned methods, we search manually rather than exhaustively over the grid.
We tune primarily on the Multi-partial subset (the full dataset). We then do narrower retuning around these Multi-partial-selected values for the Multi-complete subset, and a larger retune for Single-complete (as it is significantly smaller).

\subsection{Compute resources}\label{app:compute}

All baselines except PCA-GBT were trained and evaluated on a single NVIDIA H100 GPU; ConvDec and SFNO used H100 SXM5 GPUs, while the other GPU baselines used H100 PCIe GPUs.
Table~\ref{tab:compute} reports wall-clock training times for the learned baselines on the Multi-partial subset (the full dataset).
Hyperparameter selection via 5-fold CV multiplies these costs by approximately $5\times$ the number of candidate settings evaluated per method (Appendix~\ref{app:hyperparameter-tuning}).
Train-mean and kNN require negligible compute.

\begin{table}[htbp]
  \centering
  \small
  \renewcommand{\arraystretch}{1.1}
  \caption{Approximate wall-clock training time per run on the Multi-partial subset. PCA-GBT is trained on 32 CPU cores; all other methods use a single H100 GPU.}
  \label{tab:compute}
  \begin{tabular}{@{}lc@{}}
    \toprule
    Method & Training time (minutes)\\
    \midrule
    PCA-GBT         & 2.7 \\
    Coord-MLP       & 0.4 \\
    Coord-DeepONet  & 0.9 \\
    PCA-MLP         & 0.2 \\
    ConvDec         & 1.8 \\
    SFNO            & 2.4 \\
    PPCA-ICM        & 1.3 \\
    GPLFR           & 1.2 \\
    \bottomrule
  \end{tabular}
\end{table}

\section{Ablations}\label{app:ablations}

\subsection{Dataset ablations}\label{app:dataset-ablations}

We investigate two ablations of training set composition on Multi-partial using the standard test set.
For each ablation, we retuned the models' key hyperparameters using the same CV protocol as the main experiments (Appendix~\ref{app:hyperparameter-tuning}; selected values in Table~\ref{tab:hyperparameter-tuning}).
The first ablation removes all auxiliary-GCM simulations (ExoPlaSim, LFRic, ExoCAM-pre-2022), reducing (Multi-partial) training from 1,527 to 266 target-GCM-only simulations. The second additionally removes the remaining 48 extended-domain simulations, leaving 218 core-domain-only simulations.
(See Table~\ref{tab:dataset-composition} for a reminder of the dataset composition.)

\paragraph{Effect of auxiliary GCM data (Table~\ref{tab:aux-gcm-public-rmse-ablation}).}
The effect of auxiliary GCM data is method-dependent: PCA-GBT benefits (by 2.0\% in mean seed-0 RMSE), GPLFR is essentially unchanged ($+0.1\%$), PPCA-ICM and PCA-MLP are roughly neutral ($-2.2\%$ and $-1.6\%$), and the remaining methods are harmed, most strongly SFNO ($-13.2\%$), Coord-DeepONet ($-12.8\%$), and Coord-MLP ($-11.7\%$).
So it is plausible that the auxiliary set's outputs are systematically different enough from those of the target GCMs to be causing negative transfer.
To see if we could reduce the harm caused by auxiliary GCM data for the strongly affected deep learning models (ConvDec, SFNO, Coord-MLP, and Coord-DeepONet), we tried subtracting the same output-space linear trend function with per-GCM intercepts that the spectral-space methods used (Appendix~\ref{app:baseline-shared-settings}) before training, but this only had a small, inconsistent effect on their RMSEs ($-2.6\%$ to $+1.5\%$ in geometric-mean RMSE), and the auxiliary-GCM harm persists ($-7.8\%$ to $-14.6\%$). The negative transfer seen is therefore not explained by these per-GCM offsets.
These findings suggest that better multi-fidelity/multi-simulator transfer methods are a promising direction for improving on our baselines -- particularly for those four deep learning methods.

\paragraph{Effect of extended-domain data (Table~\ref{tab:extended-domain-ablation}).}
The 48 extended-domain target-GCM simulations constitute around 18\% of the target-GCM-only training set.
Despite their small number, they are consistently beneficial for the learned methods: removing them increases RMSE for the vast majority of variables across the learned methods.
These simulations often sit near the boundaries of the core domain and may be helping anchor predictions there.

\begin{table}[htbp]
  \centering
  \small
  \setlength{\tabcolsep}{2pt}
  \renewcommand{\arraystretch}{1.1}
  \caption{Auxiliary GCM data ablation results on the Multi-partial subset. Each entry is the percentage change in RMSE when auxiliary GCM simulations are removed from the training set. Positive values indicate auxiliary data was helping; negative values indicate it was hurting. Results are reported for training seed 0.}
  \label{tab:aux-gcm-public-rmse-ablation}
  \resizebox{\linewidth}{!}{%
  \begin{tabular}{@{}>{\raggedright\arraybackslash}p{1.8cm} *{10}{>{\raggedleft\arraybackslash}p{1.35cm}}@{}}
    \toprule
    Variable (\%) & Train-mean & kNN & PCA-GBT & Coord-MLP & Coord-DeepONet & PCA-MLP & ConvDec & SFNO & PPCA-ICM & GPLFR \\
    \midrule
    Surface temp.  & $-3.7$ & $-14.5$ & $+4.9$ & $-23.4$ & $-16.9$ & $-3.1$ & $-18.5$ & $-20.6$ & $-5.5$ & $-3.0$ \\
    Temperature    & $-3.3$ & $-4.3$ & $+6.4$ & $-27.1$ & $-15.6$ & $-2.6$ & $-16.5$ & $-22.1$ & $-2.9$ & $+3.7$ \\
    Humidity       & $-0.8$ & $-14.2$ & $+7.2$ & $-23.7$ & $-13.3$ & $+3.8$ & $-16.3$ & $-17.9$ & $+3.8$ & $+2.3$ \\
    Cloud fraction & $-6.2$ & $-3.7$ & $+2.7$ & $+5.1$ & $-9.1$ & $+2.5$ & $-10.5$ & $-10.3$ & $-3.2$ & $-1.4$ \\
    E--W wind      & $-4.4$ & $-3.8$ & $-2.7$ & $-11.5$ & $-5.5$ & $-0.6$ & $-10.6$ & $-11.0$ & $+0.7$ & $-1.8$ \\
    N--S wind      & $-7.2$ & $-5.8$ & $-6.7$ & $-6.2$ & $-16.2$ & $-1.3$ & $-4.6$ & $-8.4$ & $-0.5$ & $-5.3$ \\
    ASR            & $-25.0$ & $-6.3$ & $+0.3$ & $+14.2$ & $-14.1$ & $-6.5$ & $-2.4$ & $+0.6$ & $-6.3$ & $+4.6$ \\
    OLR            & $-18.9$ & $-10.3$ & $+4.0$ & $-21.0$ & $-11.9$ & $-4.6$ & $-6.0$ & $-15.7$ & $-3.6$ & $+1.9$ \\
    \midrule
    Mean           & $-8.7$ & $-7.9$ & $+2.0$ & $-11.7$ & $-12.8$ & $-1.6$ & $-10.7$ & $-13.2$ & $-2.2$ & $+0.1$ \\
    \bottomrule
  \end{tabular}
  }
\end{table}

\begin{table}[htbp]
  \centering
  \small
  \setlength{\tabcolsep}{2pt}
  \renewcommand{\arraystretch}{1.1}
  \caption{Extended-domain data ablation. Each entry is the percentage change in RMSE when the 48 target-GCM simulations outside the core domain are additionally removed from the target-GCM-only training set. Positive values indicate that these extended-domain data were helping. Results are reported for training seed 0.}
  \label{tab:extended-domain-ablation}
  \resizebox{\linewidth}{!}{%
  \begin{tabular}{@{}>{\raggedright\arraybackslash}p{1.8cm} *{10}{>{\raggedleft\arraybackslash}p{1.35cm}}@{}}
    \toprule
    Variable (\%) & Train-mean & kNN & PCA-GBT & Coord-MLP & Coord-DeepONet & PCA-MLP & ConvDec & SFNO & PPCA-ICM & GPLFR \\
    \midrule
    Surface temp.  & $+0.3$ & $-3.8$ & $-0.2$ & $+22.5$ & $+8.0$ & $+17.1$ & $+10.6$ & $+10.7$ & $+15.8$ & $+14.5$ \\
    Temperature    & $-0.8$ & $-6.2$ & $+1.0$ & $+12.9$ & $+5.9$ & $+14.3$ & $+5.4$ & $+12.5$ & $+10.4$ & $+10.9$ \\
    Humidity       & $-2.2$ & $-4.0$ & $-1.1$ & $+5.2$ & $+12.2$ & $+9.6$ & $+4.6$ & $+9.4$ & $+9.3$ & $+8.6$ \\
    Cloud fraction & $-0.0$ & $-5.9$ & $-1.7$ & $-6.7$ & $+7.8$ & $+0.7$ & $+3.4$ & $+7.2$ & $+2.5$ & $+7.0$ \\
    E--W wind      & $-0.2$ & $-0.3$ & $+1.7$ & $+2.4$ & $+2.6$ & $+5.7$ & $+4.9$ & $+4.5$ & $-0.4$ & $+5.2$ \\
    N--S wind      & $-0.1$ & $-1.4$ & $+3.8$ & $+0.1$ & $+9.3$ & $+5.2$ & $+1.9$ & $+1.2$ & $+0.4$ & $+5.4$ \\
    ASR            & $+0.5$ & $-10.5$ & $+0.5$ & $+2.6$ & $+13.3$ & $-1.6$ & $+18.5$ & $-1.7$ & $+7.1$ & $+5.0$ \\
    OLR            & $-1.3$ & $-6.5$ & $-4.1$ & $-6.4$ & $+7.5$ & $-2.8$ & $+10.5$ & $+5.5$ & $+2.6$ & $+4.8$ \\
    \midrule
    Mean           & $-0.5$ & $-4.8$ & $-0.0$ & $+4.1$ & $+8.3$ & $+6.0$ & $+7.5$ & $+6.2$ & $+6.0$ & $+7.7$ \\
    \bottomrule
  \end{tabular}
  }
\end{table}

\begin{table}[hbtp]
  \centering
  \small
  \setlength{\tabcolsep}{3pt}
  \renewcommand{\arraystretch}{1.1}
  \caption{The value of nonlinear regression. Each entry is the percentage increase in RMSE when PCA-MLP's MLP regressor is replaced with ridge regression. Positive values indicate ridge is worse. Results are reported for training seed 0. Unablated results are in Table~\ref{tab:rmse-partial-splits}.}
  \label{tab:subset-ablation-results}
  \begin{tabular}{@{}>{\raggedright\arraybackslash}p{2.5cm} *{3}{>{\raggedleft\arraybackslash}p{2.2cm}}@{}}
    \toprule
    Variable (\%) & Multi-partial & Multi-complete & Single-complete \\
    \midrule
    Surface temp.  & $+33.3$ & $+31.6$ & $+26.4$ \\
    Temperature    & $+39.8$ & $+41.1$ & $+32.8$ \\
    Humidity       & $+30.7$ & $+30.8$ & $+30.6$ \\
    Cloud fraction & $+18.3$ & $+22.9$ & $+11.7$ \\
    E--W wind      & $+16.4$ & $+24.9$ & $+8.2$ \\
    N--S wind      & $+15.2$ & $+14.3$ & $+7.6$ \\
    ASR            & $+20.4$ & $+26.8$ & $+30.8$ \\
    OLR            & $+26.2$ & $+28.8$ & $+29.2$ \\
    \midrule
    Mean           & $+25.0$ & $+27.7$ & $+22.1$ \\
    \bottomrule
  \end{tabular}
\end{table}

\subsection{Nonlinearity ablation}\label{app:nonlinearity-ablation}

To test how important a nonlinear input-to-latent map is, we replace the MLP in PCA-MLP with ridge regression, giving PCA-Ridge.
The ridge penalty is retuned per subset using five-fold CV, which selects $(10^{-4},10^{-6},10^{-6})$ on Multi-partial, Multi-complete, and Single-complete, respectively.
PCA-Ridge uses the same preprocessing and candidate PCA dimensions as PCA-MLP, but tunes both the number of PCs and ridge penalty independently.
PCA-Ridge has 25\% higher seed-0 RMSE than PCA-MLP averaged across variables and subsets (Table~\ref{tab:subset-ablation-results}), confirming that nonlinearity in the regression is important.
The penalty is positive for every variable and subset; averaged across subsets, it is smallest for N--S wind (12\%) and largest and most consistent for the temperature fields, indicating that the response of atmospheric temperature to planet parameters is substantially nonlinear.

\section{Additional discussion}\label{app:additional-discussion}

\subsection{Why is kNN strong on clouds, winds, and ASR?}\label{app:knn-discussion}

In Section~\ref{sec:baseline-comparison} we noted that kNN is relatively weak on temperature fields and humidity, but competitive with learned methods on cloud fraction, winds, and ASR.
Taking winds as an example, kNN is likely strong because global wind patterns are quite spatial-template-like: nearby planets often share the same broad circulation regime, so local averaging already captures much of the learnable variation.
The remaining wind errors may then be dominated by circulation regime transitions (see, e.g., \citealp{haqq-misraDemarcatingCirculation2018}) or spatial shifts of wind features like jets or vortices, which are harder to learn than the smoother global trends of temperature fields, for example, particularly under $L_2$-based objectives \citep{subichFixingDouble2025}.
Clouds and ASR (after removing the fixed insolation geometry component) are also relatively regime-determined versus trend-determined \citep{yangStabilizingCloud2013}, and so kNN could be strong on these variables for the same reason.
Cloud fraction is additionally highly dependent on both GCM and parameterization choices (as suggested by the low relative RMSE of Train-mean in Table~\ref{tab:relative-rmse}; see also, e.g., \citealp{sergeevTRAPPIST1Habitable2022}), limiting any cross-GCM positive transfer that might otherwise help the learned methods.

We note that kNN's competitiveness also partly reflects the composition of the test set (Appendix~\ref{app:test-composition}). On the literature-derived part, close neighbours in training are more common, and kNN ranks third of the ten baselines on cloud fraction and both winds, fifth on ASR, and sixth by geometric mean; on the bespoke part, kNN performs much worse, ranking seventh to ninth on each of these variables and ninth by geometric mean.

\subsection{Physically motivated GPLFR extensions}\label{app:gplfr-extensions}

GPLFR is the best-performing baseline and provides natural entry points for incorporating domain structure that are not available in the next-best methods.
We consider two physically motivated extensions:
\begin{enumerate}
  \item \textbf{Learned field--field correlations.} The default GPLFR output coregionalization matrix is $\mathbf{B}=\mathbf{I}_{D_y}$, where $D_y$ is the output dimensionality. This assumes output dimensions are conditionally independent given the latents. This is reasonable across spectral coefficients, which are approximately uncorrelated by construction, but restrictive across physical fields -- particularly different vertical levels of the same atmospheric variable. We relax this by setting $\mathbf{B}=\mathbf{I}_A\otimes\mathbf{B}_F$, where $A$ is the number of spectral coefficients per field and $\mathbf{B}_F\in\mathbb{R}^{F\times F}$ is a field--field correlation matrix ($F=53$). This retains conditional independence across spectral coefficients but couples fields within each coefficient.
  \item \textbf{Variable-group weights.} The GPLFR likelihood treats all output dimensions equally by default. However, different physical quantities differ in their predictability, so equal weighting may not allocate modelling capacity efficiently. To address this, we introduce a learned weight per variable group, where groups collect variables that we expect to have broadly similar predictability. The groups are just the variables except that ``winds'' collects both E--W and N--S winds, and ``radiation'' collects OLR and ASR. Each group's weight is broadcast to all spectral coefficients within its fields, scaling their contribution to the likelihood. To remove a global scale non-identifiability with the overall decoder scale, the weights are constrained to have unit geometric mean across groups. We place a Gaussian prior on the unconstrained log-weights and infer them jointly with the other model parameters. Predictions are mapped back to physical units by dividing out the weights before inverse preprocessing.
\end{enumerate}

\paragraph{Results.}
Table~\ref{tab:gplfr-di} reports RMSE for three configurations: vanilla GPLFR, field-coregionalization with fixed weights, and field-coregionalization with learned variable-group weights.
The effects are small: relative to vanilla GPLFR, field-coregionalization increases mean seed-0 RMSE by 0.6\% and energy score by 3.2\%, while the full extension with learned variable-group weights does manage to reduce RMSE by 1.2\% and energy score by 3.7\% (only cloud fraction gets worse).
However, these differences are notably smaller than the gap between GPLFR and the next-best baselines, suggesting that these structural relaxations offer little benefit at this dataset size and configuration.

\begin{table}[ht]
  \centering
  \small
  \setlength{\tabcolsep}{3pt}
  \renewcommand{\arraystretch}{1.1}
  \caption{The effect of domain-informed extensions to GPLFR on RMSE on the Multi-partial dataset. Scores are from training seed 0. The first column is from vanilla GPLFR like in the main text.}
  \label{tab:gplfr-di}
  \begin{tabular}{@{}>{\raggedright\arraybackslash}p{2.5cm} *{3}{>{\raggedleft\arraybackslash}p{2.2cm}}@{}}
    \toprule
    Variable & GPLFR & GPLFR + field-coreg. & GPLFR + field-coreg. + learn-weights \\
    \midrule
    Surface temp. (K)                      & 9.83 & 9.89 & 9.52 \\
    Temperature (K)                        & 8.19 & 8.24 & 7.99 \\
    Humidity (dex)                         & 0.429 & 0.430 & 0.417 \\
    Cloud fraction (1)                     & 0.0774 & 0.0778 & 0.0789 \\
    E--W wind (m\,s$^{-1}$)                & 9.45 & 9.58 & 9.41 \\
    N--S wind (m\,s$^{-1}$)                & 4.12 & 4.22 & 4.09 \\
    ASR (W\,m$^{-2}$)                      & 22.5 & 22.4 & 22.4 \\
    OLR (W\,m$^{-2}$)                      & 16.0 & 15.9 & 15.7 \\
    \bottomrule
  \end{tabular}
\end{table}

\section{Limitations}\label{app:limitations}

\paragraph{Scope.}
ThousandWorlds restricts attention to tidally locked waterworlds (\emph{aquaplanets}).
These are the most widely simulated subclass of potentially habitable exoplanets, but still only a slice of the broader parameter space (e.g., dry planets, eccentric orbits, asynchronous rotators are excluded).
Extending the benchmark to other planet classes is a clear avenue for future work.

\paragraph{GCMs.}
The auxiliary-GCM set consists primarily of ExoPlaSim (1,117 of 1,295 simulations; Table~\ref{tab:dataset-composition}), which is an intermediate-complexity GCM (Appendix~\ref{app:gcms}). These simulations are likely less useful for transfer than the same number of high-fidelity GCM runs would be, although they can still be beneficial for the right model (see Appendix~\ref{app:dataset-ablations}).
Our four high-fidelity GCMs share substantial heritage, particularly the UM with LFRic, and ExoCAM with ExoCAM-pre-2022 (see Appendix~\ref{app:gcms} for detail).
The shared-planets protocol evaluates one GCM from each lineage (the UM and ExoCAM), so it captures inter-lineage disagreement, but this is still only an estimate of the true epistemic uncertainty across exoplanet GCM space (e.g., as reported for a select planet by the THAI intercomparison; \citealp{sergeevTRAPPIST1Habitable2022}).

\paragraph{Test-set sizes.}
Test sets are small (up to 128 simulations for the largest subset, Multi-partial).
Fine-grained method comparisons should therefore be interpreted cautiously; see Appendix~\ref{app:rmse-bootstrap-intervals} for estimates of score uncertainty due to test-set composition.

\section{Additional results}\label{app:additional-results}

% BEGIN AUTO-GENERATED STD RESULT TABLES

\subsection{Result tables with seed variability}\label{app:tables-with-seed-variability}

Tables~\ref{tab:rmse-partial-splits-std} and~\ref{tab:relative-rmse-std} repeat the main RMSE and relative RMSE tables with standard deviations over random seeds.

\begin{table*}[p]
  \centering
  \scriptsize
  \setlength{\tabcolsep}{2pt}
  \renewcommand{\arraystretch}{1.1}
  \caption{RMSE by subset, variable, and method, reported as mean$\pm$standard deviation over random seeds. Deterministic baselines are shown without standard deviations. The main-text Table~\ref{tab:rmse-partial-splits} shows only the means.}
  \label{tab:rmse-partial-splits-std}
  \resizebox{\textwidth}{!}{%
  \begin{tabular}{@{}>{\centering\arraybackslash}m{0.5cm}|>{\raggedright\arraybackslash}p{1.95cm} *{2}{>{\raggedleft\arraybackslash}p{0.8cm}} *{8}{>{\raggedleft\arraybackslash}p{1.75cm}}@{}}
    \toprule
    \shortstack{Sub-\\set} & Variable & Train-mean & kNN & PCA-GBT & Coord-MLP & Coord-DeepONet & PCA-MLP & ConvDec & SFNO & PPCA-ICM & GPLFR \\
    \midrule
     & Surface temp. (K)                      & \rmseMM{10}{24.52} & \rmseMM{19}{18.79} & \rmseMM{54}{12.33$\pm$0.48} & \rmseMM{28}{17.65$\pm$1.55} & \rmseMM{46}{12.45$\pm$0.64} & \rmseMM{63}{12.24$\pm$0.30} & \rmseMM{37}{12.75$\pm$0.68} & \rmseMM{72}{12.24$\pm$0.62} & \rmseMM{81}{10.06$\pm$0.01} & \rmseMM{90}{\textbf{9.89$\pm$0.07}} \\
     & Temperature (K)                        & \rmseMM{10}{20.34} & \rmseMM{19}{14.66} & \rmseMM{54}{10.23$\pm$0.11} & \rmseMM{28}{11.94$\pm$0.97} & \rmseMM{46}{10.43$\pm$0.49} & \rmseMM{72}{10.08$\pm$0.16} & \rmseMM{63}{10.10$\pm$0.27} & \rmseMM{37}{10.52$\pm$0.21} & \rmseMM{81}{8.79$\pm$0.02} & \rmseMM{90}{\textbf{8.26$\pm$0.09}} \\
     & Humidity (dex)                         & \rmseMM{10}{1.119} & \rmseMM{19}{0.784} & \rmseMM{37}{0.536$\pm$0.013} & \rmseMM{28}{0.627$\pm$0.061} & \rmseMM{54}{0.527$\pm$0.026} & \rmseMM{72}{0.525$\pm$0.008} & \rmseMM{63}{0.525$\pm$0.012} & \rmseMM{46}{0.531$\pm$0.011} & \rmseMM{81}{0.446$\pm$0.002} & \rmseMM{90}{\textbf{0.432$\pm$0.004}} \\
    \smash{\rotatebox[origin=c]{90}{Multi-partial \textcolor{white}{aa}}}
     & Cloud fraction (1)                     & \rmseMM{10}{0.1277} & \rmseMM{28}{0.0998} & \rmseMM{54}{0.0879$\pm$0.0007} & \rmseMM{19}{0.1073$\pm$0.0024} & \rmseMM{46}{0.0897$\pm$0.0013} & \rmseMM{37}{0.0898$\pm$0.0009} & \rmseMM{81}{0.0816$\pm$0.0018} & \rmseMM{63}{0.0845$\pm$0.0012} & \rmseMM{72}{0.0836$\pm$0.0004} & \rmseMM{90}{\textbf{0.0778$\pm$0.0003}} \\
     & E--W wind (m\,s$^{-1}$)                & \rmseMM{10}{16.73} & \rmseMM{37}{10.92} & \rmseMM{54}{10.64$\pm$0.07} & \rmseMM{19}{12.12$\pm$0.28} & \rmseMM{46}{10.88$\pm$0.18} & \rmseMM{28}{11.19$\pm$0.04} & \rmseMM{72}{9.97$\pm$0.45} & \rmseMM{63}{10.46$\pm$0.25} & \rmseMM{81}{9.63$\pm$0.01} & \rmseMM{90}{\textbf{9.50$\pm$0.06}} \\
     & N--S wind (m\,s$^{-1}$)                & \rmseMM{10}{6.57} & \rmseMM{46}{4.72} & \rmseMM{37}{4.75$\pm$0.04} & \rmseMM{19}{5.13$\pm$0.32} & \rmseMM{28}{4.91$\pm$0.09} & \rmseMM{54}{4.67$\pm$0.03} & \rmseMM{90}{\textbf{4.08$\pm$0.13}} & \rmseMM{72}{4.18$\pm$0.07} & \rmseMM{63}{4.34$\pm{<}$0.01} & \rmseMM{81}{4.12$\pm$0.01} \\
     & ASR (W\,m$^{-2}$)                      & \rmseMM{19}{59.0} & \rmseMM{46}{33.2} & \rmseMM{63}{27.1$\pm$0.1} & \rmseMM{10}{77.1$\pm$9.7} & \rmseMM{54}{29.8$\pm$1.0} & \rmseMM{72}{26.2$\pm$0.7} & \rmseMM{28}{37.7$\pm$1.2} & \rmseMM{37}{33.5$\pm$1.0} & \rmseMM{81}{26.0$\pm$0.1} & \rmseMM{90}{\textbf{22.8$\pm$0.2}} \\
     & OLR (W\,m$^{-2}$)                      & \rmseMM{10}{41.0} & \rmseMM{28}{24.8} & \rmseMM{37}{19.5$\pm$0.4} & \rmseMM{19}{29.0$\pm$3.0} & \rmseMM{54}{18.9$\pm$0.2} & \rmseMM{63}{18.7$\pm$0.4} & \rmseMM{81}{18.0$\pm$0.7} & \rmseMM{46}{19.0$\pm$0.3} & \rmseMM{72}{18.2$\pm$0.1} & \rmseMM{90}{\textbf{16.1$\pm$0.1}} \\
    \arrayrulecolor{gray!50}\cmidrule(l){2-12}\arrayrulecolor{black}
     & Geometric mean                         & \rmseMM{10}{1.00} & \rmseMM{28}{0.685} & \rmseMM{54}{0.550$\pm$0.005} & \rmseMM{19}{0.752$\pm$0.023} & \rmseMM{37}{0.560$\pm$0.012} & \rmseMM{72}{0.546$\pm$0.006} & \rmseMM{63}{0.547$\pm$0.013} & \rmseMM{46}{0.551$\pm$0.007} & \rmseMM{81}{0.492$\pm$0.001} & \rmseMM{90}{\textbf{0.462$\pm$0.002}} \\
    \midrule
     & Surface temp. (K)                      & \rmseTR{10}{24.00} & \rmseTR{19}{19.58} & \rmseTR{46}{12.59$\pm$0.30} & \rmseTR{28}{16.66$\pm$0.78} & \rmseTR{72}{12.07$\pm$1.27} & \rmseTR{54}{12.36$\pm$0.23} & \rmseTR{37}{12.75$\pm$0.46} & \rmseTR{63}{12.29$\pm$0.72} & \rmseTR{81}{11.20$\pm$0.01} & \rmseTR{90}{\textbf{9.79$\pm$0.08}} \\
     & Temperature (K)                        & \rmseTR{10}{19.32} & \rmseTR{19}{15.22} & \rmseTR{54}{9.90$\pm$0.16} & \rmseTR{28}{11.44$\pm$1.15} & \rmseTR{63}{9.68$\pm$0.62} & \rmseTR{81}{9.47$\pm$0.36} & \rmseTR{46}{9.97$\pm$0.23} & \rmseTR{37}{10.40$\pm$0.40} & \rmseTR{72}{9.59$\pm{<}$0.01} & \rmseTR{90}{\textbf{8.14$\pm$0.04}} \\
     & Humidity (dex)                         & \rmseTR{10}{1.087} & \rmseTR{19}{0.765} & \rmseTR{54}{0.506$\pm$0.005} & \rmseTR{28}{0.620$\pm$0.029} & \rmseTR{72}{0.490$\pm$0.020} & \rmseTR{63}{0.506$\pm$0.009} & \rmseTR{46}{0.512$\pm$0.018} & \rmseTR{37}{0.513$\pm$0.035} & \rmseTR{81}{0.459$\pm{<}$0.001} & \rmseTR{90}{\textbf{0.404$\pm$0.002}} \\
    \smash{\rotatebox[origin=c]{90}{Multi-complete \textcolor{white}{aa}}}
     & Cloud fraction (1)                     & \rmseTR{10}{0.1356} & \rmseTR{28}{0.1074} & \rmseTR{54}{0.0913$\pm$0.0005} & \rmseTR{19}{0.1117$\pm$0.0029} & \rmseTR{37}{0.0968$\pm$0.0020} & \rmseTR{46}{0.0936$\pm$0.0005} & \rmseTR{81}{0.0865$\pm$0.0007} & \rmseTR{63}{0.0890$\pm$0.0013} & \rmseTR{72}{0.0878$\pm{<}$0.0001} & \rmseTR{90}{\textbf{0.0816$\pm$0.0003}} \\
     & E--W wind (m\,s$^{-1}$)                & \rmseTR{10}{15.25} & \rmseTR{19}{10.16} & \rmseTR{63}{8.92$\pm$0.13} & \rmseTR{28}{10.14$\pm$0.13} & \rmseTR{37}{9.63$\pm$0.27} & \rmseTR{46}{9.57$\pm$0.09} & \rmseTR{72}{8.64$\pm$0.17} & \rmseTR{54}{9.13$\pm$0.25} & \rmseTR{81}{8.63$\pm{<}$0.01} & \rmseTR{90}{\textbf{8.24$\pm$0.06}} \\
     & N--S wind (m\,s$^{-1}$)                & \rmseTR{10}{6.15} & \rmseTR{19}{4.90} & \rmseTR{54}{4.43$\pm$0.07} & \rmseTR{28}{4.89$\pm$0.12} & \rmseTR{37}{4.87$\pm$0.19} & \rmseTR{46}{4.66$\pm$0.04} & \rmseTR{90}{\textbf{4.00$\pm$0.09}} & \rmseTR{72}{4.12$\pm$0.15} & \rmseTR{63}{4.22$\pm{<}$0.01} & \rmseTR{81}{4.07$\pm$0.03} \\
     & ASR (W\,m$^{-2}$)                      & \rmseTR{19}{59.7} & \rmseTR{46}{31.5} & \rmseTR{63}{26.5$\pm$0.4} & \rmseTR{10}{68.6$\pm$7.7} & \rmseTR{54}{29.6$\pm$1.0} & \rmseTR{81}{25.5$\pm$0.5} & \rmseTR{28}{36.5$\pm$0.5} & \rmseTR{37}{31.7$\pm$1.7} & \rmseTR{72}{26.1$\pm{<}$0.1} & \rmseTR{90}{\textbf{22.1$\pm$0.1}} \\
     & OLR (W\,m$^{-2}$)                      & \rmseTR{10}{41.2} & \rmseTR{28}{23.8} & \rmseTR{37}{19.0$\pm$0.3} & \rmseTR{19}{26.2$\pm$0.9} & \rmseTR{63}{18.2$\pm$0.5} & \rmseTR{46}{18.5$\pm$0.2} & \rmseTR{81}{17.5$\pm$0.1} & \rmseTR{54}{18.3$\pm$0.7} & \rmseTR{72}{18.2$\pm{<}$0.1} & \rmseTR{90}{\textbf{15.5$\pm$0.1}} \\
    \arrayrulecolor{gray!50}\cmidrule(l){2-12}\arrayrulecolor{black}
     & Geometric mean                         & \rmseTR{10}{1.000} & \rmseTR{28}{0.700} & \rmseTR{72}{0.540$\pm$0.004} & \rmseTR{19}{0.722$\pm$0.025} & \rmseTR{37}{0.554$\pm$0.018} & \rmseTR{63}{0.542$\pm$0.006} & \rmseTR{54}{0.546$\pm$0.007} & \rmseTR{46}{0.548$\pm$0.012} & \rmseTR{81}{0.513$\pm{<}$0.001} & \rmseTR{90}{\textbf{0.458$\pm$0.002}} \\
    \midrule
     & Surface temp. (K)                      & \rmseSG{10}{20.85} & \rmseSG{28}{14.04} & \rmseSG{46}{11.63$\pm$0.41} & \rmseSG{19}{14.76$\pm$1.17} & \rmseSG{37}{11.94$\pm$0.40} & \rmseSG{54}{11.40$\pm$0.23} & \rmseSG{63}{10.72$\pm$0.28} & \rmseSG{72}{10.04$\pm$0.44} & \rmseSG{81}{9.65$\pm{<}$0.01} & \rmseSG{90}{\textbf{9.38$\pm$0.10}} \\
     & Temperature (K)                        & \rmseSG{10}{18.71} & \rmseSG{19}{11.71} & \rmseSG{28}{9.88$\pm$0.09} & \rmseSG{37}{9.85$\pm$1.00} & \rmseSG{46}{9.75$\pm$0.30} & \rmseSG{54}{9.72$\pm$0.09} & \rmseSG{63}{8.64$\pm$0.19} & \rmseSG{72}{8.55$\pm$0.08} & \rmseSG{81}{8.33$\pm{<}$0.01} & \rmseSG{90}{\textbf{8.08$\pm$0.07}} \\
     & Humidity (dex)                         & \rmseSG{10}{1.067} & \rmseSG{19}{0.583} & \rmseSG{37}{0.555$\pm$0.016} & \rmseSG{28}{0.582$\pm$0.048} & \rmseSG{46}{0.549$\pm$0.024} & \rmseSG{54}{0.523$\pm$0.006} & \rmseSG{63}{0.477$\pm$0.013} & \rmseSG{72}{0.472$\pm$0.009} & \rmseSG{81}{0.455$\pm{<}$0.001} & \rmseSG{90}{\textbf{0.447$\pm$0.003}} \\
    \smash{\rotatebox[origin=c]{90}{Single-complete \textcolor{white}{aa}}}
     & Cloud fraction (1)                     & \rmseSG{10}{0.1287} & \rmseSG{28}{0.0902} & \rmseSG{54}{0.0866$\pm$0.0010} & \rmseSG{19}{0.1021$\pm$0.0035} & \rmseSG{46}{0.0896$\pm$0.0016} & \rmseSG{37}{0.0900$\pm$0.0007} & \rmseSG{90}{\textbf{0.0761$\pm$0.0004}} & \rmseSG{72}{0.0809$\pm$0.0013} & \rmseSG{63}{0.0816$\pm{<}$0.0001} & \rmseSG{81}{0.0767$\pm$0.0003} \\
     & E--W wind (m\,s$^{-1}$)                & \rmseSG{10}{12.75} & \rmseSG{54}{8.53} & \rmseSG{46}{8.54$\pm$0.12} & \rmseSG{37}{9.01$\pm$0.27} & \rmseSG{19}{9.55$\pm$0.66} & \rmseSG{28}{9.39$\pm$0.09} & \rmseSG{90}{\textbf{7.34$\pm$0.05}} & \rmseSG{63}{8.16$\pm$0.14} & \rmseSG{72}{8.14$\pm{<}$0.01} & \rmseSG{81}{7.49$\pm$0.07} \\
     & N--S wind (m\,s$^{-1}$)                & \rmseSG{10}{5.16} & \rmseSG{63}{3.82} & \rmseSG{46}{3.95$\pm$0.04} & \rmseSG{37}{4.20$\pm$0.23} & \rmseSG{19}{4.23$\pm$0.18} & \rmseSG{28}{4.22$\pm$0.03} & \rmseSG{90}{\textbf{3.36$\pm$0.05}} & \rmseSG{72}{3.59$\pm$0.05} & \rmseSG{54}{3.95$\pm{<}$0.01} & \rmseSG{81}{3.55$\pm$0.04} \\
     & ASR (W\,m$^{-2}$)                      & \rmseSG{19}{45.6} & \rmseSG{54}{28.6} & \rmseSG{63}{28.0$\pm$0.4} & \rmseSG{10}{67.0$\pm$7.4} & \rmseSG{46}{30.5$\pm$0.5} & \rmseSG{72}{26.0$\pm$0.4} & \rmseSG{28}{36.3$\pm$1.3} & \rmseSG{37}{32.1$\pm$2.4} & \rmseSG{90}{\textbf{24.4$\pm{<}$0.1}} & \rmseSG{81}{25.1$\pm$0.4} \\
     & OLR (W\,m$^{-2}$)                      & \rmseSG{10}{33.7} & \rmseSG{28}{20.4} & \rmseSG{37}{20.0$\pm$0.3} & \rmseSG{19}{26.3$\pm$1.2} & \rmseSG{54}{19.0$\pm$0.3} & \rmseSG{46}{19.0$\pm$0.3} & \rmseSG{63}{17.7$\pm$0.5} & \rmseSG{72}{17.1$\pm$0.4} & \rmseSG{90}{\textbf{16.2$\pm{<}$0.1}} & \rmseSG{81}{16.3$\pm$0.2} \\
    \arrayrulecolor{gray!50}\cmidrule(l){2-12}\arrayrulecolor{black}
     & Geometric mean                         & \rmseSG{10}{1.000} & \rmseSG{28}{0.646} & \rmseSG{46}{0.610$\pm$0.008} & \rmseSG{19}{0.755$\pm$0.026} & \rmseSG{37}{0.630$\pm$0.015} & \rmseSG{54}{0.609$\pm$0.004} & \rmseSG{63}{0.561$\pm$0.007} & \rmseSG{72}{0.560$\pm$0.010} & \rmseSG{81}{0.538$\pm{<}$0.001} & \rmseSG{90}{\textbf{0.518$\pm$0.002}} \\
    \bottomrule
  \end{tabular}
  }
\end{table*}

\begin{table*}[p]
  \centering
  \scriptsize
  \setlength{\tabcolsep}{2pt}
  \renewcommand{\arraystretch}{1.1}
  \caption{Relative RMSE by variable under the shared-planets protocol, reported as mean$\pm$standard deviation over random seeds. Deterministic baselines are shown without standard deviations. The main-text Table~\ref{tab:relative-rmse} shows only the means.}
  \label{tab:relative-rmse-std}
  \resizebox{\textwidth}{!}{%
  \begin{tabular}{@{}>{\centering\arraybackslash}m{0.5cm}|>{\raggedright\arraybackslash}p{1.95cm} *{2}{>{\raggedleft\arraybackslash}p{0.8cm}} *{8}{>{\raggedleft\arraybackslash}p{1.55cm}}@{}}
    \toprule
    \shortstack{Sub-\\set} & Variable & Train-mean & kNN & PCA-GBT & Coord-MLP & Coord-DeepONet & PCA-MLP & ConvDec & SFNO & PPCA-ICM & GPLFR \\
    \midrule
     & Surface temp.                      & \rmseMM{10}{1.680} & \rmseMM{28}{1.283} & \rmseMM{46}{1.039$\pm$0.113} & \rmseMM{19}{1.380$\pm$0.180} & \rmseMM{63}{1.005$\pm$0.113} & \rmseMM{37}{1.057$\pm$0.059} & \rmseMM{54}{1.007$\pm$0.048} & \rmseMM{72}{0.975$\pm$0.072} & \rmseMM{81}{0.927$\pm$0.002} & \rmseMM{90}{\textbf{0.739$\pm$0.005}} \\
     & Temperature                        & \rmseMM{10}{1.655} & \rmseMM{19}{1.209} & \rmseMM{46}{0.925$\pm$0.037} & \rmseMM{28}{0.998$\pm$0.073} & \rmseMM{37}{0.949$\pm$0.079} & \rmseMM{72}{0.902$\pm$0.033} & \rmseMM{63}{0.903$\pm$0.030} & \rmseMM{54}{0.921$\pm$0.029} & \rmseMM{81}{0.829$\pm$0.002} & \rmseMM{90}{\textbf{0.664$\pm$0.008}} \\
     & Humidity                         & \rmseMM{10}{1.515} & \rmseMM{19}{1.247} & \rmseMM{37}{0.896$\pm$0.032} & \rmseMM{28}{1.008$\pm$0.117} & \rmseMM{46}{0.879$\pm$0.100} & \rmseMM{72}{0.818$\pm$0.038} & \rmseMM{54}{0.851$\pm$0.011} & \rmseMM{63}{0.824$\pm$0.012} & \rmseMM{81}{0.787$\pm$0.008} & \rmseMM{90}{\textbf{0.670$\pm$0.004}} \\
    \smash{\rotatebox[origin=c]{90}{Multi-partial}}
     & Cloud fraction                     & \rmseMM{10}{0.867} & \rmseMM{19}{0.752} & \rmseMM{46}{0.644$\pm$0.004} & \rmseMM{28}{0.733$\pm$0.015} & \rmseMM{37}{0.653$\pm$0.017} & \rmseMM{54}{0.639$\pm$0.004} & \rmseMM{81}{0.595$\pm$0.008} & \rmseMM{72}{0.606$\pm$0.008} & \rmseMM{63}{0.624$\pm$0.004} & \rmseMM{90}{\textbf{0.562$\pm$0.004}} \\
     & E--W wind                & \rmseMM{10}{1.265} & \rmseMM{63}{0.832} & \rmseMM{46}{0.897$\pm$0.012} & \rmseMM{19}{0.937$\pm$0.010} & \rmseMM{28}{0.918$\pm$0.039} & \rmseMM{37}{0.909$\pm$0.019} & \rmseMM{90}{\textbf{0.757$\pm$0.013}} & \rmseMM{54}{0.833$\pm$0.028} & \rmseMM{72}{0.785$\pm$0.001} & \rmseMM{81}{0.765$\pm$0.007} \\
     & N--S wind                & \rmseMM{10}{1.205} & \rmseMM{54}{0.828} & \rmseMM{37}{0.885$\pm$0.015} & \rmseMM{28}{0.898$\pm$0.038} & \rmseMM{19}{0.919$\pm$0.022} & \rmseMM{46}{0.837$\pm$0.009} & \rmseMM{90}{\textbf{0.714$\pm$0.010}} & \rmseMM{81}{0.746$\pm$0.014} & \rmseMM{63}{0.802$\pm{<}$0.001} & \rmseMM{72}{0.749$\pm$0.005} \\
     & ASR                      & \rmseMM{19}{1.342} & \rmseMM{46}{0.884} & \rmseMM{81}{0.699$\pm$0.015} & \rmseMM{10}{2.007$\pm$0.237} & \rmseMM{54}{0.794$\pm$0.042} & \rmseMM{72}{0.699$\pm$0.027} & \rmseMM{28}{1.044$\pm$0.030} & \rmseMM{37}{0.920$\pm$0.047} & \rmseMM{63}{0.711$\pm$0.002} & \rmseMM{90}{\textbf{0.594$\pm$0.011}} \\
     & OLR                      & \rmseMM{10}{1.571} & \rmseMM{28}{1.023} & \rmseMM{54}{0.775$\pm$0.023} & \rmseMM{19}{1.218$\pm$0.126} & \rmseMM{46}{0.776$\pm$0.022} & \rmseMM{63}{0.767$\pm$0.034} & \rmseMM{81}{0.741$\pm$0.014} & \rmseMM{72}{0.765$\pm$0.024} & \rmseMM{37}{0.776$\pm$0.004} & \rmseMM{90}{\textbf{0.640$\pm$0.007}} \\
    \arrayrulecolor{gray!50}\cmidrule(l){2-12}\arrayrulecolor{black}
      & Geometric mean                       & \rmseMM{10}{1.360} & \rmseMM{28}{0.988} & \rmseMM{46}{0.836$\pm$0.019} & \rmseMM{19}{1.094$\pm$0.029} & \rmseMM{37}{0.853$\pm$0.031} & \rmseMM{54}{0.819$\pm$0.016} & \rmseMM{72}{0.813$\pm$0.007} & \rmseMM{63}{0.815$\pm$0.010} & \rmseMM{81}{0.776$\pm$0.001} & \rmseMM{90}{\textbf{0.669$\pm$0.003}} \\
    \midrule
     & Surface temp.                      & \rmseTR{10}{1.709} & \rmseTR{19}{1.445} & \rmseTR{54}{1.011$\pm$0.071} & \rmseTR{28}{1.298$\pm$0.098} & \rmseTR{63}{1.009$\pm$0.104} & \rmseTR{72}{1.009$\pm$0.049} & \rmseTR{46}{1.026$\pm$0.050} & \rmseTR{81}{0.979$\pm$0.130} & \rmseTR{37}{1.039$\pm$0.001} & \rmseTR{90}{\textbf{0.788$\pm$0.013}} \\
     & Temperature                        & \rmseTR{10}{1.558} & \rmseTR{19}{1.317} & \rmseTR{72}{0.874$\pm$0.020} & \rmseTR{28}{0.947$\pm$0.141} & \rmseTR{54}{0.912$\pm$0.095} & \rmseTR{81}{0.850$\pm$0.059} & \rmseTR{63}{0.888$\pm$0.017} & \rmseTR{46}{0.924$\pm$0.044} & \rmseTR{37}{0.935$\pm$0.001} & \rmseTR{90}{\textbf{0.687$\pm$0.007}} \\
     & Humidity                         & \rmseTR{10}{1.534} & \rmseTR{19}{1.274} & \rmseTR{72}{0.837$\pm$0.018} & \rmseTR{28}{1.042$\pm$0.100} & \rmseTR{54}{0.857$\pm$0.046} & \rmseTR{81}{0.825$\pm$0.041} & \rmseTR{37}{0.874$\pm$0.035} & \rmseTR{46}{0.863$\pm$0.074} & \rmseTR{63}{0.853$\pm{<}$0.001} & \rmseTR{90}{\textbf{0.694$\pm$0.006}} \\
    \smash{\rotatebox[origin=c]{90}{Multi-complete}}
     & Cloud fraction                     & \rmseTR{10}{0.861} & \rmseTR{19}{0.748} & \rmseTR{46}{0.635$\pm$0.004} & \rmseTR{28}{0.721$\pm$0.017} & \rmseTR{37}{0.665$\pm$0.014} & \rmseTR{54}{0.623$\pm$0.005} & \rmseTR{81}{0.605$\pm$0.009} & \rmseTR{72}{0.610$\pm$0.016} & \rmseTR{63}{0.615$\pm{<}$0.001} & \rmseTR{90}{\textbf{0.566$\pm$0.001}} \\
     & E--W wind                & \rmseTR{10}{1.377} & \rmseTR{19}{0.974} & \rmseTR{54}{0.863$\pm$0.018} & \rmseTR{37}{0.948$\pm$0.020} & \rmseTR{28}{0.956$\pm$0.035} & \rmseTR{46}{0.940$\pm$0.017} & \rmseTR{81}{0.801$\pm$0.018} & \rmseTR{63}{0.855$\pm$0.035} & \rmseTR{72}{0.854$\pm$0.001} & \rmseTR{90}{\textbf{0.792$\pm$0.012}} \\
     & N--S wind                & \rmseTR{10}{1.164} & \rmseTR{37}{0.930} & \rmseTR{54}{0.851$\pm$0.016} & \rmseTR{28}{0.938$\pm$0.011} & \rmseTR{19}{0.963$\pm$0.052} & \rmseTR{46}{0.914$\pm$0.010} & \rmseTR{90}{\textbf{0.767$\pm$0.015}} & \rmseTR{81}{0.773$\pm$0.021} & \rmseTR{63}{0.812$\pm{<}$0.001} & \rmseTR{72}{0.779$\pm$0.009} \\
     & ASR                      & \rmseTR{19}{1.355} & \rmseTR{46}{0.853} & \rmseTR{72}{0.698$\pm$0.017} & \rmseTR{10}{1.808$\pm$0.238} & \rmseTR{54}{0.809$\pm$0.050} & \rmseTR{81}{0.686$\pm$0.016} & \rmseTR{28}{1.000$\pm$0.065} & \rmseTR{37}{0.873$\pm$0.062} & \rmseTR{63}{0.737$\pm{<}$0.001} & \rmseTR{90}{\textbf{0.600$\pm$0.004}} \\
     & OLR                      & \rmseTR{10}{1.579} & \rmseTR{28}{0.974} & \rmseTR{63}{0.763$\pm$0.011} & \rmseTR{19}{1.087$\pm$0.044} & \rmseTR{46}{0.782$\pm$0.048} & \rmseTR{72}{0.761$\pm$0.018} & \rmseTR{81}{0.741$\pm$0.014} & \rmseTR{54}{0.767$\pm$0.044} & \rmseTR{37}{0.797$\pm$0.001} & \rmseTR{90}{\textbf{0.641$\pm$0.006}} \\
    \arrayrulecolor{gray!50}\cmidrule(l){2-12}\arrayrulecolor{black}
      & Geometric mean                       & \rmseTR{10}{1.365} & \rmseTR{28}{1.040} & \rmseTR{81}{0.809$\pm$0.014} & \rmseTR{19}{1.059$\pm$0.049} & \rmseTR{37}{0.861$\pm$0.036} & \rmseTR{72}{0.816$\pm$0.019} & \rmseTR{46}{0.827$\pm$0.015} & \rmseTR{54}{0.822$\pm$0.029} & \rmseTR{63}{0.822$\pm{<}$0.001} & \rmseTR{90}{\textbf{0.688$\pm$0.006}} \\
    \bottomrule
  \end{tabular}
  }
\end{table*}

% END AUTO-GENERATED STD RESULT TABLES

\subsection{Bootstrap intervals for Multi-partial RMSE results}\label{app:rmse-bootstrap-intervals}

Table~\ref{tab:rmse-intervals} reports 95\% bootstrap confidence intervals on the Multi-partial RMSE results for training seed 0.
The intervals are notably wider than the training-seed standard deviations in Appendix~\ref{app:tables-with-seed-variability}, indicating that test-set composition is the dominant source of uncertainty in the RMSE numbers.

\begin{table*}[htbp]
  \centering
  \scriptsize
  \setlength{\tabcolsep}{0pt}
  \renewcommand{\arraystretch}{1.5}
  \newcommand{\asymci}[3]{\ensuremath{#1^{+#2}_{-#3}}}
  \caption{RMSE on the Multi-partial subset for predictions from training seed 0, reported with 95\% bootstrap intervals from 1,000 resamples of the 128 test examples. Each resample draws 128 examples with replacement and is applied to every method jointly. Lower is better.}
  \label{tab:rmse-intervals}
  \resizebox{\textwidth}{!}{%
  \begin{tabular}{@{}>{\raggedright\arraybackslash}p{0.8cm} *{11}{>{\raggedleft\arraybackslash}p{1.8cm}}@{}}
    \toprule
    Variable & \shortstack{Train-mean} & kNN & PCA-Ridge & PCA-GBT & PCA-MLP & Coord-MLP & \shortstack{Coord-DeepONet} & ConvDec & SFNO & PPCA-ICM & GPLFR \\
    \midrule
    Surface temp. (K) & \asymci{24.52}{3.55}{3.08} & \asymci{18.79}{3.15}{2.72} & \asymci{16.53}{1.82}{1.73} & \asymci{11.87}{1.74}{1.57} & \asymci{12.40}{1.83}{1.71} & \asymci{17.18}{1.84}{1.72} & \asymci{12.80}{2.27}{2.07} & \asymci{12.60}{1.62}{1.46} & \asymci{13.20}{1.85}{1.66} & \asymci{10.09}{1.75}{1.55} & \asymci{9.83}{1.59}{1.50} \\
    Temp. (K) & \asymci{20.34}{2.46}{2.07} & \asymci{14.66}{2.25}{1.90} & \asymci{14.19}{1.70}{1.38} & \asymci{10.19}{1.44}{1.30} & \asymci{10.10}{1.48}{1.22} & \asymci{12.65}{1.26}{1.14} & \asymci{10.70}{1.59}{1.42} & \asymci{10.00}{1.17}{1.10} & \asymci{10.80}{1.36}{1.22} & \asymci{8.82}{1.39}{1.21} & \asymci{8.19}{1.27}{1.04} \\
    Humidity (dex) & \asymci{1.119}{0.108}{0.101} & \asymci{0.784}{0.114}{0.096} & \asymci{0.689}{0.069}{0.072} & \asymci{0.537}{0.074}{0.060} & \asymci{0.528}{0.081}{0.067} & \asymci{0.657}{0.060}{0.056} & \asymci{0.530}{0.080}{0.072} & \asymci{0.515}{0.066}{0.057} & \asymci{0.531}{0.072}{0.062} & \asymci{0.443}{0.072}{0.061} & \asymci{0.429}{0.071}{0.059} \\
    Cloud fraction (1) & \asymci{0.1277}{0.0059}{0.0058} & \asymci{0.0998}{0.0088}{0.0081} & \asymci{0.1069}{0.0049}{0.0051} & \asymci{0.0888}{0.0054}{0.0056} & \asymci{0.0903}{0.0061}{0.0057} & \asymci{0.1051}{0.0049}{0.0050} & \asymci{0.0907}{0.0063}{0.0060} & \asymci{0.0822}{0.0061}{0.0062} & \asymci{0.0851}{0.0058}{0.0058} & \asymci{0.0839}{0.0071}{0.0069} & \asymci{0.0774}{0.0056}{0.0054} \\
    E--W wind (m\,s$^{-1}$) & \asymci{16.73}{1.28}{1.24} & \asymci{10.92}{1.36}{1.31} & \asymci{13.04}{1.02}{0.95} & \asymci{10.68}{1.15}{0.96} & \asymci{11.20}{1.14}{1.10} & \asymci{12.22}{0.99}{0.99} & \asymci{10.90}{1.20}{1.15} & \asymci{9.90}{1.19}{1.17} & \asymci{10.60}{1.07}{1.01} & \asymci{9.63}{1.20}{1.17} & \asymci{9.45}{1.12}{1.01} \\
    N--S wind (m\,s$^{-1}$) & \asymci{6.57}{0.41}{0.38} & \asymci{4.72}{0.50}{0.51} & \asymci{5.36}{0.37}{0.37} & \asymci{4.81}{0.41}{0.47} & \asymci{4.65}{0.40}{0.35} & \asymci{4.96}{0.38}{0.38} & \asymci{4.99}{0.57}{0.48} & \asymci{3.91}{0.41}{0.39} & \asymci{4.28}{0.38}{0.37} & \asymci{4.33}{0.42}{0.41} & \asymci{4.12}{0.38}{0.37} \\
    ASR (W\,m$^{-2}$) & \asymci{59.0}{5.0}{5.2} & \asymci{33.2}{4.8}{4.5} & \asymci{32.4}{4.5}{3.8} & \asymci{27.2}{4.2}{3.6} & \asymci{26.9}{4.1}{3.5} & \asymci{77.1}{4.5}{4.1} & \asymci{29.7}{4.2}{4.0} & \asymci{37.1}{3.9}{3.7} & \asymci{32.4}{4.0}{3.7} & \asymci{26.0}{4.2}{3.7} & \asymci{22.5}{4.1}{3.6} \\
    OLR (W\,m$^{-2}$) & \asymci{41.0}{3.0}{2.8} & \asymci{24.8}{3.2}{3.0} & \asymci{24.1}{2.3}{2.1} & \asymci{19.2}{2.2}{2.1} & \asymci{19.1}{2.2}{2.1} & \asymci{34.3}{2.8}{2.5} & \asymci{19.0}{2.4}{2.4} & \asymci{17.4}{2.1}{2.1} & \asymci{18.7}{2.3}{2.3} & \asymci{18.0}{2.5}{2.2} & \asymci{16.0}{2.3}{2.2} \\
    \bottomrule
  \end{tabular}
  }
\end{table*}

A paired bootstrap test shows the method ranking more clearly under this uncertainty: within each resample, we compare the geometric-mean RMSE across the eight variables for each adjacently ranked pair of methods (Table~\ref{tab:ordering-bootstrap}).
GPLFR's lead is robust, ahead of PPCA-ICM in 99.5\% of resamples, and PPCA-ICM is ahead of ConvDec in every resample.
The methods from ConvDec down to Coord-DeepONet are closely matched, with adjacent orderings reversing in 22--42\% of resamples.
Coord-DeepONet is then clearly ahead of kNN, which is in turn clearly ahead of Coord-MLP, the weakest learned method.

\begin{table}[htbp]
  \centering
  \small
  \renewcommand{\arraystretch}{1.05}
  \caption{Paired bootstrap test of the seed-0 Multi-partial RMSE ranking. Each row compares adjacent methods using the same 1,000 joint resamples as Table~\ref{tab:rmse-intervals}. We show the ratio of the lower-ranked to the higher-ranked method's geometric-mean RMSE, its 95\% interval, and the fraction of resamples in which the ordering holds. (The closely matched middle-performance methods change order relative to the five-seed means in Table~\ref{tab:rmse-partial-splits}.)}
  \label{tab:ordering-bootstrap}
  \begin{tabular}{@{}llrcr@{}}
    \toprule
    Higher-ranked & Lower-ranked & Ratio & 95\% interval & Holds (\%) \\
    \midrule
    GPLFR & PPCA-ICM & 1.070 & [1.018, 1.128] & 99.5 \\
    PPCA-ICM & ConvDec & 1.095 & [1.035, 1.159] & 100.0 \\
    ConvDec & PCA-GBT & 1.018 & [0.972, 1.062] & 77.5 \\
    PCA-GBT & PCA-MLP & 1.005 & [0.959, 1.059] & 58.0 \\
    PCA-MLP & SFNO & 1.012 & [0.976, 1.049] & 72.9 \\
    SFNO & Coord-DeepONet & 1.018 & [0.972, 1.065] & 77.2 \\
    Coord-DeepONet & kNN & 1.207 & [1.102, 1.330] & 100.0 \\
    kNN & Coord-MLP & 1.128 & [1.027, 1.234] & 99.8 \\
    \bottomrule
  \end{tabular}
\end{table}

\subsection{Anomaly correlation coefficient (ACC) results}\label{app:acc-results}

ACC measures agreement in spatial structure between prediction $\hat{\mathbf{y}}$ and truth $\mathbf{y}$, ignoring differences in global mean and amplitude.
For a single field,
\begin{equation*}
  \mathrm{ACC}(\hat{\mathbf{y}},\mathbf{y})=
  \frac{\langle\hat{\mathbf{y}}^\circ,\mathbf{y}^\circ\rangle_G}
  {\|\hat{\mathbf{y}}^\circ\|_G\,\|\mathbf{y}^\circ\|_G},
  \quad
  \mathbf{u}^\circ\equiv\mathbf{u}-\bar{u}\mathbf{1},
  \quad
  \bar{u}=\langle\mathbf{1},\mathbf{u}\rangle_G/\langle\mathbf{1},\mathbf{1}\rangle_G,
\end{equation*}
where $\langle\mathbf{u},\mathbf{v}\rangle_G\equiv\mathbf{u}^\top\mathbf{G}\mathbf{v}$ is the area-weighted inner product.
ACC $=1$ indicates perfect spatial correlation; for a random prediction, $\mathbb{E}[\mathrm{ACC}]=0$.

GPLFR achieves the highest ACC on average across variables and subsets, followed by PPCA-ICM, kNN, PCA-GBT, and SFNO at roughly similar levels (Table~\ref{tab:acc-all-splits}).
kNN's relative strength on ACC, particularly winds and humidity, compared to RMSE suggests that spatial patterns are more locally consistent in input space than global means and amplitudes.
Cloud fraction, N--S wind, and temperature have the most challenging spatial patterns as measured by ACC.
All models achieve very high ACC for ASR, reflecting the strong dependence of this field on substellar point geometry, which is consistent across these tidally locked planets.

\begin{table}[htbp]
  \centering
  \small
  \setlength{\tabcolsep}{1.5pt}
  \renewcommand{\arraystretch}{1.1}
  \caption{Anomaly correlation coefficient (ACC) by variable and dataset subset. Higher is better; maximum is 1. Values are the mean of five training seeds (one deterministic run for Train-mean and kNN).}
  \label{tab:acc-all-splits}
  \resizebox{\linewidth}{!}{%
  \begin{tabular}{@{}>{\centering\arraybackslash}m{0.65cm}|>{\raggedright\arraybackslash}p{2.2cm} >{\raggedleft\arraybackslash}p{1.0cm} >{\raggedleft\arraybackslash}p{1.05cm} *{8}{>{\raggedleft\arraybackslash}p{1.32cm}}@{}}
    \toprule
    \shortstack{Sub-\\set} & Variable & Train-mean & kNN & PCA-GBT & Coord-MLP & Coord-DeepONet & PCA-MLP & ConvDec & SFNO & PPCA-ICM & GPLFR \\
    \midrule
     & Surface temp.                      & \rmseMM{28}{0.941} & \rmseMM{72}{0.968} & \rmseMM{63}{0.963} & \rmseMM{10}{0.907} & \rmseMM{46}{0.947} & \rmseMM{37}{0.946} & \rmseMM{19}{0.927} & \rmseMM{54}{0.955} & \rmseMM{81}{0.971} & \rmseMM{90}{\textbf{0.973}} \\
     & Temperature                        & \rmseMM{10}{0.400} & \rmseMM{81}{0.634} & \rmseMM{54}{0.603} & \rmseMM{19}{0.405} & \rmseMM{46}{0.574} & \rmseMM{72}{0.631} & \rmseMM{28}{0.490} & \rmseMM{37}{0.524} & \rmseMM{63}{0.629} & \rmseMM{90}{\textbf{0.687}} \\
     & Humidity                         & \rmseMM{28}{0.682} & \rmseMM{81}{0.780} & \rmseMM{54}{0.753} & \rmseMM{10}{0.656} & \rmseMM{46}{0.732} & \rmseMM{63}{0.761} & \rmseMM{19}{0.666} & \rmseMM{37}{0.708} & \rmseMM{72}{0.775} & \rmseMM{90}{\textbf{0.792}} \\
    \smash{\rotatebox[origin=c]{90}{Multi-partial \textcolor{white}{aa}}}
     & Cloud fraction                     & \rmseMM{10}{0.535} & \rmseMM{28}{0.635} & \rmseMM{37}{0.664} & \rmseMM{19}{0.568} & \rmseMM{46}{0.665} & \rmseMM{54}{0.676} & \rmseMM{72}{0.690} & \rmseMM{63}{0.687} & \rmseMM{81}{0.691} & \rmseMM{90}{\textbf{0.712}} \\
     & E--W wind                & \rmseMM{10}{0.528} & \rmseMM{54}{0.670} & \rmseMM{37}{0.654} & \rmseMM{19}{0.583} & \rmseMM{28}{0.646} & \rmseMM{46}{0.666} & \rmseMM{63}{0.679} & \rmseMM{72}{0.687} & \rmseMM{81}{0.696} & \rmseMM{90}{\textbf{0.728}} \\
     & N--S wind                & \rmseMM{10}{0.544} & \rmseMM{54}{0.665} & \rmseMM{37}{0.646} & \rmseMM{19}{0.609} & \rmseMM{28}{0.627} & \rmseMM{46}{0.655} & \rmseMM{72}{0.688} & \rmseMM{81}{0.689} & \rmseMM{63}{0.672} & \rmseMM{90}{\textbf{0.699}} \\
     & ASR                      & \rmseMM{28}{0.994} & \rmseMM{37}{0.995} & \rmseMM{63}{0.996} & \rmseMM{10}{0.969} & \rmseMM{54}{0.995} & \rmseMM{81}{0.996} & \rmseMM{19}{0.993} & \rmseMM{46}{0.995} & \rmseMM{72}{0.996} & \rmseMM{90}{\textbf{0.997}} \\
     & OLR                      & \rmseMM{19}{0.712} & \rmseMM{28}{0.815} & \rmseMM{46}{0.855} & \rmseMM{10}{0.706} & \rmseMM{63}{0.857} & \rmseMM{37}{0.853} & \rmseMM{72}{0.868} & \rmseMM{54}{0.857} & \rmseMM{81}{0.871} & \rmseMM{90}{\textbf{0.888}} \\
    \midrule
     & Surface temp.                      & \rmseTR{46}{0.941} & \rmseTR{72}{0.965} & \rmseTR{63}{0.962} & \rmseTR{10}{0.918} & \rmseTR{28}{0.933} & \rmseTR{37}{0.938} & \rmseTR{19}{0.933} & \rmseTR{54}{0.953} & \rmseTR{90}{\textbf{0.973}} & \rmseTR{81}{0.972} \\
     & Temperature                        & \rmseTR{10}{0.422} & \rmseTR{54}{0.634} & \rmseTR{63}{0.645} & \rmseTR{19}{0.477} & \rmseTR{46}{0.590} & \rmseTR{72}{0.657} & \rmseTR{28}{0.508} & \rmseTR{37}{0.551} & \rmseTR{81}{0.663} & \rmseTR{90}{\textbf{0.700}} \\
     & Humidity                         & \rmseTR{28}{0.727} & \rmseTR{72}{0.814} & \rmseTR{63}{0.810} & \rmseTR{10}{0.708} & \rmseTR{46}{0.783} & \rmseTR{54}{0.804} & \rmseTR{19}{0.722} & \rmseTR{37}{0.762} & \rmseTR{81}{0.835} & \rmseTR{90}{\textbf{0.848}} \\
    \smash{\rotatebox[origin=c]{90}{Multi-complete \textcolor{white}{aa}}}
     & Cloud fraction                     & \rmseTR{10}{0.579} & \rmseTR{28}{0.678} & \rmseTR{46}{0.719} & \rmseTR{19}{0.643} & \rmseTR{37}{0.702} & \rmseTR{54}{0.725} & \rmseTR{63}{0.730} & \rmseTR{72}{0.732} & \rmseTR{81}{0.736} & \rmseTR{90}{\textbf{0.756}} \\
     & E--W wind                & \rmseTR{10}{0.537} & \rmseTR{28}{0.657} & \rmseTR{54}{0.683} & \rmseTR{19}{0.606} & \rmseTR{46}{0.674} & \rmseTR{37}{0.672} & \rmseTR{63}{0.705} & \rmseTR{72}{0.710} & \rmseTR{81}{0.723} & \rmseTR{90}{\textbf{0.742}} \\
     & N--S wind                & \rmseTR{10}{0.555} & \rmseTR{37}{0.647} & \rmseTR{54}{0.654} & \rmseTR{19}{0.632} & \rmseTR{28}{0.646} & \rmseTR{46}{0.650} & \rmseTR{81}{0.698} & \rmseTR{72}{0.691} & \rmseTR{63}{0.684} & \rmseTR{90}{\textbf{0.703}} \\
     & ASR                      & \rmseTR{28}{0.994} & \rmseTR{37}{0.995} & \rmseTR{81}{0.996} & \rmseTR{10}{0.973} & \rmseTR{54}{0.995} & \rmseTR{63}{0.996} & \rmseTR{19}{0.993} & \rmseTR{46}{0.995} & \rmseTR{72}{0.996} & \rmseTR{90}{\textbf{0.997}} \\
     & OLR                      & \rmseTR{10}{0.714} & \rmseTR{28}{0.821} & \rmseTR{54}{0.854} & \rmseTR{19}{0.741} & \rmseTR{46}{0.852} & \rmseTR{37}{0.844} & \rmseTR{81}{0.869} & \rmseTR{63}{0.863} & \rmseTR{72}{0.865} & \rmseTR{90}{\textbf{0.893}} \\
    \midrule
     & Surface temp.                      & \rmseSG{28}{0.955} & \rmseSG{81}{0.973} & \rmseSG{54}{0.971} & \rmseSG{10}{0.948} & \rmseSG{19}{0.954} & \rmseSG{37}{0.955} & \rmseSG{46}{0.964} & \rmseSG{72}{0.973} & \rmseSG{63}{0.971} & \rmseSG{90}{\textbf{0.978}} \\
     & Temperature                        & \rmseSG{10}{0.448} & \rmseSG{81}{0.685} & \rmseSG{54}{0.611} & \rmseSG{19}{0.517} & \rmseSG{37}{0.581} & \rmseSG{63}{0.628} & \rmseSG{28}{0.571} & \rmseSG{46}{0.598} & \rmseSG{72}{0.680} & \rmseSG{90}{\textbf{0.692}} \\
     & Humidity                         & \rmseSG{10}{0.738} & \rmseSG{81}{0.832} & \rmseSG{63}{0.822} & \rmseSG{19}{0.744} & \rmseSG{37}{0.793} & \rmseSG{46}{0.798} & \rmseSG{28}{0.778} & \rmseSG{54}{0.801} & \rmseSG{72}{0.823} & \rmseSG{90}{\textbf{0.848}} \\
    \smash{\rotatebox[origin=c]{90}{Single-complete \textcolor{white}{aa}}}
     & Cloud fraction                     & \rmseSG{10}{0.599} & \rmseSG{37}{0.706} & \rmseSG{54}{0.719} & \rmseSG{19}{0.652} & \rmseSG{46}{0.708} & \rmseSG{28}{0.701} & \rmseSG{81}{0.732} & \rmseSG{63}{0.722} & \rmseSG{72}{0.724} & \rmseSG{90}{\textbf{0.749}} \\
     & E--W wind                & \rmseSG{10}{0.510} & \rmseSG{72}{0.717} & \rmseSG{46}{0.676} & \rmseSG{19}{0.621} & \rmseSG{37}{0.673} & \rmseSG{28}{0.668} & \rmseSG{81}{0.732} & \rmseSG{63}{0.712} & \rmseSG{54}{0.708} & \rmseSG{90}{\textbf{0.741}} \\
     & N--S wind                & \rmseSG{10}{0.566} & \rmseSG{81}{0.706} & \rmseSG{46}{0.662} & \rmseSG{19}{0.620} & \rmseSG{54}{0.666} & \rmseSG{28}{0.637} & \rmseSG{90}{\textbf{0.721}} & \rmseSG{63}{0.690} & \rmseSG{37}{0.658} & \rmseSG{72}{0.699} \\
     & ASR                      & \rmseSG{28}{0.993} & \rmseSG{46}{0.995} & \rmseSG{63}{0.996} & \rmseSG{10}{0.973} & \rmseSG{37}{0.995} & \rmseSG{72}{0.996} & \rmseSG{19}{0.993} & \rmseSG{54}{0.995} & \rmseSG{81}{0.996} & \rmseSG{90}{\textbf{0.996}} \\
     & OLR                      & \rmseSG{19}{0.797} & \rmseSG{54}{0.882} & \rmseSG{37}{0.863} & \rmseSG{10}{0.776} & \rmseSG{46}{0.871} & \rmseSG{28}{0.843} & \rmseSG{90}{\textbf{0.912}} & \rmseSG{72}{0.899} & \rmseSG{63}{0.897} & \rmseSG{81}{0.911} \\
    \bottomrule
  \end{tabular}
  }
\end{table}

\subsection{RAMSE results}\label{app:ramse-results}

Squared-error metrics reward smooth predictions: a misplaced feature is penalized both at its true and at its predicted location (the so-called ``double penalty''), so damping fine-scale amplitude lowers the expected score \citep{subichFixingDouble2025}.

To check to what extent this drives the method ranking, we report RAMSE, the square root of the adjusted MSE as defined in \citet{subichFixingDouble2025}.
For a single field with spherical-harmonic coefficients $\hat{a}_{\ell m}$ (prediction) and $a_{\ell m}$ (truth), the area-weighted MSE decomposes over degrees as
\begin{equation*}
  \|\hat{\mathbf{y}}-\mathbf{y}\|_G^2=\sum_\ell\left[\left(\sqrt{\hat{P}_\ell}-\sqrt{P_\ell}\right)^2+2\sqrt{\hat{P}_\ell P_\ell}\left(1-\mathrm{Coh}_\ell\right)\right],
  \qquad
  \mathrm{Coh}_\ell=\frac{\sum_m \hat{a}_{\ell m}a_{\ell m}}{\sqrt{\hat{P}_\ell P_\ell}},
\end{equation*}
where $\hat{P}_\ell=\sum_m\hat{a}_{\ell m}^2$ and $P_\ell$ are the band powers of prediction and truth and $\mathrm{Coh}_\ell$ is the coherence at degree $\ell$.
The $\sqrt{\hat{P}_\ell P_\ell}$ prefactor is what rewards blurring: minimizing the degree-$\ell$ term over $\hat{P}_\ell$ at fixed coherence gives $\sqrt{\hat{P}_\ell}=\mathrm{Coh}_\ell\sqrt{P_\ell}$, so the score-optimal band amplitude is the true amplitude damped by the coherence.
The adjusted MSE replaces the prefactor with $\max(\hat{P}_\ell,P_\ell)$, whose minimizer instead matches the true band power at every degree even where coherence is limited.
We compute RAMSE in the T21 spectral representation and report it per field, averaged over test examples, as for RMSE (Table~\ref{tab:ramse-all-splits}).

The RAMSE ranking closely follows the RMSE ranking. Across methods and subsets, the ratio between RAMSE and RMSE geometric means spans 1.09--1.14, so the smoothing reward of squared-error evaluation is close to method-neutral here.
This is plausibly because the smoothing preference of the L2 objective (+ the quite red spectra of most of the target fields) dominates any differences in the models' inductive biases towards smoothness. 

\begin{table}[htbp]
  \centering
  \small
  \setlength{\tabcolsep}{1.5pt}
  \renewcommand{\arraystretch}{1.1}
  \caption{RAMSE by variable and dataset subset. Lower is better. Values are the mean of five training seeds (one deterministic run for Train-mean and kNN).}
  \label{tab:ramse-all-splits}
  \resizebox{\linewidth}{!}{%
  \begin{tabular}{@{}>{\centering\arraybackslash}m{0.65cm}|>{\raggedright\arraybackslash}p{2.38cm} >{\raggedleft\arraybackslash}p{1.0cm} >{\raggedleft\arraybackslash}p{1.05cm} *{8}{>{\raggedleft\arraybackslash}p{1.32cm}}@{}}
    \toprule
    \shortstack{Sub-\\set} & Variable & Train-mean & kNN & PCA-GBT & Coord-MLP & Coord-DeepONet & PCA-MLP & ConvDec & SFNO & PPCA-ICM & GPLFR \\
    \midrule
     & Surface temp. (K)                      & \rmseMM{10}{25.1} & \rmseMM{19}{19.1} & \rmseMM{63}{12.6} & \rmseMM{28}{18.8} & \rmseMM{46}{12.8} & \rmseMM{72}{12.5} & \rmseMM{37}{13.4} & \rmseMM{54}{12.7} & \rmseMM{81}{10.4} & \rmseMM{90}{\textbf{10.1}} \\
     & Temperature (K)                        & \rmseMM{10}{20.9} & \rmseMM{19}{15.0} & \rmseMM{63}{10.6} & \rmseMM{28}{12.9} & \rmseMM{54}{10.9} & \rmseMM{72}{10.5} & \rmseMM{46}{11.3} & \rmseMM{37}{11.3} & \rmseMM{81}{9.18} & \rmseMM{90}{\textbf{8.56}} \\
     & Humidity (dex)                         & \rmseMM{10}{1.15} & \rmseMM{19}{0.800} & \rmseMM{54}{0.560} & \rmseMM{28}{0.677} & \rmseMM{63}{0.558} & \rmseMM{72}{0.547} & \rmseMM{37}{0.582} & \rmseMM{46}{0.568} & \rmseMM{81}{0.468} & \rmseMM{90}{\textbf{0.450}} \\
    \smash{\rotatebox[origin=c]{90}{Multi-partial \textcolor{white}{aa}}}
     & Cloud fraction (1)                     & \rmseMM{10}{0.151} & \rmseMM{28}{0.118} & \rmseMM{54}{0.103} & \rmseMM{19}{0.132} & \rmseMM{37}{0.106} & \rmseMM{46}{0.105} & \rmseMM{81}{0.0946} & \rmseMM{72}{0.0971} & \rmseMM{63}{0.0971} & \rmseMM{90}{\textbf{0.0895}} \\
     & E--W wind (m\,s$^{-1}$)                & \rmseMM{10}{19.4} & \rmseMM{46}{13.2} & \rmseMM{54}{13.0} & \rmseMM{19}{14.7} & \rmseMM{37}{13.2} & \rmseMM{28}{13.7} & \rmseMM{72}{12.0} & \rmseMM{63}{12.6} & \rmseMM{81}{11.5} & \rmseMM{90}{\textbf{11.3}} \\
     & N--S wind (m\,s$^{-1}$)                & \rmseMM{10}{8.28} & \rmseMM{46}{5.78} & \rmseMM{37}{5.82} & \rmseMM{19}{6.24} & \rmseMM{28}{6.01} & \rmseMM{54}{5.65} & \rmseMM{90}{\textbf{4.87}} & \rmseMM{72}{4.97} & \rmseMM{63}{5.25} & \rmseMM{81}{4.91} \\
     & ASR (W\,m$^{-2}$)                      & \rmseMM{19}{61.2} & \rmseMM{46}{35.1} & \rmseMM{63}{28.8} & \rmseMM{10}{83.6} & \rmseMM{54}{32.0} & \rmseMM{72}{27.9} & \rmseMM{28}{40.0} & \rmseMM{37}{35.3} & \rmseMM{81}{27.9} & \rmseMM{90}{\textbf{24.2}} \\
     & OLR (W\,m$^{-2}$)                      & \rmseMM{10}{43.8} & \rmseMM{28}{27.5} & \rmseMM{37}{21.1} & \rmseMM{19}{32.9} & \rmseMM{46}{20.5} & \rmseMM{63}{20.4} & \rmseMM{81}{19.5} & \rmseMM{54}{20.5} & \rmseMM{72}{19.8} & \rmseMM{90}{\textbf{17.3}} \\
    \midrule
     & Surface temp. (K)                      & \rmseTR{10}{24.5} & \rmseTR{19}{19.9} & \rmseTR{46}{12.8} & \rmseTR{28}{17.7} & \rmseTR{72}{12.5} & \rmseTR{63}{12.7} & \rmseTR{37}{13.4} & \rmseTR{54}{12.7} & \rmseTR{81}{11.5} & \rmseTR{90}{\textbf{9.99}} \\
     & Temperature (K)                        & \rmseTR{10}{19.8} & \rmseTR{19}{15.6} & \rmseTR{54}{10.2} & \rmseTR{28}{12.3} & \rmseTR{63}{10.2} & \rmseTR{81}{9.82} & \rmseTR{37}{11.1} & \rmseTR{46}{11.1} & \rmseTR{72}{9.92} & \rmseTR{90}{\textbf{8.44}} \\
     & Humidity (dex)                         & \rmseTR{10}{1.12} & \rmseTR{19}{0.785} & \rmseTR{63}{0.530} & \rmseTR{28}{0.669} & \rmseTR{72}{0.521} & \rmseTR{54}{0.530} & \rmseTR{37}{0.562} & \rmseTR{46}{0.547} & \rmseTR{81}{0.482} & \rmseTR{90}{\textbf{0.422}} \\
    \smash{\rotatebox[origin=c]{90}{Multi-complete \textcolor{white}{aa}}}
     & Cloud fraction (1)                     & \rmseTR{10}{0.159} & \rmseTR{28}{0.127} & \rmseTR{54}{0.105} & \rmseTR{19}{0.135} & \rmseTR{37}{0.114} & \rmseTR{46}{0.108} & \rmseTR{81}{0.100} & \rmseTR{63}{0.102} & \rmseTR{72}{0.101} & \rmseTR{90}{\textbf{0.0940}} \\
     & E--W wind (m\,s$^{-1}$)                & \rmseTR{10}{17.8} & \rmseTR{19}{12.4} & \rmseTR{63}{10.8} & \rmseTR{28}{12.2} & \rmseTR{37}{11.6} & \rmseTR{46}{11.4} & \rmseTR{72}{10.3} & \rmseTR{54}{11.1} & \rmseTR{81}{10.2} & \rmseTR{90}{\textbf{9.81}} \\
     & N--S wind (m\,s$^{-1}$)                & \rmseTR{10}{7.63} & \rmseTR{19}{6.03} & \rmseTR{54}{5.38} & \rmseTR{37}{5.93} & \rmseTR{28}{5.96} & \rmseTR{46}{5.64} & \rmseTR{90}{\textbf{4.78}} & \rmseTR{72}{4.91} & \rmseTR{63}{5.07} & \rmseTR{81}{4.86} \\
     & ASR (W\,m$^{-2}$)                      & \rmseTR{19}{61.8} & \rmseTR{46}{33.5} & \rmseTR{63}{28.1} & \rmseTR{10}{74.6} & \rmseTR{54}{32.1} & \rmseTR{81}{27.2} & \rmseTR{28}{38.9} & \rmseTR{37}{33.5} & \rmseTR{72}{27.9} & \rmseTR{90}{\textbf{23.5}} \\
     & OLR (W\,m$^{-2}$)                      & \rmseTR{10}{44.0} & \rmseTR{28}{26.4} & \rmseTR{37}{20.7} & \rmseTR{19}{29.3} & \rmseTR{63}{19.8} & \rmseTR{46}{20.2} & \rmseTR{81}{19.0} & \rmseTR{54}{19.8} & \rmseTR{72}{19.6} & \rmseTR{90}{\textbf{16.6}} \\
    \midrule
     & Surface temp. (K)                      & \rmseSG{10}{21.2} & \rmseSG{28}{14.3} & \rmseSG{46}{11.9} & \rmseSG{19}{15.5} & \rmseSG{37}{12.4} & \rmseSG{54}{11.7} & \rmseSG{63}{11.1} & \rmseSG{72}{10.3} & \rmseSG{81}{9.89} & \rmseSG{90}{\textbf{9.56}} \\
     & Temperature (K)                        & \rmseSG{10}{19.1} & \rmseSG{19}{12.1} & \rmseSG{46}{10.2} & \rmseSG{28}{10.4} & \rmseSG{37}{10.2} & \rmseSG{54}{10.1} & \rmseSG{63}{9.29} & \rmseSG{72}{8.98} & \rmseSG{81}{8.68} & \rmseSG{90}{\textbf{8.36}} \\
     & Humidity (dex)                         & \rmseSG{10}{1.09} & \rmseSG{28}{0.597} & \rmseSG{46}{0.572} & \rmseSG{19}{0.616} & \rmseSG{37}{0.573} & \rmseSG{54}{0.544} & \rmseSG{63}{0.500} & \rmseSG{72}{0.490} & \rmseSG{81}{0.474} & \rmseSG{90}{\textbf{0.463}} \\
    \smash{\rotatebox[origin=c]{90}{Single-complete \textcolor{white}{aa}}}
     & Cloud fraction (1)                     & \rmseSG{10}{0.151} & \rmseSG{46}{0.106} & \rmseSG{54}{0.101} & \rmseSG{19}{0.126} & \rmseSG{28}{0.107} & \rmseSG{37}{0.107} & \rmseSG{81}{0.0892} & \rmseSG{72}{0.0945} & \rmseSG{63}{0.0957} & \rmseSG{90}{\textbf{0.0877}} \\
     & E--W wind (m\,s$^{-1}$)                & \rmseSG{10}{15.2} & \rmseSG{54}{10.3} & \rmseSG{46}{10.3} & \rmseSG{37}{11.0} & \rmseSG{19}{11.7} & \rmseSG{28}{11.6} & \rmseSG{90}{\textbf{8.79}} & \rmseSG{63}{9.97} & \rmseSG{72}{9.81} & \rmseSG{81}{8.85} \\
     & N--S wind (m\,s$^{-1}$)                & \rmseSG{10}{6.29} & \rmseSG{63}{4.60} & \rmseSG{54}{4.76} & \rmseSG{37}{5.12} & \rmseSG{19}{5.25} & \rmseSG{28}{5.22} & \rmseSG{90}{\textbf{4.03}} & \rmseSG{72}{4.31} & \rmseSG{46}{4.78} & \rmseSG{81}{4.21} \\
     & ASR (W\,m$^{-2}$)                      & \rmseSG{19}{47.6} & \rmseSG{54}{30.3} & \rmseSG{63}{29.5} & \rmseSG{10}{73.2} & \rmseSG{46}{32.5} & \rmseSG{72}{27.3} & \rmseSG{28}{38.9} & \rmseSG{37}{33.5} & \rmseSG{90}{\textbf{25.7}} & \rmseSG{81}{26.4} \\
     & OLR (W\,m$^{-2}$)                      & \rmseSG{10}{36.8} & \rmseSG{28}{21.8} & \rmseSG{37}{21.5} & \rmseSG{19}{29.4} & \rmseSG{54}{20.5} & \rmseSG{46}{20.6} & \rmseSG{63}{18.7} & \rmseSG{72}{18.1} & \rmseSG{81}{17.3} & \rmseSG{90}{\textbf{17.2}} \\
    \bottomrule
  \end{tabular}
  }
\end{table}

\subsection{Probabilistic evaluation results}\label{app:probabilistic-results}

PPCA-ICM and GPLFR are our two naturally probabilistic baselines, and the best-performing ones by RMSE.
Here we show their energy scores (Table~\ref{tab:energy-score}) and spread--skill ratios (Table~\ref{tab:ssr}) on the three subsets.

\paragraph{Energy score.}
The energy score is
\begin{equation*}
  \mathrm{ES}(p,\mathbf{y}) = \mathbb{E}_{\mathbf{y}'\sim p}\lVert\mathbf{y}'-\mathbf{y}\rVert_G - \tfrac{1}{2}\mathbb{E}_{\mathbf{y}',\mathbf{y}''\sim p}\lVert\mathbf{y}'-\mathbf{y}''\rVert_G,
\end{equation*}
where $p$ is the predictive distribution, $\mathbf{y}$ is the ground-truth output, and $\lVert\mathbf{e}\rVert_G = \sqrt{\mathbf{e}^\top\mathbf{G}\mathbf{e}}$ with $\mathbf{G}$ containing latitude area weights.
The first term penalizes each draw's distance from the truth, which grows with both the predictive distribution's misplacement and its spread; the second rewards spread, at a rate balanced such that under- and overconfidence both worsen the expected score; lower is better.
We estimate the energy score using 64 posterior predictive samples.
Its results largely follow the RMSE results, with GPLFR beating PPCA-ICM on most variables and subsets (22 of the 24 variable--subset combinations) (Table~\ref{tab:energy-score}).

\paragraph{Spread--skill ratio.}
SSR is a first-order diagnostic of ensemble calibration.
We define it as
\begin{equation*}
  \mathrm{SSR}=\sqrt{\frac{\sum_i\mathrm{Spread}_i^2}{\sum_i\mathrm{MSE}_i}},
\end{equation*}
where for each test example $i$,
\begin{equation*}
  \mathrm{Spread}_i^2=\frac{1}{M-1}\sum_{m=1}^M\|\mathbf{y}^{[m]}_i-\hat{\mathbf{y}}_i\|_G^2,
  \qquad
  \mathrm{MSE}_i=\|\hat{\mathbf{y}}_i-\mathbf{y}_i\|_G^2,
\end{equation*}
and $\hat{\mathbf{y}}_i$ is the ensemble mean.
A well-calibrated ensemble has SSR $\approx 1$; values below 1 indicate overconfidence (too little spread) and values above 1 indicate underconfidence.

On SSR, PPCA-ICM is consistently overconfident, while GPLFR is more balanced between over- and underconfidence, leaning overconfident on Single-complete (Table~\ref{tab:ssr}).
Both models improve going from Single-complete to the Multi- subsets.
By this measure, GPLFR is better calibrated than PPCA-ICM on all variables and subsets.

\begin{table}[t]
  \centering
  \small
  \setlength{\tabcolsep}{3pt}
  \renewcommand{\arraystretch}{1.1}
  \caption{Energy score by variable and subset. Lower is better. Scores are the mean of five training seeds.}
  \label{tab:energy-score}
  \begin{tabular}{@{}>{\raggedright\arraybackslash}p{2.45cm} *{6}{>{\raggedleft\arraybackslash}p{1.65cm}}@{}}
    \toprule
    & \multicolumn{2}{c}{Multi-partial} & \multicolumn{2}{c}{Multi-complete} & \multicolumn{2}{c}{Single-complete} \\
    \cmidrule(lr){2-3}\cmidrule(lr){4-5}\cmidrule(lr){6-7}
    Variable & PPCA-ICM & GPLFR & PPCA-ICM & GPLFR & PPCA-ICM & GPLFR \\
    \midrule
    Surface temp. (K)                      & \metricMM{90}{\textbf{7.82}} & \metricMM{81}{7.98} & \metricTR{81}{8.75} & \metricTR{90}{\textbf{8.03}} & \metricSG{81}{8.34} & \metricSG{90}{\textbf{7.57}} \\
    Temperature (K)                        & \metricMM{81}{6.89} & \metricMM{90}{\textbf{6.68}} & \metricTR{81}{7.64} & \metricTR{90}{\textbf{6.67}} & \metricSG{81}{7.18} & \metricSG{90}{\textbf{6.41}} \\
    Humidity (dex)                         & \metricMM{90}{\textbf{0.345}} & \metricMM{81}{0.358} & \metricTR{81}{0.355} & \metricTR{90}{\textbf{0.341}} & \metricSG{81}{0.399} & \metricSG{90}{\textbf{0.354}} \\
    Cloud fraction (1)                     & \metricMM{81}{0.0674} & \metricMM{90}{\textbf{0.0586}} & \metricTR{81}{0.0707} & \metricTR{90}{\textbf{0.0623}} & \metricSG{81}{0.0725} & \metricSG{90}{\textbf{0.0566}} \\
    E--W wind (m\,s$^{-1}$)                & \metricMM{81}{8.02} & \metricMM{90}{\textbf{7.31}} & \metricTR{81}{7.27} & \metricTR{90}{\textbf{6.36}} & \metricSG{81}{7.31} & \metricSG{90}{\textbf{5.93}} \\
    N--S wind (m\,s$^{-1}$)                & \metricMM{81}{3.58} & \metricMM{90}{\textbf{3.17}} & \metricTR{81}{3.48} & \metricTR{90}{\textbf{3.12}} & \metricSG{81}{3.59} & \metricSG{90}{\textbf{2.77}} \\
    ASR (W\,m$^{-2}$)                      & \metricMM{81}{20.7} & \metricMM{90}{\textbf{18.7}} & \metricTR{81}{20.8} & \metricTR{90}{\textbf{18.3}} & \metricSG{81}{21.9} & \metricSG{90}{\textbf{19.8}} \\
    OLR (W\,m$^{-2}$)                      & \metricMM{81}{14.2} & \metricMM{90}{\textbf{12.7}} & \metricTR{81}{14.2} & \metricTR{90}{\textbf{12.3}} & \metricSG{81}{14.2} & \metricSG{90}{\textbf{12.9}} \\
    \bottomrule
  \end{tabular}
\end{table}

\begin{table}[t]
  \centering
  \small
  \setlength{\tabcolsep}{3pt}
  \renewcommand{\arraystretch}{1.1}
  \caption{Spread--skill ratio (SSR) by variable and subset. Closer to one is better. Values are the mean of five training seeds.}
  \label{tab:ssr}
  \begin{tabular}{@{}>{\raggedright\arraybackslash}p{2.45cm} *{6}{>{\raggedleft\arraybackslash}p{1.65cm}}@{}}
    \toprule
    & \multicolumn{2}{c}{Multi-partial} & \multicolumn{2}{c}{Multi-complete} & \multicolumn{2}{c}{Single-complete} \\
    \cmidrule(lr){2-3}\cmidrule(lr){4-5}\cmidrule(lr){6-7}
    Variable & PPCA-ICM & GPLFR & PPCA-ICM & GPLFR & PPCA-ICM & GPLFR \\
    \midrule
    Surface temp.                      & \metricMM{81}{0.686} & \metricMM{90}{\textbf{1.091}} & \metricTR{81}{0.628} & \metricTR{90}{\textbf{1.113}} & \metricSG{81}{0.241} & \metricSG{90}{\textbf{0.887}} \\
    Temperature                        & \metricMM{81}{0.626} & \metricMM{90}{\textbf{1.031}} & \metricTR{81}{0.541} & \metricTR{90}{\textbf{0.996}} & \metricSG{81}{0.239} & \metricSG{90}{\textbf{0.911}} \\
    Humidity                         & \metricMM{81}{0.741} & \metricMM{90}{\textbf{1.166}} & \metricTR{81}{0.694} & \metricTR{90}{\textbf{1.214}} & \metricSG{81}{0.205} & \metricSG{90}{\textbf{0.847}} \\
    Cloud fraction                     & \metricMM{81}{0.394} & \metricMM{90}{\textbf{1.138}} & \metricTR{81}{0.377} & \metricTR{90}{\textbf{1.199}} & \metricSG{81}{0.181} & \metricSG{90}{\textbf{0.926}} \\
    E--W wind                & \metricMM{81}{0.305} & \metricMM{90}{\textbf{0.678}} & \metricTR{81}{0.285} & \metricTR{90}{\textbf{0.682}} & \metricSG{81}{0.174} & \metricSG{90}{\textbf{0.764}} \\
    N--S wind                & \metricMM{81}{0.374} & \metricMM{90}{\textbf{0.853}} & \metricTR{81}{0.341} & \metricTR{90}{\textbf{0.837}} & \metricSG{81}{0.155} & \metricSG{90}{\textbf{0.717}} \\
    ASR                      & \metricMM{81}{0.393} & \metricMM{90}{\textbf{1.041}} & \metricTR{81}{0.401} & \metricTR{90}{\textbf{1.104}} & \metricSG{81}{0.145} & \metricSG{90}{\textbf{0.678}} \\
    OLR                      & \metricMM{81}{0.474} & \metricMM{90}{\textbf{1.087}} & \metricTR{81}{0.489} & \metricTR{90}{\textbf{1.139}} & \metricSG{81}{0.204} & \metricSG{90}{\textbf{0.966}} \\
    \bottomrule
  \end{tabular}
\end{table}

\subsection{Relative energy score results}\label{app:relative-energy-score-results}

\begin{table}[htbp]
  \centering
  \small
  \setlength{\tabcolsep}{3pt}
  \renewcommand{\arraystretch}{1.1}
  \caption{Relative energy score by variable under the shared-planets protocol for probabilistic methods. Bold indicates lowest (best) score within each subset. Scores are the mean of five training seeds.}
  \label{tab:relative-energy-score}
  \begin{tabular}{@{}>{\raggedright\arraybackslash}p{2.45cm} *{4}{>{\raggedleft\arraybackslash}p{1.65cm}}@{}}
    \toprule
    & \multicolumn{2}{c}{Multi-partial} & \multicolumn{2}{c}{Multi-complete} \\
    \cmidrule(lr){2-3}\cmidrule(lr){4-5}
    Variable & PPCA-ICM & GPLFR & PPCA-ICM & GPLFR \\
    \midrule
    Surface temp.                      & \metricMM{81}{0.712} & \metricMM{90}{\textbf{0.623}} & \metricTR{81}{0.802} & \metricTR{90}{\textbf{0.673}} \\
    Temperature                        & \metricMM{81}{0.638} & \metricMM{90}{\textbf{0.551}} & \metricTR{81}{0.724} & \metricTR{90}{\textbf{0.578}} \\
    Humidity                         & \metricMM{81}{0.590} & \metricMM{90}{\textbf{0.563}} & \metricTR{81}{0.639} & \metricTR{90}{\textbf{0.591}} \\
    Cloud fraction                     & \metricMM{81}{0.505} & \metricMM{90}{\textbf{0.421}} & \metricTR{81}{0.495} & \metricTR{90}{\textbf{0.430}} \\
    E--W wind                & \metricMM{81}{0.636} & \metricMM{90}{\textbf{0.581}} & \metricTR{81}{0.702} & \metricTR{90}{\textbf{0.605}} \\
    N--S wind                & \metricMM{81}{0.653} & \metricMM{90}{\textbf{0.585}} & \metricTR{81}{0.657} & \metricTR{90}{\textbf{0.602}} \\
    ASR                      & \metricMM{81}{0.562} & \metricMM{90}{\textbf{0.476}} & \metricTR{81}{0.582} & \metricTR{90}{\textbf{0.487}} \\
    OLR                      & \metricMM{81}{0.607} & \metricMM{90}{\textbf{0.504}} & \metricTR{81}{0.620} & \metricTR{90}{\textbf{0.511}} \\
    \midrule
    Geometric mean                       & \metricMM{81}{0.610} & \metricMM{90}{\textbf{0.534}} & \metricTR{81}{0.647} & \metricTR{90}{\textbf{0.555}} \\
    \bottomrule
  \end{tabular}
\end{table}

Table~\ref{tab:relative-energy-score} reports relative energy scores for the probabilistic methods (PPCA-ICM, GPLFR) under the shared-planets protocol (Section~\ref{sec:eval}).
Following the same ratio structure as the relative RMSE, we define
\begin{equation*}
  \mathrm{ES}_\mathrm{rel} = \frac{\mathrm{ES}(p,\mathbf{y}^s)}{\|\mathbf{y}^{s'}-\mathbf{y}^s\|_G},
\end{equation*}
where the denominator is the other GCM's energy score, which reduces to the RMSE because GCMs only produce point predictions (i.e., delta-function predictive distributions).
Probabilistic emulators that produce well-calibrated spreads can therefore achieve lower relative energy scores than relative RMSEs, reflecting the additional information in a calibrated predictive distribution over a point estimate.

Both methods achieve relative energy score below 1 on every variable in both Multi- subsets.
GPLFR's geometric mean relative energy score is 0.534 on Multi-partial, 20\% lower than its geometric mean relative RMSE (0.669; Table~\ref{tab:relative-rmse}), indicating the value of a distributional prediction over a point prediction (since both metrics share a denominator).
PPCA-ICM shows a similar drop of about 21\%; the Multi-complete drops are 19\% and 21\%.

\subsection{Per-example relative RMSE results}\label{app:per-example-relative-rmse}

\begin{figure}[t]
  \centering
  \includegraphics[width=\linewidth]{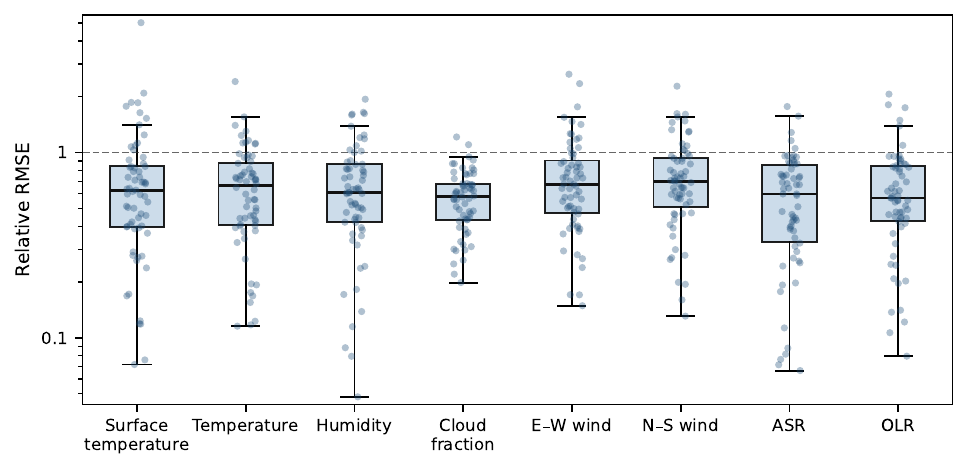}
  \caption{Per-example relative RMSE distribution for GPLFR on Multi-partial, grouped by variable. Each point is one example ((planet, GCM) combination). The dashed line at 1 marks the GCM-disagreement threshold. Boxes show the interquartile range; whiskers extend to $1.5\times$ IQR. Scores are from training seed 0.}
  \label{fig:per-example-relative-rmse}
\end{figure}

Figure~\ref{fig:per-example-relative-rmse} shows the distribution of per-example relative RMSEs for GPLFR on the Multi-partial dataset.
While the medians are well below 1 for all variable groups (ranging from 0.54 for ASR to 0.68 for N--S wind), the spread is significant.
Surface temperature has the widest distribution, with two examples exceeding a relative RMSE of 2 (2.1 and 5.0 respectively), meaning the emulator's error is far larger than GCM disagreement for those examples.
The two worst examples overall by geometric-mean relative RMSE (simulations 28 and 1581, at 1.55 and 1.51; use \texttt{simulation\_id} in the code/data) are both ExoCAM cases, consistent with it having fewer training data than the UM.

The high relative RMSE tails suggest that certain (planet, GCM) combinations remain difficult to emulate reliably and represent natural targets for future method development or additional training data.
We show predictions for these two outliers alongside two more typical examples (simulations 16 and 1629) in Figures~\ref{fig:example-temperature-winds}, \ref{fig:example-asr}, and~\ref{fig:example-vertical-profiles}.

\FloatBarrier
\subsection{Even more results tables}\label{app:more-results-tables}
Per-variable and per-level result tables for every combination of subset, applicable metric, and evaluation protocol are shown for all baselines trained with training seed 0 here in the benchmark code: \url{https://github.com/astroautomata/ThousandWorlds/tree/main/results/tables}.

\FloatBarrier
\section{Example predictions and climate diagnostics}\label{app:example-predictions}

\begin{figure}[ht]
  \centering
  \includegraphics[width=\linewidth]{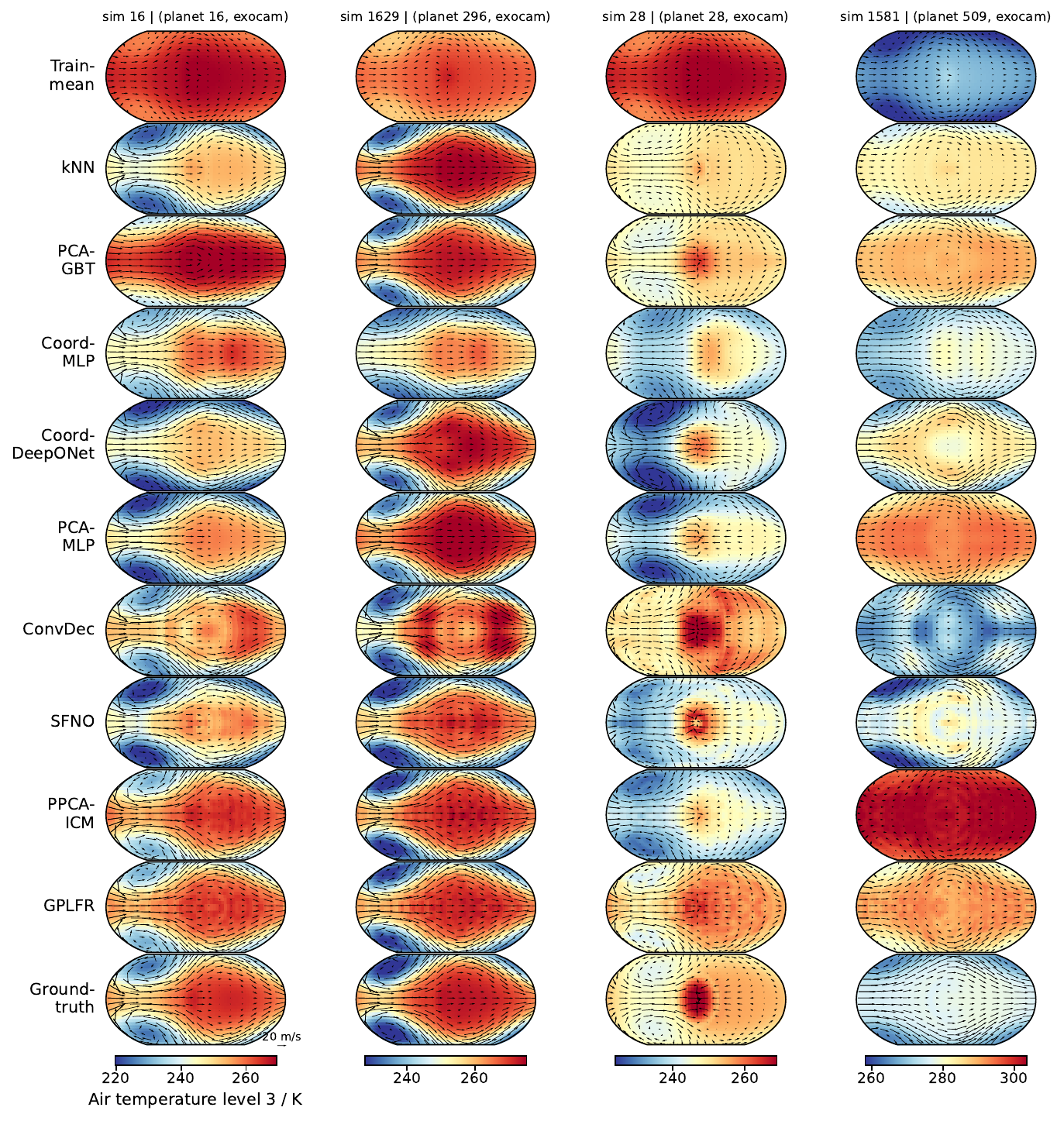}
  \caption{Spatial maps of temperature with superimposed wind vectors at relative isobar $\sigma_3\approx 0.72$ (this would be around the mid-troposphere on Earth) for four test examples. Simulations 28 and 1581 are the two worst predicted by our strongest baseline (GPLFR), while simulations 16 and 1629 give more typical predictions.}
  \label{fig:example-temperature-winds}
\end{figure}

\begin{figure}[t]
  \centering
  \includegraphics[width=\linewidth]{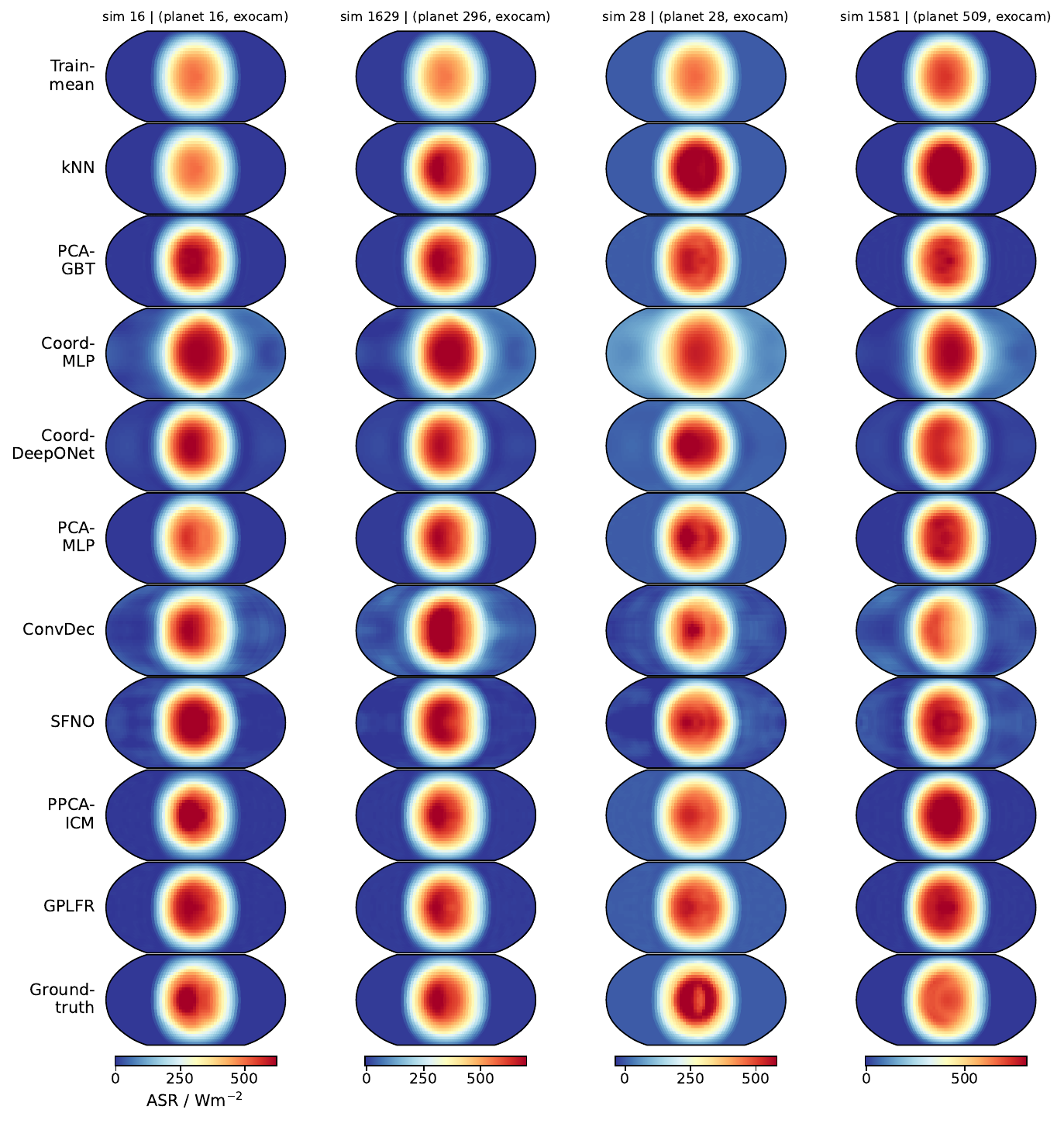}
  \caption{Absorbed shortwave radiation (ASR) maps for four test examples. Simulations 28 and 1581 are the two worst predicted by our strongest baseline (GPLFR), while simulations 16 and 1629 give more typical predictions.}
  \label{fig:example-asr}
\end{figure}

\begin{figure}[t]
  \centering
  \includegraphics[width=\linewidth]{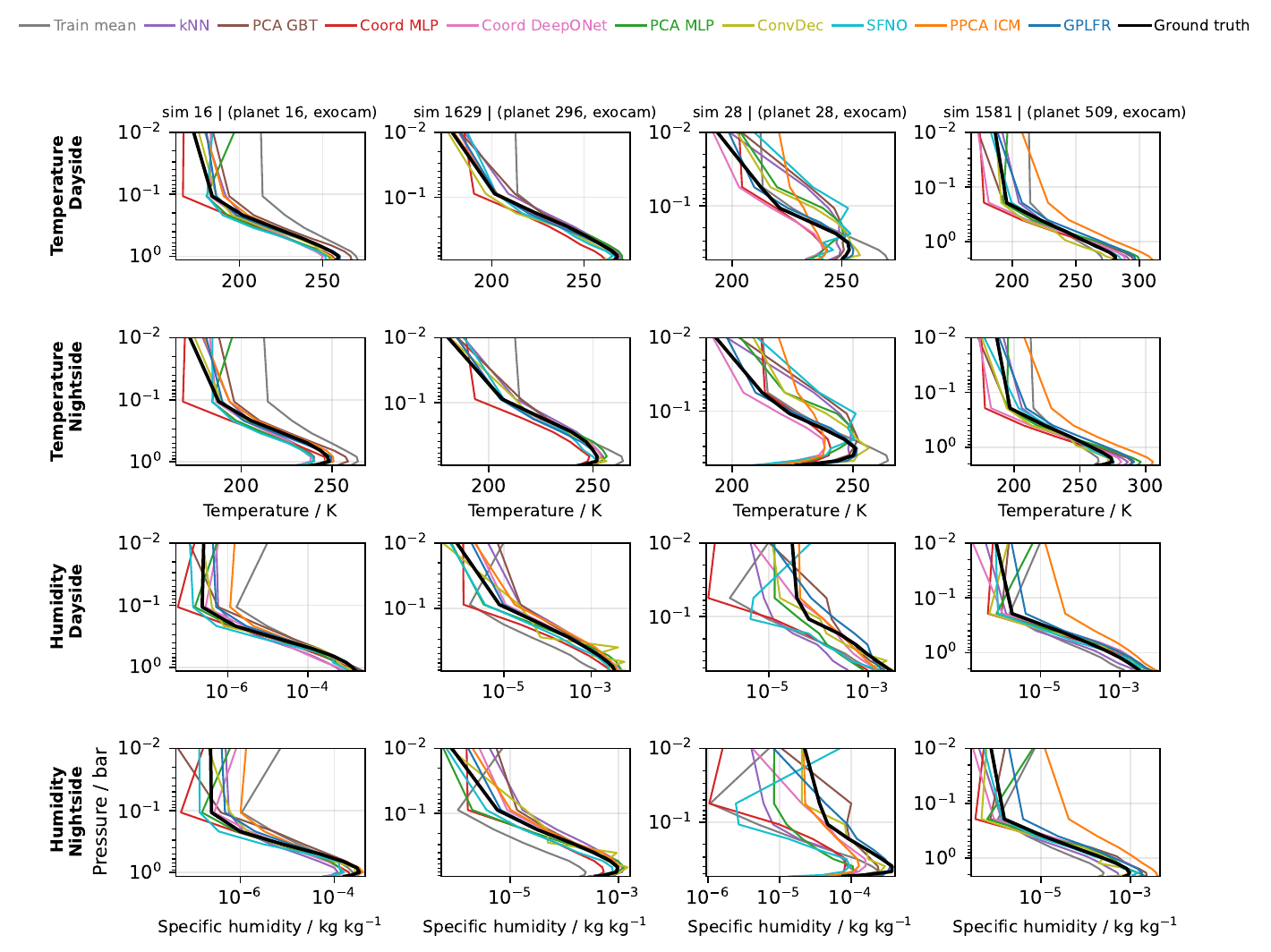}
  \caption{Dayside and nightside vertical profiles of area-weighted mean temperature and specific humidity for four test examples. Simulations 28 and 1581 are the two worst predicted by our strongest baseline (GPLFR), while simulations 16 and 1629 give more typical predictions.}
  \label{fig:example-vertical-profiles}
\end{figure}

\begin{figure}[t]
  \centering
  \includegraphics[width=\linewidth]{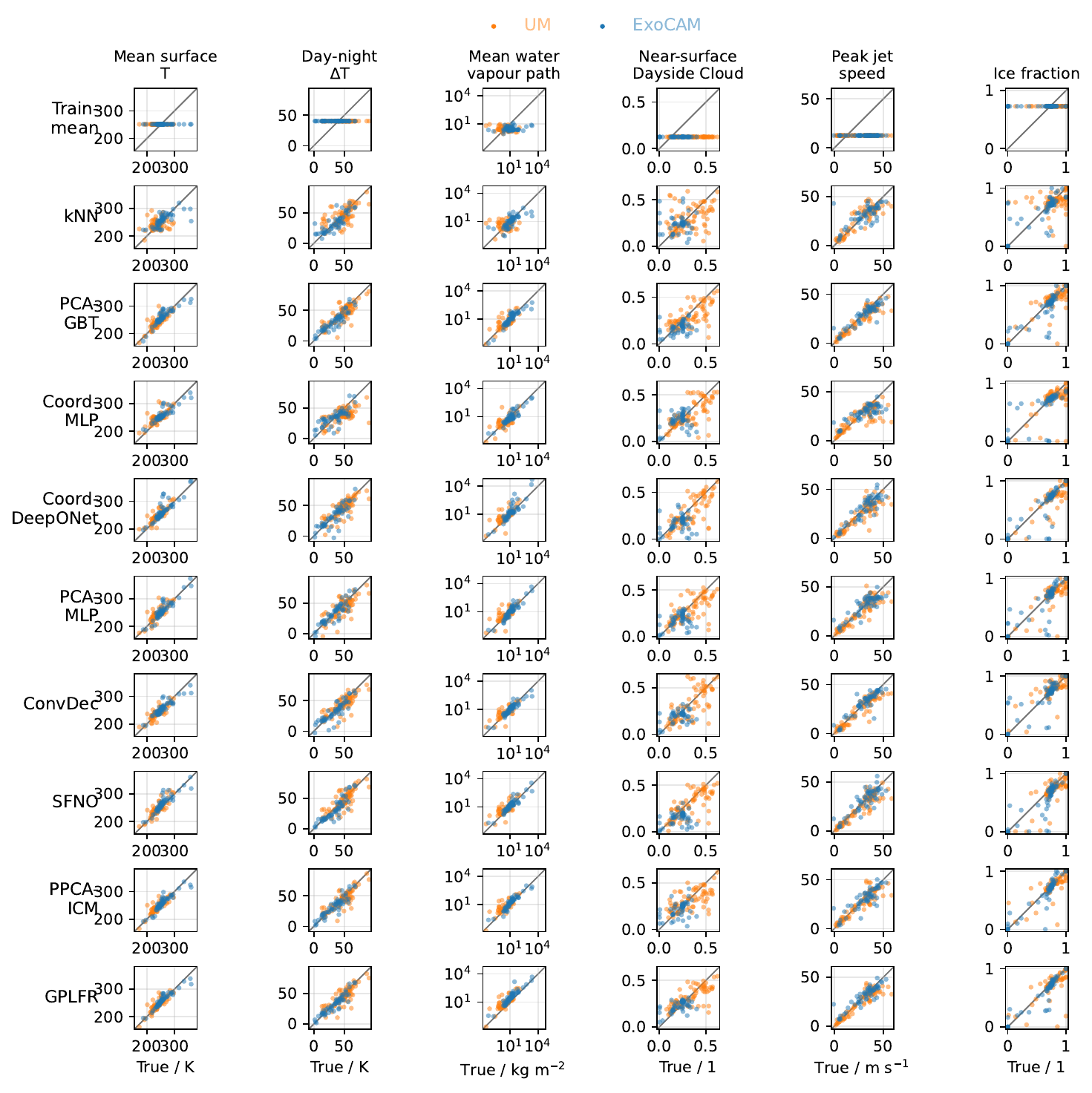}
  \caption{Predicted (y-axis) versus true (x-axis) values for six climate diagnostics across the Multi-partial test set. Each point is one test example; the line marks perfect prediction. Points are coloured by the source GCM. Mean surface $T$: area-weighted global-mean surface temperature. Day--night $\Delta T$: area-weighted mean dayside minus nightside surface temperature. Water vapour path: total mass of water vapour in an atmospheric column, integrated over pressure levels and area-weighted over the globe. Near-surface dayside cloud: cloud fraction at the lowest pressure level, averaged over dayside grid cells. Peak jet speed: maximum of the longitudinally averaged E--W wind over latitudes and pressure levels. Ice fraction: fraction of surface area with $T_\text{surf} < 273\,$K.}
  \label{fig:climate-diagnostics}
\end{figure}

Here we show some ground truth and baseline predictions using common plots and climate diagnostics from exoplanet science.
Figures~\ref{fig:example-temperature-winds}, \ref{fig:example-asr}, and \ref{fig:example-vertical-profiles} show plots for four test examples: 28 and 1581 are two of the worst predicted by our strongest baseline (GPLFR); the other two are more typical predictions.
Figure~\ref{fig:climate-diagnostics} shows six scalar climate diagnostics for all planets in the Multi-partial test set.

\clearpage
\section*{NeurIPS Paper Checklist}

\begin{enumerate}

\item {\bf Claims}
    \item[] Question: Do the main claims made in the abstract and introduction accurately reflect the paper's contributions and scope?
    \item[] Answer: \answerYes{}
    \item[] Justification: The abstract and introduction (Section~\ref{sec:intro}) state three contributions: the dataset, evaluation protocols, and baseline results. The Dataset and Experiments sections (Sections~\ref{sec:dataset}, \ref{sec:experiments}) match these claims. Scope limitations (tidally locked waterworlds only, low-data regime) are stated explicitly in the Introduction and Dataset sections (Sections~\ref{sec:intro}, \ref{sec:dataset}).

\item {\bf Limitations}
    \item[] Question: Does the paper discuss the limitations of the work performed by the authors?
    \item[] Answer: \answerYes{}
    \item[] Justification: Appendix~\ref{app:limitations} discusses scope restrictions (single planet class), the composition and shared heritage of the GCMs, and test-set sizes. Section~\ref{sec:challenges} discusses inter-GCM disagreement and structured missingness as fundamental dataset limitations.

\item {\bf Theory assumptions and proofs}
    \item[] Question: For each theoretical result, does the paper provide the full set of assumptions and a complete (and correct) proof?
    \item[] Answer: \answerNA{}
    \item[] Justification: The paper is a dataset and benchmark contribution; it does not contain theoretical results or proofs.

    \item {\bf Experimental result reproducibility}
    \item[] Question: Does the paper fully disclose all the information needed to reproduce the main experimental results of the paper to the extent that it affects the main claims and/or conclusions of the paper (regardless of whether the code and data are provided or not)?
    \item[] Answer: \answerYes{}
    \item[] Justification: All preprocessing, data splits, model details, hyperparameter search grids, and selected values are documented, briefly in main text (Section~\ref{sec:dataset}) and extensively in the Appendix (Appendix~\ref{app:baselines}). The dataset is publicly available at \url{https://doi.org/10.57967/hf/8695}. Code including scripts for running all baseline models is at \url{https://github.com/astroautomata/ThousandWorlds}.

\item {\bf Open access to data and code}
    \item[] Question: Does the paper provide open access to the data and code, with sufficient instructions to faithfully reproduce the main experimental results, as described in supplemental material?
    \item[] Answer: \answerYes{}
    \item[] Justification: The dataset is available at \url{https://doi.org/10.57967/hf/8695}. The code is at \url{https://github.com/astroautomata/ThousandWorlds}.

\item {\bf Experimental setting/details}
    \item[] Question: Does the paper specify all the training and test details (e.g., data splits, hyperparameters, how they were chosen, type of optimizer) necessary to understand the results?
    \item[] Answer: \answerYes{}
    \item[] Justification: The main high-level information is given in Sections~\ref{sec:subsets}--\ref{sec:baselines}. The low-level details are given in Appendices~\ref{app:baseline-shared-settings}--\ref{app:compute}.

\item {\bf Experiment statistical significance}
    \item[] Question: Does the paper report error bars suitably and correctly defined or other appropriate information about the statistical significance of the experiments?
    \item[] Answer: \answerYes{}
    \item[] Justification: Learned-method variability across five random training seeds is reported as mean $\pm$ standard deviation for the main text results (Appendix~\ref{app:tables-with-seed-variability}). Finite-test-set uncertainty is also reported using paired bootstrap intervals over test examples (Appendix~\ref{app:rmse-bootstrap-intervals}). 

\item {\bf Experiments compute resources}
    \item[] Question: For each experiment, does the paper provide sufficient information on the computer resources (type of compute workers, memory, time of execution) needed to reproduce the experiments?
    \item[] Answer: \answerYes{}
    \item[] Justification: Baseline training hardware and wall-clock times are reported in Appendix~\ref{app:compute}.
    
\item {\bf Code of ethics}
    \item[] Question: Does the research conducted in the paper conform, in every respect, with the NeurIPS Code of Ethics \url{https://neurips.cc/public/EthicsGuidelines}?
    \item[] Answer: \answerYes{}
    \item[] Justification: The research conforms to the NeurIPS Code of Ethics. No humans, animals, or planets were harmed in the making of this dataset.

\item {\bf Broader impacts}
    \item[] Question: Does the paper discuss both potential positive societal impacts and negative societal impacts of the work performed?
    \item[] Answer: \answerYes{}
    \item[] Justification: Positive: the benchmark supports exoplanet climate research, and indirectly Earth climate research as well, and may reduce GCM compute in future studies. Negative: we could not identify any -- exoplanets are too far away.
    
\item {\bf Safeguards}
    \item[] Question: Does the paper describe safeguards that have been put in place for responsible release of data or models that have a high risk for misuse (e.g., pre-trained language models, image generators, or scraped datasets)?
    \item[] Answer: \answerNA{}
    \item[] We do not foresee any misuse risks from this dataset.

\item {\bf Licenses for existing assets}
    \item[] Question: Are the creators or original owners of assets (e.g., code, data, models), used in the paper, properly credited and are the license and terms of use explicitly mentioned and properly respected?
    \item[] Answer: \answerYes{}
    \item[] Justification: Source publications for all literature simulations are cited in Table~\ref{tab:dataset-composition}. GCM implementations (ExoCAM, UM, ExoPlaSim, LFRic) are credited with citations and URLs where relevant (Appendix~\ref{app:gcms}).

\item {\bf New assets}
    \item[] Question: Are new assets introduced in the paper well documented and is the documentation provided alongside the assets?
    \item[] Answer: \answerYes{}
    \item[] Justification: The ThousandWorlds dataset is documented in Section~\ref{sec:dataset}. The benchmark dataset is released on Hugging Face with DOI-bearing metadata plus accompanying documentation and a license. The code is released on GitHub with documentation and a license too.

\item {\bf Crowdsourcing and research with human subjects}
    \item[] Question: For crowdsourcing experiments and research with human subjects, does the paper include the full text of instructions given to participants and screenshots, if applicable, as well as details about compensation (if any)? 
    \item[] Answer: \answerNA{}
    \item[] Justification: No crowdsourcing or human subjects research was involved.

\item {\bf Institutional review board (IRB) approvals or equivalent for research with human subjects}
    \item[] Question: Does the paper describe potential risks incurred by study participants, whether such risks were disclosed to the subjects, and whether Institutional Review Board (IRB) approvals (or an equivalent approval/review based on the requirements of your country or institution) were obtained?
    \item[] Answer: \answerNA{}
    \item[] Justification: No human subjects research was involved.

\item {\bf Declaration of LLM usage}
    \item[] Question: Does the paper describe the usage of LLMs if it is an important, original, or non-standard component of the core methods in this research? Note that if the LLM is used only for writing, editing, or formatting purposes and does \emph{not} impact the core methodology, scientific rigor, or originality of the research, declaration is not required.
    %this research? 
    \item[] Answer: \answerNA{}
    \item[] Justification: LLMs were not used as an important, original, or non-standard component of the core methodology.

\end{enumerate}

\end{document}